\documentclass[11pt, letterpaper, logo, onecolumn, copyright]{main}

\usepackage[authoryear, sort&compress, round]{natbib}

\usepackage[inkscapeformat=png]{svg}

\usepackage[most, breakable, skins]{tcolorbox}
\usepackage{caption}
\tcbuselibrary{skins}
\usepackage{lipsum}
\usepackage{tabularx}
\usepackage{afterpage}
\usepackage{booktabs}
\usepackage{xurl} 
\usepackage{subcaption}
\usepackage{makecell}
\usepackage{amssymb}
\usepackage{mathtools}
\usepackage{multirow}
\usepackage{bm}
\usepackage{multicol}
\usepackage{array}
\usepackage{float}
\usepackage{placeins}
\usepackage{listings, listings-rust}
\usepackage{fontawesome5}
\usepackage{hyperref}
\usepackage{graphicx}
\usepackage[dvipsnames]{xcolor}
\usepackage{cleveref}
\usepackage{longtable}
\usepackage{pdflscape}
\usepackage{adjustbox}
\usepackage{nicematrix}
\usepackage{CJKutf8}
\usepackage{ragged2e}
\usepackage{colortbl}
\usepackage{enumitem}
\usepackage{bbm}
\usepackage{pifont}
\usepackage{hyphenat}
\usepackage{amsmath}
\usepackage{amsthm}  
\usepackage{circledsteps}
\usepackage{diagbox}
\usepackage{bookmark}
\usepackage{textcomp}

\usepackage{algorithm} 
\usepackage{algpseudocode}
\providecommand{\STATE}{\State}
\providecommand{\IF}{\If}
\providecommand{\ENDIF}{\EndIf}
\providecommand{\FOR}{\For}
\providecommand{\ENDFOR}{\EndFor}
\providecommand{\COMMENT}{\Comment}

\newtheorem{definition}{Definition}

\usepackage{wrapfig}
\usepackage{xspace}
\usepackage{tikz}
\usepackage[normalem]{ulem}

\usepackage{pgfplots}
\usepackage{pgfplotstable}
\pgfplotsset{compat=1.18}
\usepackage{microtype}
\usepackage[table]{xcolor}
\usepackage{tcolorbox}
\tcbuselibrary{skins, breakable}
\usepackage{xcolor}
\definecolor{humanBg}{RGB}{235, 248, 255}  
\definecolor{modelBg}{RGB}{255, 250, 235}  
\newcommand{\treeline}{\textcolor{black!40}{\scriptsize\ttfamily{|}}} 
\newcommand{\treebranch}{\textcolor{black!40}{\scriptsize\ttfamily{+--}}}
\newcommand{\treelast}{\textcolor{black!40}{\scriptsize\ttfamily{`--}}}
\newcommand{\treegap}{\hspace{0.8em}}
\newcommand{\Tr}[1]{{\footnotesize\sffamily\bfseries\color{black} #1}\par}
\newcommand{\Tone}[1]{{\scriptsize\sffamily\bfseries\color{black!85} #1}\par}
\newcommand{\Ttwo}[1]{{\scriptsize\sffamily\color{black!75} #1}\par}

\newcommand{\Tp}[1]{{\scriptsize\sffamily\itshape\color{black!60} #1}\par}
\newcommand{\Sgl}{\textcolor{red!60!black}{\scriptsize\textbf{[Singleton]}}}

\definecolor{houseSlate}{HTML}{3A5F8A}
\definecolor{houseClay}{HTML}{CC785C}
\definecolor{houseTeal}{HTML}{63A69F}
\definecolor{houseInk}{HTML}{30343B}
\definecolor{houseMuted}{HTML}{7A8087}
\colorlet{tintDR}{houseClay!13}
\colorlet{tintBU}{houseSlate!11}
\colorlet{tintGroup}{black!5}
\colorlet{tintRef}{black!6}
\makeatletter
\@namedef{vendoricon@claude-color}{\includegraphics[height=0.92em]{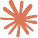}}
\@namedef{vendoricon@deepseek-color}{\includegraphics[height=0.92em]{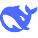}}
\@namedef{vendoricon@doubao-color}{\includegraphics[height=0.92em]{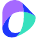}}
\@namedef{vendoricon@gemini-color}{\includegraphics[height=0.92em]{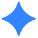}}
\@namedef{vendoricon@grok}{\includegraphics[height=0.92em]{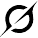}}
\@namedef{vendoricon@kimi-color}{\includegraphics[height=0.92em]{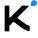}}
\@namedef{vendoricon@openai}{\includegraphics[height=0.92em]{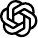}}
\@namedef{vendoricon@perplexity-color}{\includegraphics[height=0.92em]{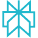}}
\@namedef{vendoricon@qwen-color}{\includegraphics[height=0.92em]{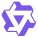}}
\newcommand{\vendoricon}[1]{\raisebox{-0.11em}{\@nameuse{vendoricon@#1}}}
\makeatother
\newcommand{\modelname}[2]{\mbox{\vendoricon{#1}\hspace{0.28em}#2}}
\newcommand{\tablegroupc}[3]{\rowcolor{#1}\multicolumn{#2}{l}{#3}}
\newcommand{\tablegroup}[2]{\tablegroupc{tintGroup}{#1}{#2}}
\newcommand{\grouplabel}[2]{\textsc{#1}\if\relax\detokenize{#2}\relax\else\hspace{0.6em}{\itshape\color{houseMuted}#2}\fi}

\usepackage[most]{tcolorbox}
\definecolor{lightblue}{RGB}{210, 220, 250}

\definecolor{medgray55}{gray}{0.55}
\definecolor{medgray}{gray}{0.7}
\definecolor{litegray}{gray}{0.9}
\definecolor{gblue}{RGB}{210, 227, 252}
\definecolor{gred}{RGB}{250, 210, 207}
\definecolor{gyellow}{RGB}{254, 239, 195}
\definecolor{ggreen}{RGB}{206, 234, 214}
\definecolor{gorange}{RGB}{254, 223, 200}

\definecolor{gblue9}{RGB}{23, 78, 166}
\definecolor{gred9}{RGB}{165, 14, 14}
\definecolor{gyellow9}{RGB}{227, 116, 0}
\definecolor{ggreen9}{RGB}{13, 101, 45}
\definecolor{gorange9}{RGB}{176, 96, 0}

\definecolor{myblue}{rgb}{0,0,1}
\definecolor{myred}{rgb}{1,0,0}
\definecolor{mylightgray}{gray}{0.95}
\definecolor{myCite}{HTML}{1C4587}

\definecolor{highlightblue}{HTML}{185ABC}
\definecolor{cellHighlight}{HTML}{dbefff}
\definecolor{lightgray}{RGB}{211, 211, 211}
\definecolor{lightfont}{gray}{0.3}

\usepackage{minitoc}

\noptcrule


\newcolumntype{L}[1]{>{\raggedright\let\newline\\\arraybackslash\hspace{0pt}}m{#1}}
\newcolumntype{C}[1]{>{\centering}m{#1}}

\newcolumntype{R}[1]{>{\raggedleft\let\newline\\\arraybackslash\hspace{0pt}}m{#1}}

\definecolor{ao}{rgb}{0.0, 0.0, 1.0}

\newcommand\vcent[1]{\vcenter{\hbox{#1}}}
\newcommand\loudspeaker[1][3]{\ensuremath{\vcent{\rule{.6ex}{.6ex}}\kern-.5ex
  \vcent{\scalebox{.6}[1]{\rotatebox[origin=center]{90}{$\blacktriangle$}}}
  \ifnum#1>0\relax\kern.05ex\vcent{\scalebox{.4}{\ttfamily)}}
  \ifnum#1>1\relax\kern-.4ex\vcent{\scalebox{.56}{\ttfamily)}}
  \ifnum#1>2\relax\kern-.55ex\vcent{\scalebox{.7}{\ttfamily)}}
  \fi\fi\fi}
}

\newcommand{\github}{\raisebox{-0.5pt}{\faGithub}\xspace}
\newcommand{\huggingface}{\raisebox{-1.5pt}{\includegraphics[height=1.05em]{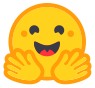}}\xspace}

\makeatletter
\renewcommand\subparagraph{
 \@startsection {subparagraph}{5}{\z@ }{3.25ex \@plus 1ex
 \@minus .2ex}{-1em}{\normalfont \normalsize \bfseries }}
\makeatother

\let\cite\citep
\hypersetup{
  citecolor = myCite,  
  linkcolor = myCite,   
  urlcolor  = myCite
}
\newcommand{\myheaderbreak}{\\}
\title{TaxoBench: Can Deep Research Agents Retrieve and Organize? \myheaderbreak Evaluating the Synthesis Gap with Expert Taxonomies}

\author{
  \textbf{Ming Zhang}\textsuperscript{\rm 1,2\dag*},
  \textbf{Jiabao Zhuang}\textsuperscript{\rm 1*},
  \textbf{Wenqing Jing}\textsuperscript{\rm 1*},
  \textbf{Kexin Tan}\textsuperscript{\rm 1*},\\
  \textbf{Ziyu Kong}\textsuperscript{\rm 1},
  \textbf{Jingyi Deng}\textsuperscript{\rm 1},
  \textbf{Yujiong Shen}\textsuperscript{\rm 1,2},
  \textbf{Yuhui Wang}\textsuperscript{\rm 1},
  \textbf{Zhenghao Xiang}\textsuperscript{\rm 1,2},\\
  \textbf{Qiyuan Peng}\textsuperscript{\rm 1},
  \textbf{Mingqi Wu}\textsuperscript{\rm 1},
  \textbf{Long Ma}\textsuperscript{\rm 1},
  \textbf{Zhangyue Yin}\textsuperscript{\rm 1,2},
  \textbf{Shihan Dou}\textsuperscript{\rm 1,2},\\
  \textbf{Maxm Pan}\textsuperscript{\rm 2\dag},
  \textbf{Tao Gui}\textsuperscript{\rm 1},
  \textbf{Qi Zhang}\textsuperscript{\rm 1\dag},
  \textbf{Xuanjing Huang}\textsuperscript{\rm 1} \\
  \vspace{0.3cm} 
  \normalsize 
  \textsuperscript{1}Fudan University 
  \textsuperscript{2}Hunyuan Team, Tencent \\ 
  \texttt{\normalsize mingzhang23@m.fudan.edu.cn, maxmpan@tencent.com, qz@fudan.edu.cn} \\
  \small
  \href{https://github.com/KongLongGeFDU/TaxoBench}{\github\,\texttt{KongLongGeFDU/TaxoBench}}
  \hspace{1.2em}
  \href{https://huggingface.co/datasets/konglongge/TaxoBench}{\huggingface\,\texttt{konglongge/TaxoBench}}
}
\correspondingauthor{Ming Zhang, Maxm Pan, and Qi Zhang. \textsuperscript{*}Equal contribution.}

\vspace{-50pt}

\vspace{-50pt}
\begin{abstract}
Deep Research Agents increasingly automate survey writing, yet existing benchmarks do not jointly test whether they retrieve the papers experts consider essential and organize those papers into paper-grounded taxonomies. We introduce \textsc{TaxoBench}, a benchmark built from 72 highly cited LLM surveys, 3{,}815 cited papers, and their expert-authored taxonomies. \textsc{TaxoBench} evaluates systems in two settings: \textbf{Deep Research} mode measures end-to-end retrieval and organization from a topic, while \textbf{Bottom-Up} mode provides the expert paper set and isolates organization. We evaluate leaf-level assignments with ARI and V-Measure and hierarchy-level structure with \textsc{US-TED}, \textsc{US-NTED}, and \textsc{Sem-Path}. Across 7 Deep Research Agents and 16 LLM configurations, the best agent retrieves only 20.92\% of expert-cited papers, and none of 70 standard Bottom-Up runs reaches the experts' average depth of 4.86. A controlled probe shows that models which match this depth do so by fragmenting the taxonomy, reducing alignment with the expert reference. We further find that raw \textsc{Sem-Path} remains near a no-organization floor even when a newer model generation gains 3.68\,pp ARI; after depth matching, humans lead on all 10 matched surveys by 13.27\,pp. These results identify retrieval and hierarchical organization as separate bottlenecks and show why hierarchy metrics must be calibrated before they are used to compare models.
\end{abstract}

\begin{document}
\doparttoc
\faketableofcontents

\vspace{-30pt}
\maketitle
\renewcommand{\myheaderbreak}{ } 

\vspace{-15pt}

\section{Introduction}

\begin{wrapfigure}[13]{r}{0.5\textwidth}
    \centering
    \vspace{-15pt}
    \includegraphics[width=0.6\linewidth]{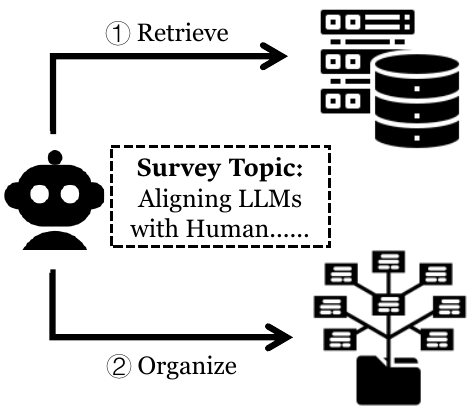}
    \vspace{1pt}
    \caption{Deep Research Agents for survey generation: given a topic, retrieve papers, then organize them into a taxonomy.}
    \label{fig:agent_process}
\end{wrapfigure}
Deep Research (DR) Agents increasingly automate survey writing by searching the web, collecting papers, and producing structured overviews~\citep{schmidgall2025agent,zhang2025web,wang2024autosurvey,yan2025surveyforge}, as sketched in Figure~\ref{fig:agent_process}. A recent DR survey~\citep{shi2025deep} identifies four sub-components: Query Planning, Information Acquisition, Memory Management, and Answer Generation. Concurrent benchmarks evaluate report-quality and rubric-following aspects of Answer Generation~\citep{zhang2025far,sharma2025researchrubrics}. Whether these systems match human experts at two foundational abilities at the intersection of \emph{Information Acquisition} and \emph{structural Answer Generation} remains unclear: retrieving essential papers and organizing them into expert-like taxonomies. Expert surveys do not arise from text generation alone; experts read broadly, identify a set of core works, and iteratively synthesize a hierarchical taxonomy that becomes the backbone of the survey. Current benchmarks focus on writing quality, factuality, or citation correctness~\citep{eldifrawi2024automated,wadden2020fact}, rather than on whether an agent retrieves the papers experts would cite and organizes them in an expert-like structure.

\begin{figure*}[t]
 \centering
 \includegraphics[width=\textwidth]{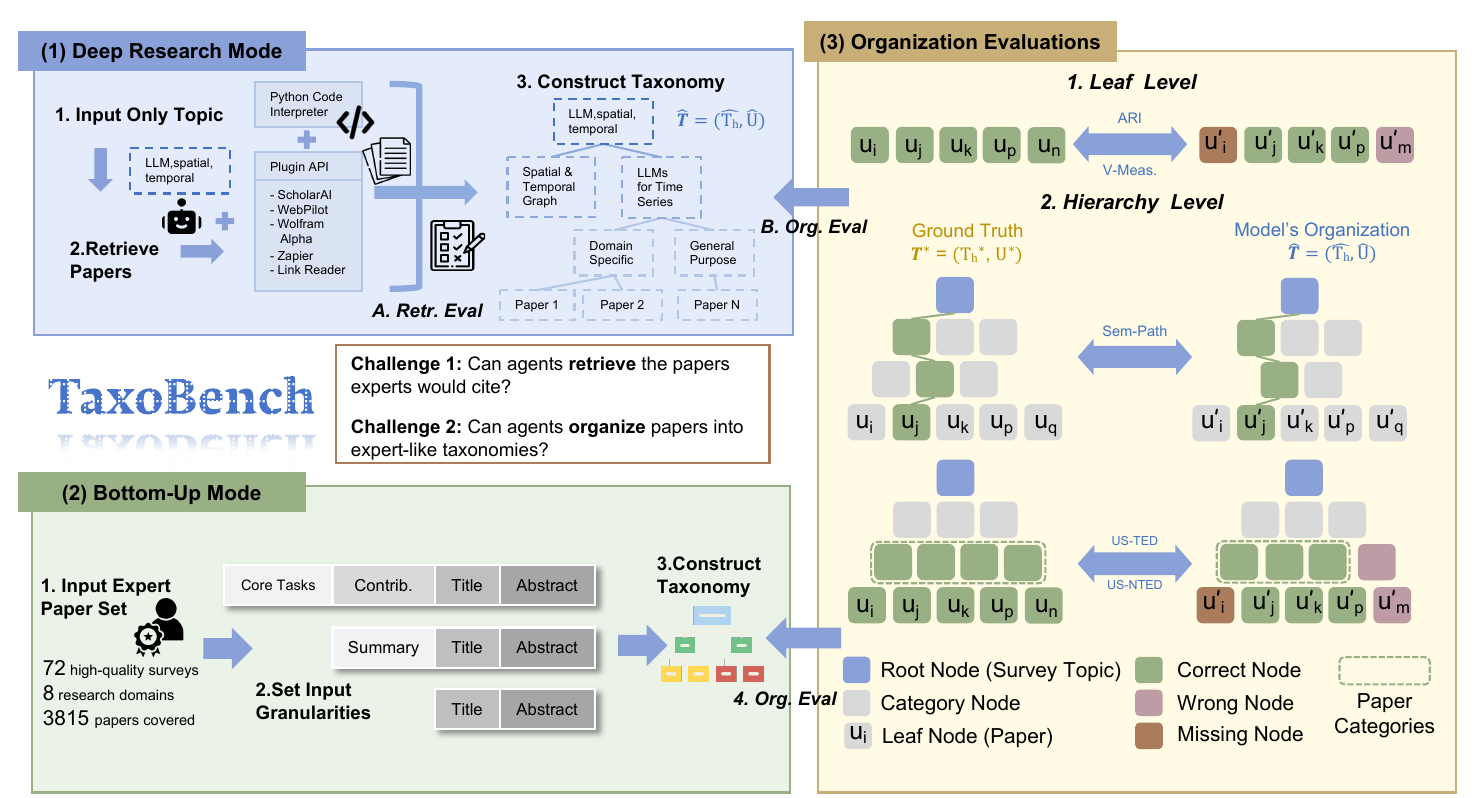}
 \caption{Overview of \textsc{TaxoBench}. In \textbf{Deep Research mode}, agents receive only a topic, retrieve papers with web tools, and build a taxonomy. In \textbf{Bottom-Up mode}, we provide the expert paper set to isolate organization. We compare the output with the expert reference at the leaf level using ARI and V-Measure and at the hierarchy level using \textsc{US-TED}/\textsc{US-NTED} and \textsc{Sem-Path}.}
 \label{fig:main_fig}
\end{figure*}

We identify two challenges. \textbf{C1: Retrieval.} Existing benchmarks do not provide expert-curated paper sets for survey-oriented retrieval at scale, although standard Recall, Precision, and F1 suffice once such ground truth exists. \textbf{C2: Organization.} Standard clustering metrics such as ARI and V-Measure capture only partitions. They cannot distinguish confusion between sibling categories from errors that move content across distant branches. We therefore evaluate both the \emph{leaf level} of paper-to-category assignment and the \emph{hierarchy level} of taxonomy structure, including node semantics and parent--child relations.

We address these challenges with \textsc{TaxoBench}, which contains 72 highly cited LLM surveys, their expert-authored taxonomies, and 3{,}815 cited papers manually mapped to expert categories. Figure~\ref{fig:main_fig} shows our two evaluation modes. \textbf{Deep Research mode} tests end-to-end performance from a topic, while \textbf{Bottom-Up mode} provides the expert paper set and isolates organization. We propose three hierarchy-aware metrics, \textsc{US-TED}, \textsc{US-NTED}, and \textsc{Sem-Path}, to measure structure beyond flat clustering. We also separate \textbf{alignment-based} findings, which measure agreement with the expert reference, from \textbf{capability-based} findings, which use retrieval Recall and judge-free tree statistics (Section~\ref{subsec:alignment_vs_capability}).

We evaluate 7 Deep Research Agents and 16 frontier LLM configurations under this distinction. On the capability side, the best agent retrieves only 20.92\% of expert-cited papers, and the deepest of the 70 Bottom-Up runs is still 0.89 levels short of the expert depth of 4.86; models pushed to that depth fragment their taxonomies instead. On the alignment side, every model given the exact expert paper set scores 28--30\% \textsc{Sem-Path}, just above the 27.49\% no-organization floor. Depth matching restores a 9.45\,pp spread, and most of the remaining gap to the expert comes from category vocabulary rather than paper placement.

The contributions are threefold:
\begin{enumerate}[topsep=0pt,parsep=0pt,itemsep=2pt,leftmargin=*]
    \item We introduce \textsc{TaxoBench}, which pairs 72 expert-authored survey taxonomies with 3{,}815 cited papers and supports separate evaluation of retrieval and organization (Section~\ref{sec:dataset}).
    \item We develop three hierarchy-aware metrics, \textsc{US-TED}, \textsc{US-NTED}, and \textsc{Sem-Path}, and separate alignment-based from capability-based findings. Calibrating \textsc{Sem-Path} reveals that it is nearly flat for shallow trees; after depth matching, a 13.27\,pp paired human advantage remains on all 10 matched surveys (Apps.~\ref{app:metric_calibration} and~\ref{app:human_baseline}).
    \item We evaluate 7 Deep Research Agents and 16 frontier LLM configurations. The best retrieval Recall is 20.92\%, and a depth deficit persists across 70 runs and two model generations. A controlled probe shows that forcing depth fragments taxonomies rather than improving their organization (App.~\ref{app:depth_probe}).
\end{enumerate}

\section{\textsc{TaxoBench}}
\label{sec:dataset}
\begin{figure}[!ht]
\centering
\begin{minipage}[b]{0.46\linewidth}
\centering
\small
\setlength{\tabcolsep}{4pt}
\begin{tabular}{lc}
\toprule
Statistic & Value \\
\midrule
\tablegroup{2}{\grouplabel{Survey collection}{}} \\
Surveys / total papers & 72 / 3{,}815 \\
Avg.\ citations per survey & 354.5 \\
\tablegroup{2}{\grouplabel{Per taxonomy}{mean $\pm$ std}} \\
Papers per taxonomy & 53.0 $\pm$ 20.6 \\
Hierarchy depth & 4.9 $\pm$ 0.8 \\
Paper categories & 14.0 $\pm$ 6.5 \\
Papers per category & 3.8 $\pm$ 3.1 \\
Branching factor & 2.5 $\pm$ 0.5 \\
\bottomrule
\end{tabular}
\captionof{table}{\textbf{\textsc{TaxoBench} dataset statistics.}}
\label{tab:stats}
\end{minipage}\hfill
\begin{minipage}[b]{0.50\linewidth}
\centering
\includegraphics[width=\linewidth]{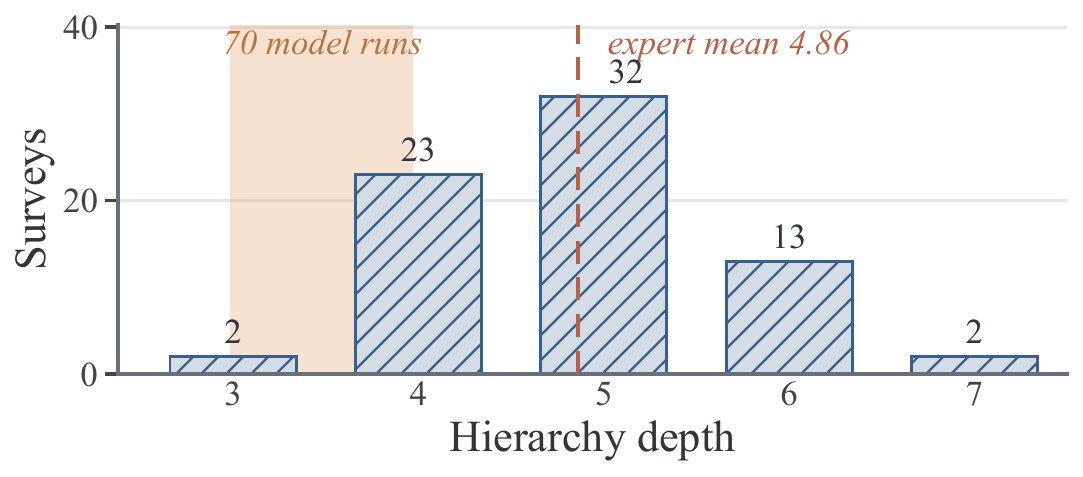}
\caption{\textbf{Expert taxonomy depths.} Only 2 of 72 expert taxonomies fall inside the shaded band of Bottom-Up run depths.}
\label{fig:expert_depth}
\end{minipage}
\end{figure}

We define a \emph{taxonomy} as a \emph{category hierarchy} over topics together with a \emph{paper-to-category assignment} that maps each paper to a leaf category. We extract these taxonomies from published expert surveys and use them as references for evaluating agents. Section~\ref{sec:metrics} gives the formal definitions, Table~\ref{tab:stats} summarizes the dataset, and Figure~\ref{fig:expert_depth} shows how reference depth is distributed over the 72 surveys.

\paragraph{Survey collection.}
We begin with more than 200 candidate LLM-related surveys and apply four filters. We retain surveys with an explicit taxonomy figure, then filter by citation count; the retained surveys average 354.5 citations. We exclude surveys whose scope is too broad for a coherent taxonomy, and Ph.D.-level researchers verify the remaining set. The final 72 surveys span 8 AI/ML subdomains. App.~\ref{app:survey_selection_pipeline} reports the count after each stage. We also record each survey's publication date $t_s$ as the retrieval cutoff described in Section~\ref{subsec:retrieval}.

\paragraph{Taxonomy extraction.}
Our Ph.D.-level annotators transcribe each published taxonomy figure rather than redesigning it. They then read the cited papers and map them to categories. We disambiguate and deduplicate papers and check each placement against the survey text (App.~\ref{app:dataset_details}). In a separate baseline, human annotators build taxonomies from scratch under the same Bottom-Up requirements (App.~\ref{app:human_baseline}).

\paragraph{Multi-category papers.}
Some papers belong to several categories. Because ARI and V-Measure require one assignment, we retain the most contextually appropriate category; removing all multi-category papers changes \textsc{Sem-Path} by less than 1\,pp (App.~\ref{app:multi_category}).

\section{Evaluation Framework}
\label{sec:metrics}

\subsection{Preliminaries and Notation}
\label{subsec:prelim}

\begin{definition}[Taxonomy]
A \textbf{taxonomy} $T=(T_h,U)$ consists of (i)~a \textbf{category hierarchy} $T_h=(\mathcal{C},E)$, a rooted tree over category labels $\mathcal{C}$ with parent--child edges $E$; and (ii)~a \textbf{paper-to-category assignment} $U:\mathcal{P}\to\mathcal{C}_p$ sending each paper to a \textbf{paper category} $c_p\in\mathcal{C}_p\subseteq\mathcal{C}$, the leaves of $T_h$ under which papers are attached.
\end{definition}

\paragraph{Notation and modes.}
We write $T^*=(T_h^*,U^*)$ for the expert taxonomy and $\hat{T}=(\hat{T}_h,\hat{U})$ for a model output. We denote the expert-cited and model-retrieved paper sets by $\mathcal{P}^*$ and $\hat{\mathcal{P}}$, the corresponding paper-category sets by $\mathcal{C}_p^*$ and $\hat{\mathcal{C}}_p$, the subtree size at $u$ by $|T_u|$, and its children by $\mathrm{Ch}(u)$. \textbf{Deep Research} evaluates retrieval and organization end to end; \textbf{Bottom-Up} sets $\mathcal{P}=\mathcal{P}^*$ and asks models to organize a fixed paper set.

\subsection{Retrieval Evaluation}
\label{subsec:retrieval}
We evaluate retrieval only in Deep Research mode, where the model must recover $\mathcal{P}^*$ from a topic query, with set-based metrics:
\begin{equation}
\begin{gathered}
\textsc{Recall}=\frac{|\mathcal{P}^*\cap\hat{\mathcal{P}}|}{|\mathcal{P}^*|},\enspace
\textsc{Precision}=\frac{|\mathcal{P}^*\cap\hat{\mathcal{P}}|}{|\hat{\mathcal{P}}|},\\
\textsc{F1}=\frac{2\cdot\textsc{Precision}\cdot\textsc{Recall}}{\textsc{Precision}+\textsc{Recall}}.
\end{gathered}
\end{equation}
Recall is micro-averaged over pooled hits and Precision macro-averaged; F1 is their harmonic mean (App.~\ref{app:aggregation}).

\paragraph{Temporal cutoff specification.}
We use each survey's publication date $t_s$ as the retrieval cutoff: $\mathcal{P}^*_s$ contains only papers published before $t_s$, and Precision and F1 use the full output (App.~\ref{app:temporal_filtering}).

\paragraph{Source surveys.}
Agents retrieve the source survey itself in 119 of 504 runs. A survey is not in its own expert paper set, so these cases do not change Recall (App.~\ref{app:leakage_audit}).

\subsection{Organization Evaluation}
\label{subsec:organization}
We evaluate organization at the \textbf{Leaf Level}, using paper-to-category assignments, and at the \textbf{Hierarchy Level}, using taxonomy structure.

\subsubsection{Leaf-Level: Paper-to-Category Assignment}
\label{subsubsec:leaf_level}
Given a paper universe $\mathcal{P}$, we compare the expert assignment $U^*:\mathcal{P}\to\mathcal{C}_p^*$ with the model assignment $\hat{U}:\mathcal{P}\to\hat{\mathcal{C}}_p$. Since ARI and V-Measure evaluate partitions without encoding hierarchical structure, we use them here and introduce hierarchy-level metrics in Section~\ref{subsubsec:hier_level}.

\paragraph{Leaf-level metrics.}
We use the chance-corrected \textbf{Adjusted Rand Index} (\textsc{ARI})~\citep{hubert1985comparing} and \textbf{V-Measure}~\citep{rosenberg2007v}; the latter reports homogeneity (\textsc{Hom}.\,$=1-H(U^*|\hat U)/H(U^*)$) and completeness (\textsc{Comp}.\,$=1-H(\hat U|U^*)/H(\hat U)$) and combines them as $\textsc{V-Measure}=\frac{2\,\textsc{Hom}.\,\textsc{Comp}.}{\textsc{Hom}.+\textsc{Comp}.}$. Full forms are in App.~\ref{sec:ari_details}.

\paragraph{Deep Research mode.}
Retrieval limits what an agent can organize, so we report two views. Our \textbf{retrieval-conditioned} view restricts $\mathcal{P}$ to $\mathcal{P}^*\cap\hat{\mathcal{P}}$ and gives \textsc{ARI}$_\cap$ and \textsc{V-Measure}$_\cap$. Our \textbf{end-to-end} view sets $\mathcal{P}=\mathcal{P}^*$ and labels unretrieved papers $\perp$, so it includes the retrieval bottleneck.

\subsubsection{Hierarchy-Level: Taxonomy Hierarchy Structure}
\label{subsubsec:hier_level}
Flat partition metrics such as ARI and V-Measure cannot distinguish sibling confusion from errors that move content across distant branches. Soft-cardinality metrics such as \textsc{NSR}/\textsc{NSP}~\citep{franti2023soft} are \emph{structure-blind} for the same reason, and can be perfect even when subtrees are entirely rewired (App.~\ref{sec:metric_nsr_nsp_discussion}). We therefore compare the hierarchies $T_h^*$ and $\hat{T}_h$ directly, keeping only internal categories and edges and excluding paper nodes, with two complementary metrics: \textsc{US-(N)TED} gives a \emph{global} edit distance under unordered sibling matching, and \textsc{Sem-Path} a \emph{local}, per-paper measure of root-to-leaf path consistency.

\noindent\textbf{Unordered Semantic Tree Edit Distance (\textsc{US-TED}).}
We extend the STED framework~\citep{wang2025sted} to the unordered siblings of a taxonomy, so sibling permutation does not affect \textsc{US-TED}. Using all-MiniLM-L6-v2 embeddings $\mathbf{e}(\cdot)$~\citep{all-MiniLM-L6-v2}, we define semantic label similarity and renaming cost as
\begin{equation}
\begin{aligned}
\mathrm{Sim}(x,y)&\coloneqq\max\!\bigl(0,\cos(\mathbf{e}(x),\mathbf{e}(y))\bigr),\\
\mathrm{cost}_{\mathrm{ren}}(x\!\to\!y)&\coloneqq 1-\mathrm{Sim}(x,y)\in[0,1],
\end{aligned}
\label{eq:sim}
\end{equation}
with insertion/deletion cost $1$ per node. Then $D(u,v)\coloneqq\mathrm{cost}_{\mathrm{ren}}(u\!\to\!v)+\mathrm{MatchCost}(\mathrm{Ch}(u),\mathrm{Ch}(v))$, where $\mathrm{MatchCost}$ is a minimum-cost bipartite matching (Hungarian) and unmatched children correspond to deleting or inserting whole subtrees, charged by subtree size. The tree-level distance is $\textsc{US-TED}(T_h^*,\hat{T}_h)\coloneqq D(r^*,\hat{r})$ at the roots, which we normalize by hierarchy size for cross-instance comparability: $\textsc{US-NTED}\coloneqq\textsc{US-TED}(T_h^*,\hat{T}_h)/(|T_h^*|+|\hat{T}_h|)$. Lower values indicate closer structural alignment; the full assignment formulation, boundedness, and properties are in App.~\ref{sec:metric_us_ted}. We report \textsc{US-NTED} as a percentage in all tables.

\noindent\textbf{Semantic Path Similarity (\textsc{Sem-Path}).}
\textsc{US-(N)TED} captures global tree divergence but does not test the ancestor chain of each paper. We therefore define \textsc{Sem-Path}. Let $D_a\subseteq\mathcal{P}^*\times\hat{\mathcal{P}}$ be the aligned paper pairs described in App.~\ref{sec:metric_paper_alignment}. In Bottom-Up mode, $D_a$ covers $\mathcal{P}^*$; in Deep Research mode, it covers $\mathcal{P}^*\cap\hat{\mathcal{P}}$. For each pair $(d,\hat{d})\in D_a$, we remove the paper node from the root-to-paper paths in $T^*$ and $\hat{T}$ to obtain ancestor-label sequences $S=(s_1,\ldots,s_m)$ and $\hat{S}=(\hat{s}_1,\ldots,\hat{s}_n)$. With clipped cosine distance $\delta(x,y)\!\coloneqq\!1{-}\mathrm{Sim}(x,y)$ and $m\le n$ without loss of generality, we take the order-preserving minimum-cost alignment over strictly increasing $f:\{1,\ldots,m\}\!\to\!\{1,\ldots,n\}$:
\begin{equation}
J(S,\hat{S})\coloneqq\min_{f}\sum_{i=1}^{m}\delta(s_i,\hat{s}_{f(i)})\;+\;\lambda\,(n-m),
\end{equation}
with $\lambda\!=\!1$, which charges an unmatched ancestor exactly as much as the maximum distance between two matched labels; a sensitivity analysis over $\lambda\in\{0.5,1,1.5,2\}$ is in App.~\ref{app:lambda_sensitivity}. Letting $J_d$ be the minimum cost for $(d,\hat d)$ over its candidate ancestor-path pairs,
\begin{equation}
\textsc{Sem-Path}\coloneqq\frac{1}{|D_a|}\sum_{(d,\hat{d})\in D_a}\frac{1}{1+J_d}.
\end{equation}
Unlike \textsc{ARI}, \textsc{Sem-Path} has no chance correction, so raw values need the floors and the depth-matched variant of App.~\ref{app:metric_calibration}.

\subsection{LLM-as-Judge and Diagnostic Split}
\label{subsec:llm_judge}
\label{subsec:alignment_vs_capability}
\paragraph{LLM-as-Judge.} We also prompt GPT-4o to compare $T_h^*$ and $\hat{T}_h$ on four 1--5 scales: \textbf{Coverage}, \textbf{Organization} or MECE, \textbf{Logic} or parent--child consistency, and \textbf{Topology}. Agreement with human evaluators reaches Cohen's $\kappa=0.89$. Details are in Apps.~\ref{app:judge_co_with_human} and~\ref{sec:prompt_judge}.

\paragraph{Alignment vs.\ capability.} \textbf{L1} metrics (\textsc{ARI}, \textsc{V-Measure}, \textsc{US-(N)TED}, \textsc{Sem-Path}) measure agreement with the full expert hierarchy $T_h^*$. \textbf{L2} comparisons use only scalar statistics of $T_h^*$, such as depth and category count, and \textbf{L3} retrieval Recall uses $\mathcal{P}^*$ but no taxonomy.

\section{Benchmarking LLMs on \textsc{TaxoBench}}
\label{sec:findings}
We address three research questions. In \textbf{Deep Research mode}, we evaluate 7 agents end to end from a topic, each in its product interface: o3, Grok, Gemini, Perplexity, DeepSeek, Qwen, and Doubao. In \textbf{Bottom-Up mode}, we provide $\mathcal{P}^*$ to the newest reachable model from each of seven families: Claude-Sonnet-5, GPT-5.6-sol, Gemini-3.1-Pro, DeepSeek-V4-Pro, Qwen3.8-Max, Kimi-K3, and Grok-4. We run each model at its lowest and highest reasoning effort and add a second GPT tier as a within-family control, for 16 configurations in total (App.~\ref{subsec:reasoning_control}); the 12 configurations used to build the benchmark are in App.~\ref{app:prevgen}. Following~\citet{zhang2026opennovelty}, we test \textbf{Title+Abstract}, \textbf{+Summary}, and \textbf{+Core-task\&Contributions} inputs. Claude-Sonnet-4.6 produces the auxiliary text; hand-written inputs shift \textsc{Sem-Path} by at most 0.75\,pp. Apps.~\ref{app:model_list},~\ref{subsec:bottomup_results_input_forms}, and~\ref{app:human_input_control} provide details.

\subsection{\texorpdfstring{RQ\,I}{RQ I}: Can current agents replicate expert-level literature discovery?}

\begin{table}[t]
\centering
\small
\setlength{\tabcolsep}{5pt}
\begin{tabular}{lccc}
\toprule
Deep Research agent & Recall$\uparrow$ & Precision$\uparrow$ & F1$\uparrow$ \\
\midrule
\modelname{openai}{o3} & \textbf{20.92} & 29.29 & \textbf{24.41} \\
\modelname{grok}{Grok} & 12.82 & \textbf{29.35} & 17.85 \\
\modelname{gemini-color}{Gemini} & 15.23 & 18.92 & 16.88 \\
\modelname{perplexity-color}{Perplexity} & 6.61 & 7.47 & 7.01 \\
\modelname{deepseek-color}{DeepSeek} & 4.61 & 14.26 & 6.97 \\
\modelname{qwen-color}{Qwen} & 4.35 & 7.83 & 5.59 \\
\modelname{doubao-color}{Doubao} & 3.15 & 3.83 & 3.46 \\
\bottomrule
\end{tabular}
\caption{\textbf{Retrieval capability} of Deep Research Agents when given the same survey topic as human experts. All values are percentages; best results are \textbf{bold}; 3-run Recall std $\le 3.01$\,pp (App.~\ref{app:error_bars}).}
\label{tab:retrieval_capability_dpa}
\end{table}

\textbf{Finding 1: Deep Research Agents exhibit severe retrieval bottlenecks.}
We find that current agents do not retrieve core literature at scale. The best agent, o3, reaches 20.92\% Recall and 24.41\% F1, so it misses nearly 80\% of expert-cited papers. Four of the seven agents fall below 10\% F1. Precision is also limited: Grok leads at 29.35\%, indicating that peripheral papers dominate the retrieved sets. Any pipeline built on these agents inherits a Recall bound of at most 21\%. A re-test with three current agents given a neutral topic description (App.~\ref{app:dr_retest}) reaches at most 19.74\% F1 and 25.14\% Recall; these agents raise Recall mainly by listing more papers, which lowers Precision.

\subsection{\texorpdfstring{RQ\,II}{RQ II}: How well can agents organize the retrieved subset?}

\begin{figure*}[t]
    \centering
    \begin{subfigure}[b]{0.49\linewidth}
        \centering
        \includegraphics[width=\linewidth]{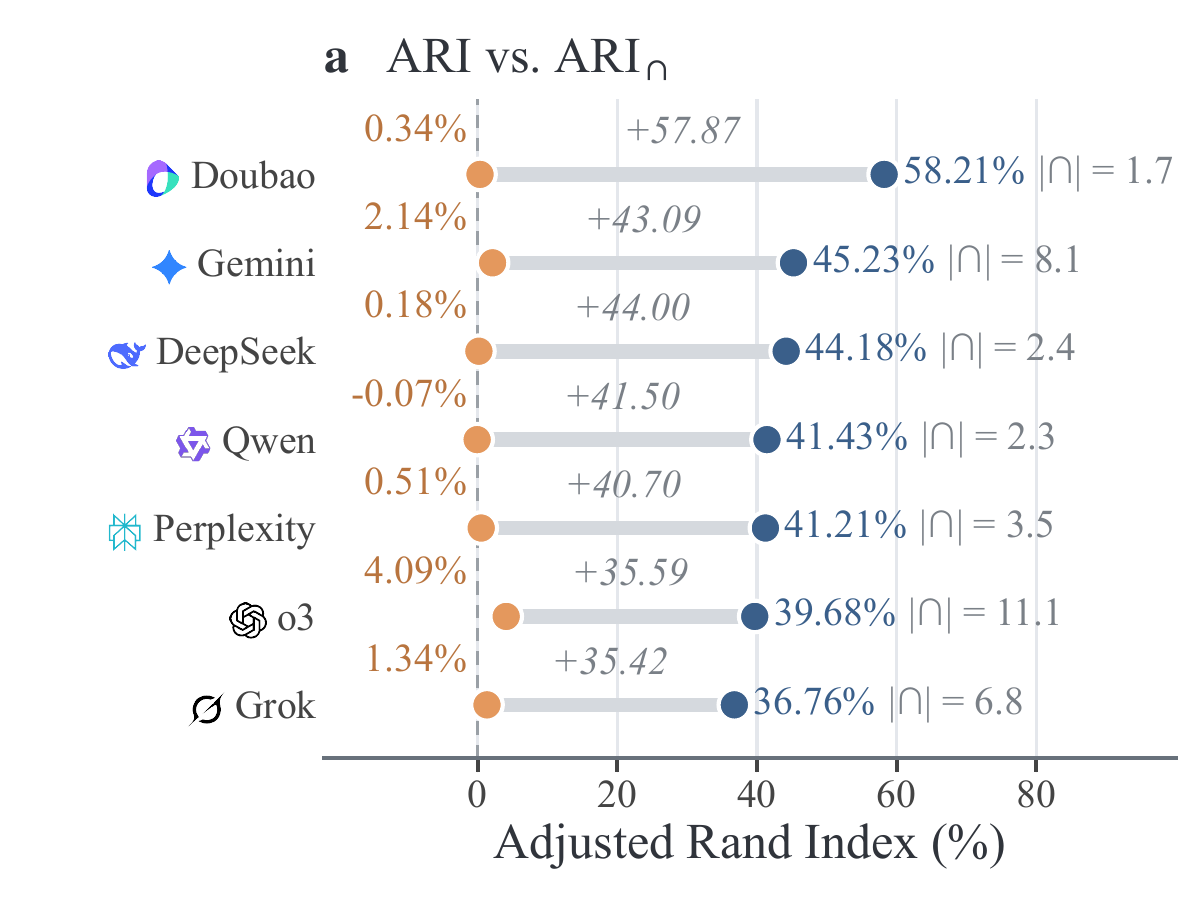}
        \phantomsubcaption
        \label{fig:ari_comparison}
    \end{subfigure}\hfill
    \begin{subfigure}[b]{0.49\linewidth}
        \centering
        \includegraphics[width=\linewidth]{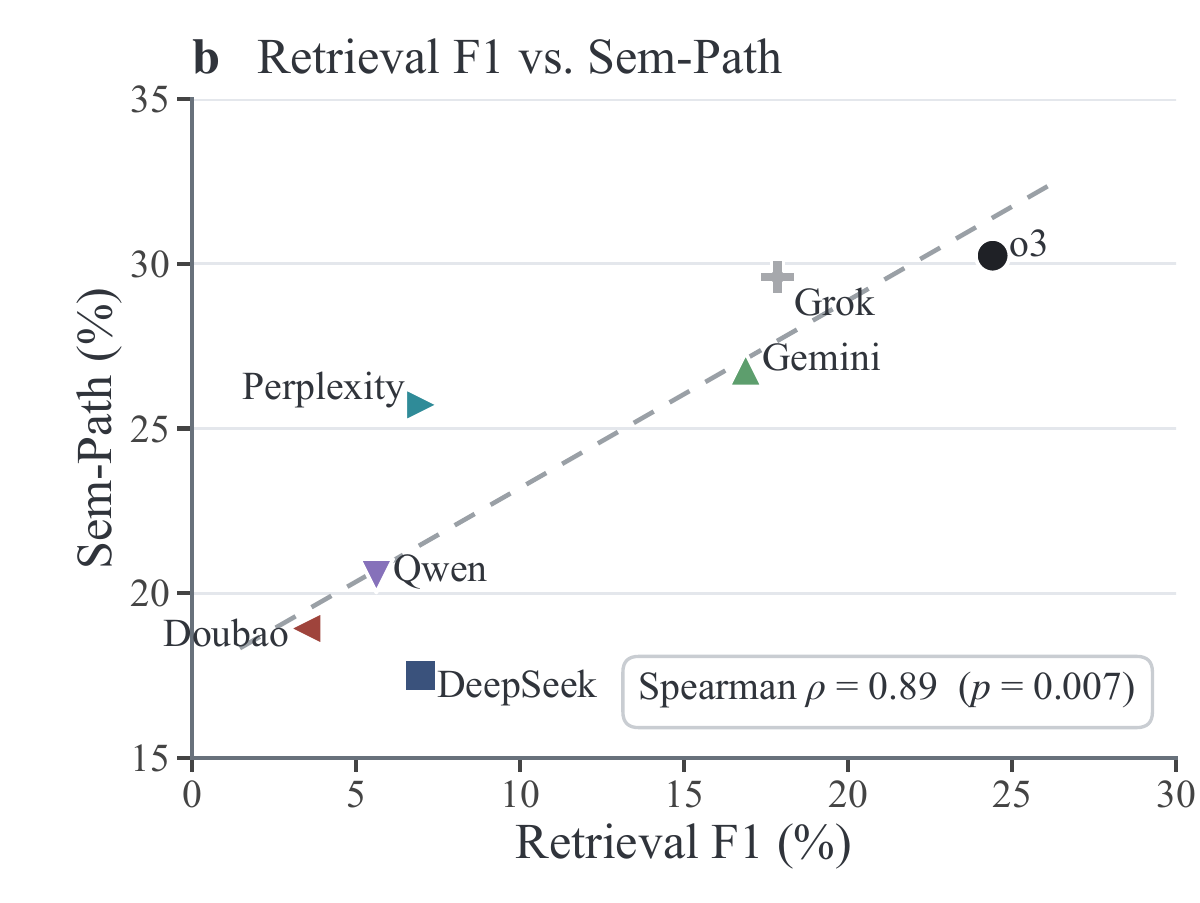}
        \phantomsubcaption
        \label{fig:retrieval_hierarchy}
    \end{subfigure}
    \caption{Retrieval is the dominant bottleneck in Deep Research mode. (a) End-to-end ARI vs.\ ARI$_\cap$ on retrieved expert papers, with intersection size shown (App.~\ref{app:ari_intersection}). (b) F1 vs.\ \textsc{Sem-Path} across the 7 agents ($\rho=0.89$, $p=0.007$).}
    \label{fig:dr_bottleneck}
\end{figure*}

\textbf{Finding 2: Conditioning on retrieved papers raises every score, but intersection size confounds the gain.}
When we restrict evaluation to $\mathcal{P}^*\cap\hat{\mathcal{P}}$, ARI rises from below 5\% to 37--58\% for all seven agents; Table~\ref{tab:org_deep_research_organization} and Figure~\ref{fig:dr_bottleneck}a report the result. Agents can thus separate core papers once they find them, but the ranking reverses with retrieval quality: Doubao scores 58.21\% on intersections averaging only 1.7 papers, o3 39.68\% on 11.1 papers. Small intersections make clustering easier, so ARI$_\cap$ mainly shows how much retrieval limits the end-to-end score (App.~\ref{app:ari_intersection}).

\textbf{Finding 3: Retrieval quality correlates with hierarchy-level structure.}
We observe a strong association between F1 and \textsc{Sem-Path} across the 7 agents, with Spearman $\rho=0.89$ and $p=0.007$ in Figure~\ref{fig:dr_bottleneck}b. o3 leads on both measures, while Doubao and Qwen score low on both. Off-topic papers plausibly pull agents toward a different hierarchy.

\begin{table*}[t]
\centering
\small
\setlength{\tabcolsep}{4pt}
\begin{tabular}{lcccccccc}
\toprule
\multirow{2}{*}{Deep Research agent} & \multicolumn{4}{c}{Leaf level} & \multicolumn{3}{c}{Hierarchy level} \\
\cmidrule(lr){2-5} \cmidrule(lr){6-8}
& ARI$\uparrow$ & V-Meas.$\uparrow$ & $|\mathcal{P}^*\!\cap\!\hat{\mathcal{P}}|$ & ARI$_\cap$$\uparrow$ / V-Meas.$_\cap$$\uparrow$ & \textsc{US-TED}$\downarrow$ & \textsc{US-NTED}$\downarrow$ & \textsc{Sem-Path}$\uparrow$ \\
\midrule
\modelname{openai}{o3} & \textbf{4.09\%} & \textbf{26.05\%} & \textbf{11.1} & 39.68\% / 75.94\% & 29.25 & 74.02\% & \textbf{30.25\%} \\
\modelname{grok}{Grok} & 1.34\% & 20.03\% & 6.8 & 36.76\% / 76.80\% & \textbf{27.14} & \textbf{72.80\%} & 29.60\% \\
\modelname{gemini-color}{Gemini} & 2.14\% & 23.57\% & 8.1 & 45.23\% / \textbf{82.24\%} & 40.56 & 79.02\% & 26.79\% \\
\modelname{perplexity-color}{Perplexity} & 0.51\% & 12.79\% & 3.5 & 41.21\% / 79.13\% & 56.65 & 83.93\% & 25.72\% \\
\modelname{deepseek-color}{DeepSeek} & 0.18\% & 8.33\% & 2.4 & 44.18\% / 81.70\% & 29.97 & 74.96\% & 17.50\% \\
\modelname{qwen-color}{Qwen} & $-$0.07\% & 7.59\% & 2.3 & 41.43\% / 79.64\% & 45.09 & 76.36\% & 20.53\% \\
\modelname{doubao-color}{Doubao} & 0.34\% & 6.57\% & 1.7 & \textbf{58.21\%} / 81.28\% & 39.06 & 75.13\% & 18.93\% \\
\bottomrule
\end{tabular}
\caption{\textbf{Organization capability} in \textbf{Deep Research mode}. $|\mathcal{P}^*\!\cap\!\hat{\mathcal{P}}|$ is the mean number of retrieved expert papers; the $_\cap$ metrics average over surveys with at least two aligned papers (App.~\ref{app:ari_intersection}). Best results are \textbf{bold}; 3-run \textsc{Sem-Path} std $\le 2.14$\,pp.}
\label{tab:org_deep_research_organization}
\end{table*}

\subsection{\texorpdfstring{RQ\,III}{RQ III}: How well can LLMs organize given perfect retrieval?}

\begin{figure*}[t]
    \centering
    \begin{subfigure}[b]{0.49\linewidth}
        \centering
        \includegraphics[width=\linewidth]{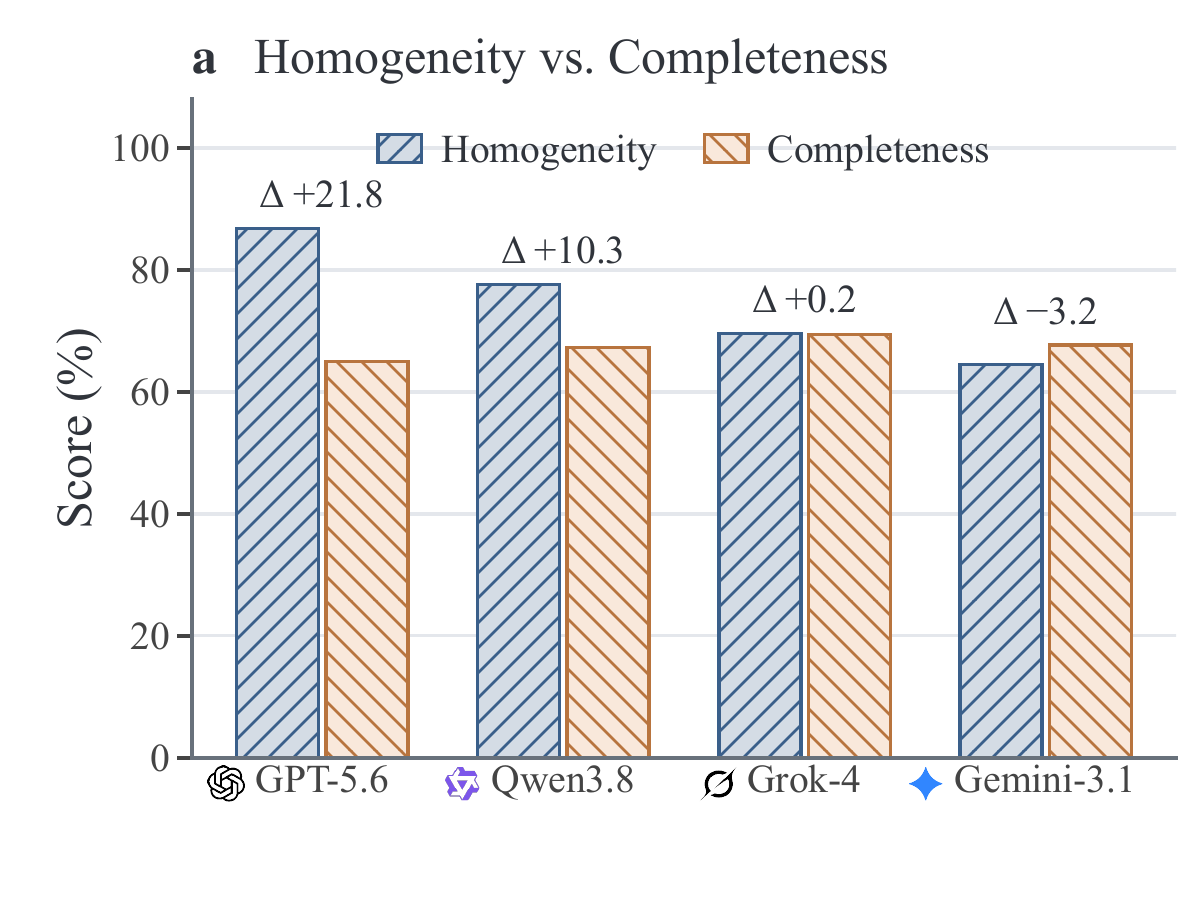}
        \phantomsubcaption
        \label{fig:hom_comp}
    \end{subfigure}\hfill
    \begin{subfigure}[b]{0.49\linewidth}
        \centering
        \includegraphics[width=\linewidth]{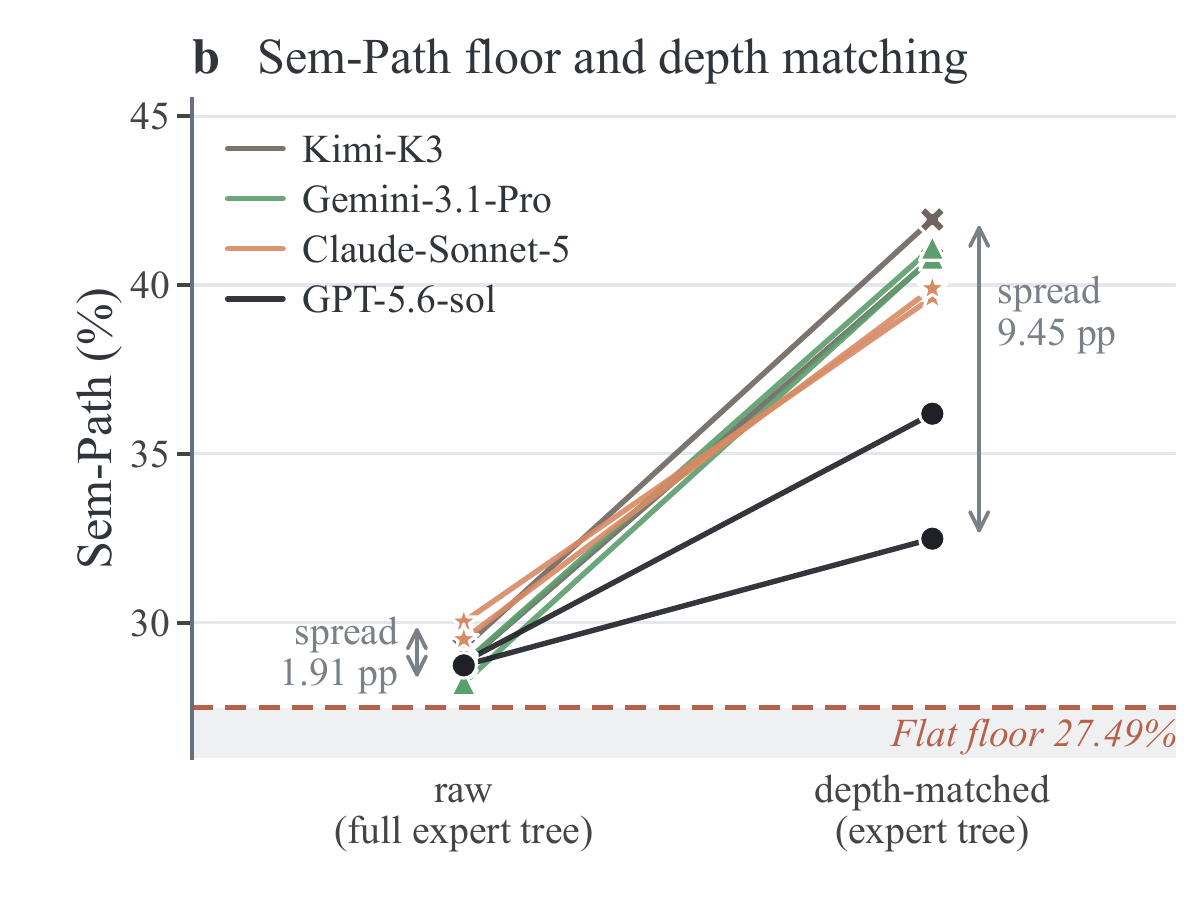}
        \phantomsubcaption
        \label{fig:sem_path_convergence}
    \end{subfigure}
    \caption{Two organization failures in Bottom-Up mode. (a) Homogeneity above Completeness indicates over-segmentation; the gap $\Delta$ (pp) is largest for GPT-5.6-sol and negative for Gemini-3.1-Pro. (b) Every raw \textsc{Sem-Path} score lies within 2.5\,pp of the 27.49\% \textsc{Flat} floor; depth matching raises scores to 32--42\% (App.~\ref{app:metric_calibration}). Panels show 4 and 8 of the 16 configurations for legibility.}
    \label{fig:bottomup_failures}
\end{figure*}

\textbf{Finding 4: Most models over-segment at the leaf level.}
Figure~\ref{fig:hom_comp} shows higher Homogeneity than Completeness for every model except Gemini-3.1-Pro. The gap is largest for GPT-5.6-sol at 86.78\% versus 65.02\%, which indicates that this model splits topics into fine-grained clusters instead of broader expert-style categories (example in App.~\ref{app:error_and_case_study}). Gemini-3.1-Pro reverses the gap at 64.54\% versus 67.71\% and has the lowest singleton rate at 5.1\%, compared with 17.0\% for the expert. On 15 surveys, constraining previous-generation GPT-5 to the expert category count raises ARI from 13.05\% to 20.54\% and lowers Homogeneity from 73.84\% to 62.87\%, confirming that over-segmentation can drive high Homogeneity (App.~\ref{app:constrained_k}). Higher reasoning effort raises ARI for 3 of the 8 models and lowers it for 5 (App.~\ref{subsec:reasoning_control}).

\begin{table*}[t]
\centering
\footnotesize
\setlength{\tabcolsep}{4.5pt}
{%
\begin{tabular}{lcccccccc}
\toprule
\multirow{2}{*}{Model} & \multicolumn{4}{c}{Leaf level} & \multicolumn{2}{c}{Hierarchy level} & \multicolumn{2}{c}{Judge-free shape} \\
\cmidrule(lr){2-5} \cmidrule(lr){6-7} \cmidrule(lr){8-9}
& ARI$\uparrow$ & Hom.$\uparrow$ & Comp.$\uparrow$ & \textsc{US-NTED}$\downarrow$ & \textsc{Sem-Path}$\uparrow$ & Depth & \#Cat & Singl. \\
\midrule
\tablegroupc{tintBU}{9}{\grouplabel{Lowest reasoning effort}{\textup{\texttt{none}} for GPT, \textup{\texttt{minimal}} elsewhere; App.~\ref{subsec:reasoning_control}}} \\
\modelname{claude-color}{Claude-Sonnet-5} & \textbf{34.92\%} & 75.62\% & \textbf{69.71\%} & \textbf{74.66\%} & \textbf{30.03\%} & 3.43 & 15.9 & 21.2\% \\
\modelname{kimi-color}{Kimi-K3} & 33.23\% & 76.42\% & 68.69\% & 78.74\% & 29.27\% & 3.06 & 16.2 & 16.1\% \\
\modelname{grok}{Grok-4} & 31.44\% & 69.60\% & 69.40\% & 79.08\% & 28.13\% & 3.01 & 12.9 & 11.3\% \\
\modelname{deepseek-color}{DeepSeek-V4-Pro} & 29.81\% & 75.59\% & 68.68\% & 78.15\% & 28.72\% & 3.18 & 18.4 & 28.3\% \\
\modelname{qwen-color}{Qwen3.8-Max} & 29.81\% & 77.63\% & 67.29\% & 78.13\% & 29.48\% & 3.28 & 18.5 & 23.2\% \\
\modelname{gemini-color}{Gemini-3.1-Pro} & 29.67\% & 64.54\% & 67.71\% & 77.79\% & 28.22\% & 2.99 & 11.0 & 5.1\% \\
\modelname{openai}{GPT-5.6-sol} & 21.98\% & \textbf{86.78\%} & 65.02\% & 77.46\% & 28.84\% & 3.56 & 28.3 & 43.5\% \\
\midrule
\rowcolor{tintRef}
\emph{Expert reference} & --- & --- & --- & --- & --- & \emph{4.86} & \emph{14.0} & \emph{17.0\%} \\
\bottomrule
\end{tabular}%
}
\caption{\textbf{Organization capability} in \textbf{Bottom-Up mode}: the newest reachable model of each family on Title+Abstract inputs for all 72 surveys, at the lowest reasoning effort. Highest-effort rows, a second GPT tier, the previous generation, and 3-run standard deviations are in Apps.~\ref{subsec:reasoning_control}, \ref{app:prevgen}, and~\ref{app:error_bars}. Best results are \textbf{bold}. Every model is shallower than the expert depth of 4.86, and \textsc{Sem-Path} lies only 0.6--2.5\,pp above the 27.49\% \textsc{Flat} floor.}
\label{tab:org_bottom_up_mode}
\end{table*}

\begin{figure*}[t]
    \centering
    \begin{subfigure}[t]{0.49\linewidth}
        \centering
        \includegraphics[width=\linewidth]{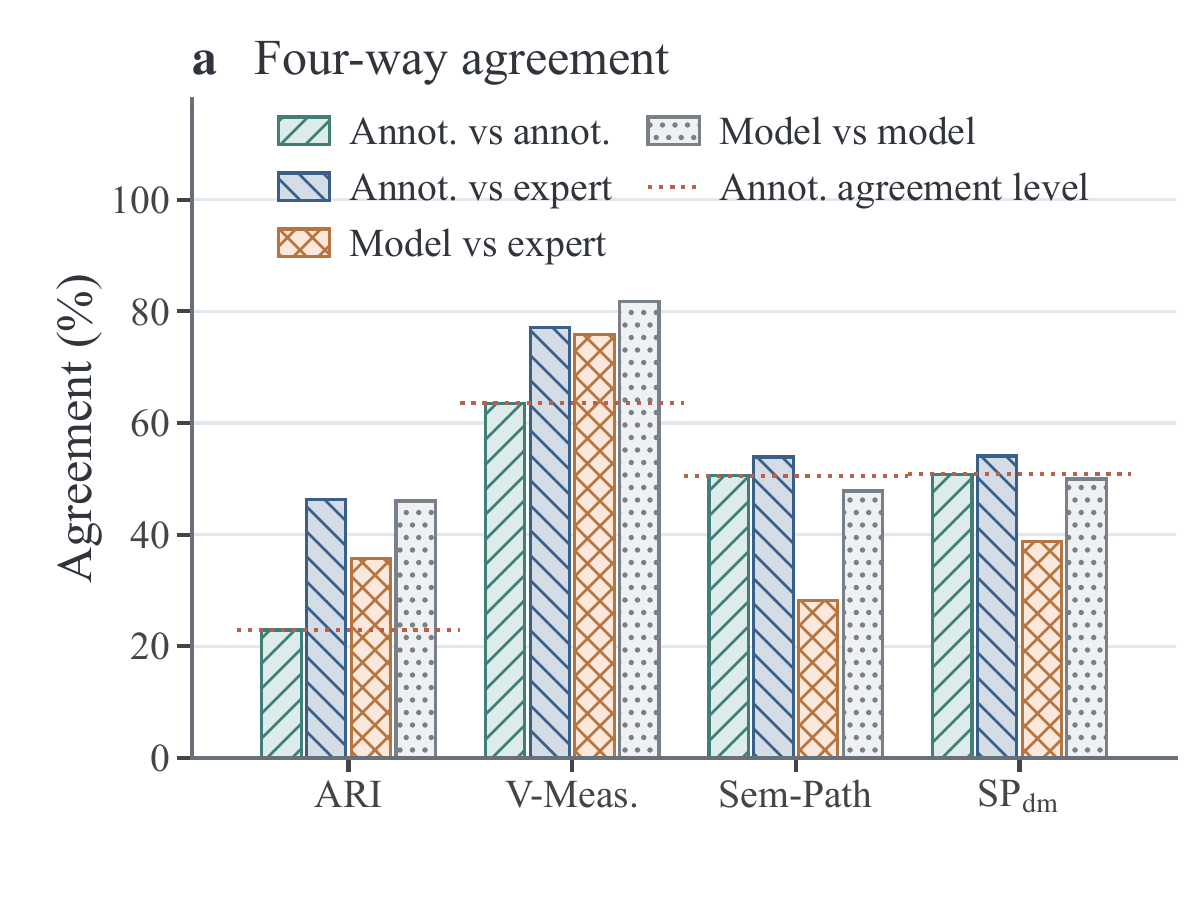}
        \phantomsubcaption
        \label{fig:iaa_fourway_plot}
    \end{subfigure}
    \hfill
    \begin{subfigure}[t]{0.49\linewidth}
        \centering
        \includegraphics[width=\linewidth]{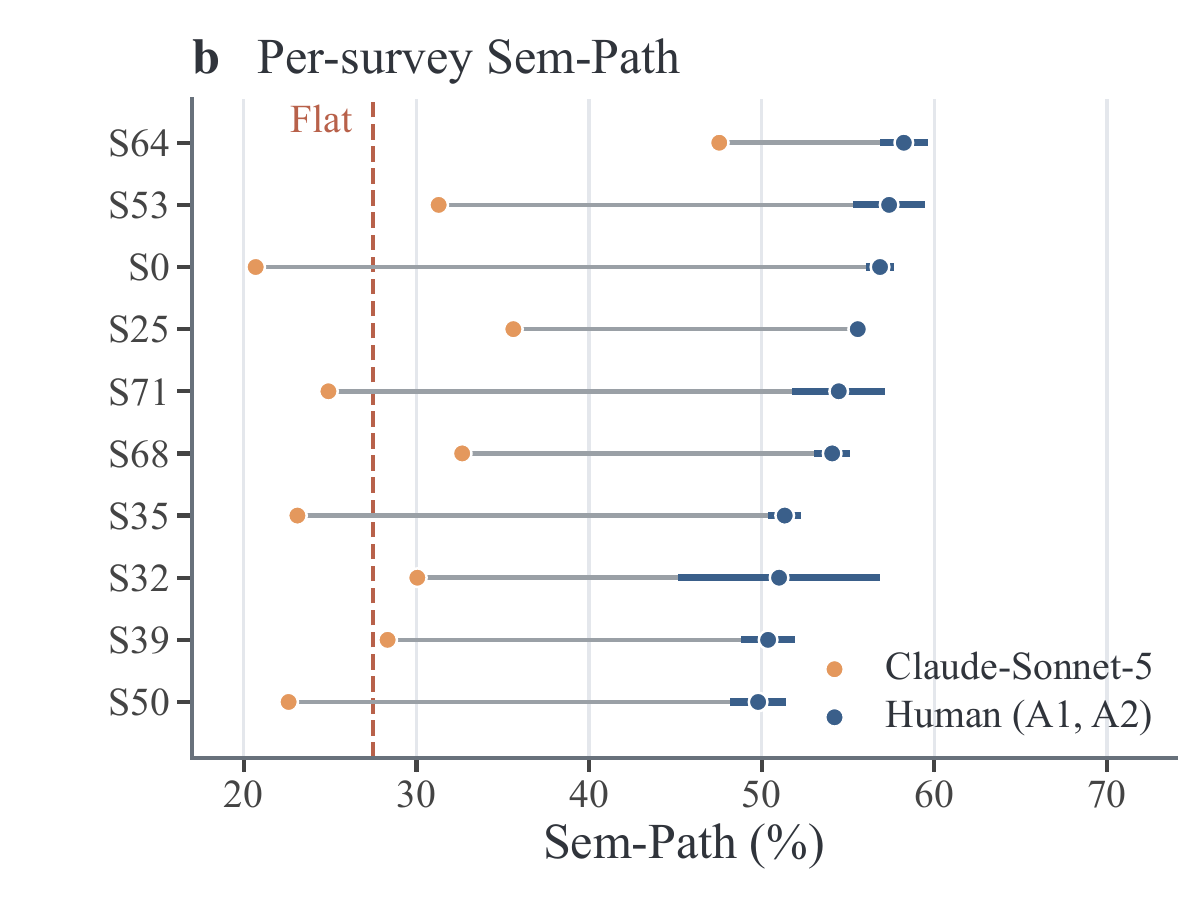}
        \phantomsubcaption
        \label{fig:human_per_survey_plot}
    \end{subfigure}
    \caption{\textbf{Human baseline on the matched 10 surveys} (App.~\ref{app:human_baseline}). (a) Agreement between annotators, expert, and models; the dotted line marks annotator--annotator agreement, which models exceed on ARI and \textsc{V-Measure} but not on \textsc{Sem-Path}. (b) Per-survey \textsc{Sem-Path}: the bar spans the two annotators, the slate marker is their mean, and the clay marker is the strongest model configuration. Humans lead on all 10 surveys.}
    \label{fig:human_baseline}
\end{figure*}

\textbf{Finding 5: Raw \textsc{Sem-Path} is uninformative in this regime; depth-matching recovers a meaningful gap.}
We observe raw \textsc{Sem-Path} scores of 28.12--30.03\% across all 16 configurations (Figure~\ref{fig:sem_path_convergence}), a floor effect: a \textsc{Flat} taxonomy with no organization scores 27.49\% and randomly rewired expert labels 28.11\%, so the weakest configurations sit within 0.02\,pp of the rewired baseline (App.~\ref{app:metric_calibration}). Across model generations, best ARI rises 3.68\,pp while raw \textsc{Sem-Path} moves only 0.87\,pp (App.~\ref{app:prevgen}). Two independent annotators given the same Bottom-Up requirements average 53.91\% \textsc{Sem-Path} on a matched 10-survey subset and agree with each other at 50.56\%, where models reach 28.21\%, and the gap holds on every survey (Figure~\ref{fig:human_baseline}). ARI inverts this and is saturated: models agree with the expert at 35.76\%, the annotators with each other at only 22.92\% (App.~\ref{app:human_baseline}).

We trace the floor effect to the missing chance correction and the depth deficit of Finding~6. Collapsing the expert tree to each model's depth raises scores to 32--42\%, widens the spread from 1.91 to \textbf{9.45\,pp}, and leaves a paired human advantage on all 10 matched surveys ($+13.27$\,pp; 95\% CI $[10.69,16.06]$). Randomizing paper placement while keeping expert labels still scores 77.29\%, 35\,pp above models, so \textsc{Sem-Path} mainly penalizes category vocabulary rather than paper placement (Apps.~\ref{app:metric_calibration} and~\ref{app:human_baseline}).

\textbf{Finding 6: Models do not spontaneously build deep taxonomies, and forcing depth does not help.}
We compare model and expert trees using judge-free, threshold-free statistics in App.~\ref{app:reference_independent}. Expert taxonomies average 4.86 category levels, while all 70 Bottom-Up runs (16 current-generation configurations and 54 previous-generation runs across four input conditions) average 2.99--3.97, a band that contains only 2 of the 72 expert taxonomies (Figure~\ref{fig:expert_depth}). Even the deepest configuration remains 0.89 levels short. Details are in App.~\ref{app:prevgen}. Under the same Bottom-Up requirements, two independent annotators on the matched 10-survey subset reach depth 4.60 and 4.70, matching the 4.70 expert mean on those surveys (App.~\ref{app:human_baseline}), so the deficit is largely model-specific. Because \textsc{Sem-Path} charges $\lambda=1$ per unmatched ancestor, this shortfall alone caps the per-paper score at 45.8--52.4\% (Eq.~\ref{eq:depth_ceiling}) and feeds the floor effect of Finding~5.

With an explicit depth target in the prompt (four previous-generation models; App.~\ref{app:depth_probe}), three of the four comply and GPT-5 reaches the expert depth of 4.86, but matching depth does not improve organization. GPT-5's \textsc{Sem-Path} falls by 2.77\,pp, its ARI drops from 29.02\% to 14.65\%, and the added levels fragment the taxonomy, raising singleton leaves to 66.7\% against the expert's 17.0\%. No model--condition pair gains more than 0.3\,pp \textsc{Sem-Path}. The depth deficit is thus a failure of coherent hierarchical refinement rather than an inability to add levels. The 16 current configurations show the same tension without prompt manipulation: depth predicts singleton rate at $\rho=0.83$ and $p=10^{-4}$ but does not predict ARI at $\rho=-0.27$.

\section{Related Work}

\paragraph{Deep-research-aided survey generation.} Most pipelines adopt a structure-then-content scheme: AutoSurvey~\citep{wang2024autosurvey} uses two-stage generation, \citet{hu2024taxonomy} build taxonomies from citation networks, and SurveyForge~\citep{yan2025surveyforge} pairs outline heuristics learned from human surveys with retrieval~\citep{citron2025deep,li2025webthinker,oepnaideep,zheng2025deepresearcher}. Closest to our organization setting, TaxoAlign/CS-TaxoBench~\citep{lahiri2025taxoalign} evaluates scholarly taxonomy generation against human-written survey taxonomies. \textsc{TaxoBench} differs by grounding each taxonomy in paper-to-category assignments, separating retrieval from organization, and auditing hierarchy metrics under expert paper sets rather than evaluating taxonomy outlines alone.

\paragraph{Taxonomy induction and its evaluation.} Taxonomy construction predates LLMs~\citep{zhang2018taxogen,shen2018hiexpan,zeng2024chainoflayer}, but its evaluation is edge-based: TExEval-2~\citep{bordea2016texeval} scores is-a edges and hierarchical P/R/F1~\citep{kiritchenko2006hierarchical} credits ancestor overlap, both presupposing a shared vocabulary. Model-generated taxonomies invent their own labels, so edge matching is undefined; \textsc{US-TED} and \textsc{Sem-Path} match labels semantically instead.

\paragraph{Deep-research evaluation.} Existing work targets multi-hop retrieval~\citep{Chen2025,Ai-ling2025}, web browsing~\citep{deng2023mind2web,wu2025webwalker}, and report generation~\citep{li2025reportbench,du2025deepresearch}. \textsc{FINDER}~\citep{zhang2025far} scores reports against structured checklists, while \textsc{ResearchRubrics}~\citep{sharma2025researchrubrics} uses human rubrics for reports. \textsc{TaxoBench} instead isolates Information Acquisition and structural Answer Generation with expert-authored references rather than LLM-judged outlines~\citep{sun2025surveybench,yan2025surveyforge}.

\section{Conclusion}
Survey generation requires both finding the right evidence and organizing it. We introduced \textsc{TaxoBench}, which pairs 72 expert-authored taxonomies with 3{,}815 mapped papers and separates these abilities through complementary Deep Research and Bottom-Up modes, making the source of failure observable. Retrieval F1 is strongly associated with \textsc{Sem-Path} ($\rho{=}0.89$) across agents, yet perfect retrieval leaves a 13.27\,pp depth-matched human gap and a 0.89-level depth deficit.

To measure organization beyond flat assignments, we developed \textsc{US-TED}, \textsc{US-NTED}, and \textsc{Sem-Path}, separated alignment-based from capability-based claims, and calibrated them against synthetic controls and a two-annotator human baseline. The calibration bounds what each metric can show: raw \textsc{Sem-Path} moves only 0.87\,pp when a model generation gains 3.68\,pp ARI, and ARI is saturated, since models agree with the expert more than two independent annotators agree with each other. \textsc{TaxoBench} therefore supports a more diagnostic protocol: separate retrieval from organization, distinguish reference alignment from reference-light evidence, and calibrate against controls and human agreement.

\section*{Limitations}
\label{sec:limitations}
\paragraph{Alignment vs.\ capability.}
\textsc{Sem-Path} and \textsc{US-TED}/\textsc{US-NTED} measure agreement with the chosen expert taxonomy, not absolute quality: model taxonomies differ from the references, but not every difference is an error (App.~\ref{app:reference_independent}).

\paragraph{Prompt anchoring of hierarchy depth.}
Given an explicit depth target, three of four models reach 4.0--4.9 levels without gaining more than 0.3\,pp \textsc{Sem-Path}, so the depth deficit is behavioral rather than a capability ceiling; the probe covers four models in one input condition (App.~\ref{app:depth_probe}).

\paragraph{Single-reference evaluation and taxonomy subjectivity.}
Each survey has one expert taxonomy, though others may be valid. Two independent annotators agree at 50.56\% \textsc{Sem-Path}, while models reach only 0.56$\times$ that level and sit 2.4$\times$ further from the expert on \textsc{US-NTED} than the annotators sit from each other (App.~\ref{app:human_baseline}).

\paragraph{Annotation reliability and human-baseline coverage.}
We report no inter-annotator agreement for the paper-to-category mapping, the highest-priority gap (App.~\ref{app:multi_category} gives the closest check), and the human baseline covers 10 of 72 surveys.

\paragraph{Domain scope.}
All 72 surveys are LLM-related, so absolute numbers may not transfer to other fields, although two interdisciplinary surveys show the same pattern (App.~\ref{app:cross_domain}).

\paragraph{Evaluation workflow simplification.}
Agents are evaluated as deployed, without richer agentic scaffolding.

\paragraph{\textsc{Sem-Path} lacks a chance correction.}
Absolute values are interpretable only against the floors of App.~\ref{app:metric_calibration}; for current shallow trees the dynamic range is about 1\,pp, and our depth-matched variant corrects this only after the fact. A chance-corrected hierarchy metric is the clearest open problem.

\paragraph{Encoder and judge dependency.}
Absolute scores depend on the encoder and the GPT-4o judge, but \textsc{US-TED} and \textsc{Sem-Path} rankings are unchanged across three encoders (Kendall's $\tau{=}1.0$; App.~\ref{app:embedding_sensitivity}) and the judge agrees with humans at $\kappa{=}0.89$ (App.~\ref{app:judge_co_with_human}).

\section*{Ethical Considerations}
\label{sec:ethics}
\paragraph{Data provenance and privacy.}
We construct the dataset entirely from publicly available computer science surveys and their cited papers. We use scientific content such as titles and abstracts and do not collect non-public personal or sensitive information. We use these materials only for non-commercial academic research.

\paragraph{Human subjects.}
Most experiments evaluate AI systems. Two auxiliary studies involve human participants: expert agreement with GPT-4o judgments in App.~\ref{app:judge_co_with_human} and non-author human taxonomy construction in App.~\ref{app:human_baseline}. These minimal-risk tasks collect professional judgments about scientific taxonomies, not sensitive personal information. Participants were recruited from CS graduate students familiar with AI/ML literature, and the paper reports only aggregated results under pseudonymous identifiers. Everyone who annotated or rated data for this work, including the dataset annotators, was paid 100 RMB per hour. App.~\ref{app:human_protocol} states what participants were told before starting, including that they could withdraw. We do not release participant-level material.

\paragraph{Broader impact and potential misuse.}
\textsc{TaxoBench} can support research on automated survey generation and help researchers navigate a growing scientific literature. The same technology can also produce low-quality reviews and create academic-integrity risks. Our experiments show that current agents remain far below expert performance, so we recommend AI-assisted survey writing with expert oversight rather than full automation.

\bibliography{main}

\newpage

\appendix

\renewcommand{\dbltopfraction}{0.95}
\renewcommand{\dblfloatpagefraction}{0.6}
\setcounter{dbltopnumber}{4}
\renewcommand{\topfraction}{0.95}
\renewcommand{\bottomfraction}{0.6}
\renewcommand{\textfraction}{0.05}
\renewcommand{\floatpagefraction}{0.6}
\setcounter{topnumber}{4}
\setcounter{totalnumber}{6}
\makeatletter
\setlength{\@fptop}{0pt}
\setlength{\@dblfptop}{0pt}
\makeatother

\section*{Appendix Roadmap}
Apps.~\ref{app:metric_details}--\ref{app:reference_independent} give formal metric definitions, calibration, and the validity framing. Apps.~\ref{app:ari_intersection}--\ref{app:robustness} report the supplementary experiments behind the findings, together with the judge validation, error analysis, human baseline, and robustness checks. Apps.~\ref{app:survey_selection_pipeline}--\ref{app:prompts} collect dataset construction, model details, reproducibility, the use of AI assistants, and all prompts.

\section{Metric Details}
\label{app:metric_details}

\subsection{Adjusted Rand Index (ARI)}
\label{sec:ari_details}

The Adjusted Rand Index (ARI) is a chance-corrected measure of similarity between two data clusterings. Given a set of $N$ papers, let $U = \{u_1, \dots, u_R\}$ be the ground truth partition (expert taxonomy) and $V = \{v_1, \dots, v_C\}$ be the model-generated partition.

We first consider all $\binom{N}{2}$ pairs of papers and categorize them into four types based on their assignment in $U$ and $V$:
\begin{itemize}
    \item \textbf{TP (True Positive)}: The number of pairs that are in the same cluster in both $U$ and $V$.
    \item \textbf{TN (True Negative)}: The number of pairs that are in different clusters in both $U$ and $V$.
    \item \textbf{FP (False Positive)}: The number of pairs that are in different clusters in $U$ but in the same cluster in $V$.
    \item \textbf{FN (False Negative)}: The number of pairs that are in the same cluster in $U$ but in different clusters in $V$.
\end{itemize}

The standard Rand Index (RI) measures the percentage of correct decisions:
\begin{equation}
    RI = \frac{TP + TN}{TP + FP + FN + TN}
\end{equation}

However, the RI does not yield a value of 0 for random partitions. The ARI corrects this by normalizing the RI using its expected value under a random permutation model:
\begin{equation}
    ARI = \frac{RI - E[RI]}{\max(RI) - E[RI]}
\end{equation}

Specifically, let $n_{ij}$ be the number of papers in both class $u_i$ and cluster $v_j$, and let $a_i = \sum_j n_{ij}$ and $b_j = \sum_i n_{ij}$ be the row and column sums of the contingency table. The terms in the ARI formula are calculated as follows:
\begin{align}
    \text{Index} &= \sum_{i,j} \binom{n_{ij}}{2} = TP \\
    \text{Max Index} &= \frac{1}{2} \left[ \sum_i \binom{a_i}{2} + \sum_j \binom{b_j}{2} \right] \\
    \text{Expected Index} &= \frac{\left[ \sum_i \binom{a_i}{2} \right] \left[ \sum_j \binom{b_j}{2} \right]}{\binom{N}{2}}
\end{align}

Combining these, the computational formula for ARI is:
\begin{equation}
    ARI = \frac{\sum_{ij} \binom{n_{ij}}{2} - \left[ \sum_i \binom{a_i}{2} \sum_j \binom{b_j}{2} \right] / \binom{N}{2}}{ \frac{1}{2} \left[ \sum_i \binom{a_i}{2} + \sum_j \binom{b_j}{2} \right] - \left[ \sum_i \binom{a_i}{2} \sum_j \binom{b_j}{2} \right] / \binom{N}{2} }
\end{equation}

This linear transformation ensures that ARI has a maximum value of 1 for perfect agreement and an expected value of 0 for random clustering, making it a robust metric for comparing taxonomies with different numbers of clusters.

\subsection{Semantic Label Similarity and Cost Calibration}
\label{sec:metric_similarity_calibration}

\paragraph{Embedding similarity.}
For each topic label (node name) $x$, we compute a dense embedding
$\mathbf{e}(x)\in\mathbb{R}^d$ using a fixed sentence encoder.
We measure semantic relatedness by cosine similarity
\begin{equation}
c(x,y)\coloneqq \cos(\mathbf{e}(x),\mathbf{e}(y))
=\frac{\mathbf{e}(x)^\top\mathbf{e}(y)}{\|\mathbf{e}(x)\|\,\|\mathbf{e}(y)\|}.
\end{equation}
Since cosine similarity can be negative in practice, we adopt a clipped
similarity used consistently by both US-TED/US-NTED and \textsc{Sem-Path}:
\begin{equation}
\mathrm{Sim}(x,y)\coloneqq \max\!\bigl(0,\,c(x,y)\bigr)\in[0,1].
\end{equation}

\paragraph{Edit-cost calibration.}
We convert similarity to a renaming cost by
\begin{equation}
\mathrm{cost}_{\mathrm{ren}}(x\!\rightarrow\! y)\coloneqq 1-\mathrm{Sim}(x,y)\in[0,1],
\end{equation}
and set insertion/deletion costs to $1$ \emph{per node}.
This calibration has two practical benefits.
First, clipping ensures that renaming cost stays on the same scale as
insertion/deletion (all in $[0,1]$), which stabilizes normalization in
US-NTED and avoids pathological cases where negative cosine would yield
$\mathrm{cost}_{\mathrm{ren}}>1$.
Second, in the embedding space used, a negative cosine value does not reliably indicate
semantic opposition for short topic labels. We therefore treat
it as \emph{no semantic match}, with $\mathrm{Sim}=0$, rather than assigning an
exaggerated penalty.

\subsection{Unordered Semantic Tree Edit Distance (US-TED/US-NTED) Explanation}
\label{sec:metric_us_ted}

\subsubsection{Definition}
\label{sec:metric_us_ted_def}

We compare the expert hierarchy $T_h^*$ and the model hierarchy $\hat{T}_h$ using an
\emph{unordered} semantic tree edit distance, where sibling order is ignored and
children are matched by minimum-cost assignment. Both hierarchies contain only
internal-topic nodes; we exclude paper nodes. The leaves of $T_h$ are \textbf{paper categories},
the terminal category nodes to which papers are assigned.

Let $\mathrm{Ch}(u)$ denote the multiset of children of node $u$, and let
$|T_u|$ denote the number of nodes in the subtree rooted at $u$ (including $u$).
We use the calibrated edit costs from Section~\ref{sec:metric_similarity_calibration}:
insertion and deletion cost $1$ per node, and renaming $x\!\rightarrow\!y$ costs
$\mathrm{cost}_{\mathrm{ren}}(x\!\rightarrow\!y)=1-\mathrm{Sim}(x,y)\in[0,1]$.

We define the unordered node-to-node distance $D(u,v)$ recursively:
\begin{equation}
D(u,v)
~\coloneqq~
\mathrm{cost}_{\mathrm{ren}}(u\!\rightarrow\!v)
~+~
\mathrm{MatchCost}\bigl(\mathrm{Ch}(u),\mathrm{Ch}(v)\bigr).
\end{equation}
If both $u$ and $v$ are leaves, then $\mathrm{MatchCost}=0$ and $D(u,v)$ reduces to
the renaming cost. If one node is a leaf and the other has children, the unmatched
subtrees are charged by unit per-node insertion/deletion via $|T_u|$.

To define $\mathrm{MatchCost}$ when both nodes have children, let
$\mathrm{Ch}(u)=\{u_1,\ldots,u_m\}$ and $\mathrm{Ch}(v)=\{v_1,\ldots,v_n\}$, and set
$k=\max(m,n)$. We build a $k\times k$ cost matrix $C$ by padding with dummy children:
\begin{equation}
C_{ij}=
\begin{cases}
D(u_i,v_j), & 1\le i\le m,\ 1\le j\le n,\\
|T_{u_i}|,  & 1\le i\le m,\ n< j\le k,\\
|T_{v_j}|,  & m< i\le k,\ 1\le j\le n,\\
0,          & m< i\le k,\ n< j\le k,
\end{cases}
\end{equation}
where matching a real child to a dummy child corresponds to deleting or inserting
the entire subtree at unit cost per node, and dummy--dummy matches contribute zero.
The unordered children matching cost is defined by the minimum assignment:
\begin{equation}
\mathrm{MatchCost}\bigl(\mathrm{Ch}(u),\mathrm{Ch}(v)\bigr)
~\coloneqq~
\min_{\sigma\in\mathfrak{S}_k}\sum_{i=1}^{k} C_{i,\sigma(i)},
\end{equation}
which we solve via the Hungarian algorithm.

Finally, letting $r^*$ and $\hat{r}$ be the roots of $T_h^*$ and $\hat{T}_h$, the
tree-level distance is
\begin{equation}
\mathrm{US\text{-}TED}(T_h^*,\hat{T}_h)\coloneqq D(r^*,\hat{r}).
\end{equation}
We normalize by the total number of hierarchy nodes:
\begin{equation}
\mathrm{US\text{-}NTED}(T_h^*,\hat{T}_h)
\coloneqq
\frac{\mathrm{US\text{-}TED}(T_h^*,\hat{T}_h)}{|T_h^*|+|\hat{T}_h|}.
\end{equation}

\subsubsection{Basic Properties}
\label{sec:metric_us_ted_props}

\paragraph{Range and normalization.}
\textbf{Proposition.}
For any two hierarchies $T_h^*$ and $\hat{T}_h$,
\begin{equation}
0 \le \mathrm{US\text{-}TED}(T_h^*,\hat{T}_h) \le |T_h^*|+|\hat{T}_h|,
\qquad
\end{equation}
\begin{equation}
0 \le \mathrm{US\text{-}NTED}(T_h^*,\hat{T}_h) \le 1.
\end{equation}
\textbf{Proof sketch.}
Non-negativity holds since all edit costs are non-negative.
For the upper bound, consider a valid (not necessarily optimal) edit sequence that
deletes every node in $\hat{T}_h$ (cost $|\hat{T}_h|$) and then inserts every node in
$T_h^*$ (cost $|T_h^*|$). Since US-TED is defined as a minimum edit cost under the
same per-node insertion/deletion convention, it cannot exceed $|T_h^*|+|\hat{T}_h|$.
Dividing by $|T_h^*|+|\hat{T}_h|$ yields $0\le \mathrm{US\text{-}NTED}\le 1$.

\paragraph{Permutation invariance.}
\textbf{Proposition.}
US-TED is invariant to any permutation of sibling order in either tree.
\textbf{Proof sketch.}
At each pair of nodes $(u,v)$, the children matching cost is defined by a minimum
assignment over a cost matrix $C$ whose rows/columns correspond to the children of
$u$ and $v$ (plus dummies). Permuting sibling order permutes rows and/or columns of
$C$ but does not change the optimal assignment cost, hence leaves
$\mathrm{MatchCost}$ and $D(u,v)$ unchanged.

\paragraph{Why unordered.}
Taxonomy siblings do not have a canonical left-to-right order. An ordered tree edit
distance would penalize pure sibling permutations as structural errors and conflate sibling
ordering with organization. By using unordered matching via minimum-cost
assignment, US-TED focuses on structural organization (parent--child relations and
subtree composition) rather than arbitrary sibling ordering.

\paragraph{Relation to unordered tree edit distance.}
We state the positioning of US-TED precisely. General unordered tree edit distance is
NP-hard, so US-TED is not a computation of it. By matching children level by level
under a bijection and recursing, US-TED is an instance of the classical
\emph{constrained} (equivalently, top-down / degree-two) tree edit distance family,
for which the minimum-cost assignment at each node yields a polynomial-time algorithm;
our contribution is not the matching scheme but the semantic cost calibration
(clipped-cosine renaming against unit insert/delete) and its use as a taxonomy
evaluation measure. Two consequences follow. First, US-TED upper-bounds the true
unordered edit distance, since every constrained edit script is a valid edit script.
Second, we make no claim that US-TED satisfies the triangle inequality: constrained
edit distances are metrics under uniform costs, but our renaming cost
$1-\mathrm{Sim}$ derived from cosine similarity is not itself a metric, so US-TED
should be read as a calibrated dissimilarity rather than a metric in the formal sense.
This does not affect any reported result, all of which use US-TED only to compare a
model tree against a fixed expert tree.

\subsection{Semantic Path Similarity (\textsc{Sem-Path})}
\label{sec:metric_sem_path}

\subsubsection{Aligned-paper set and path extraction}
\label{sec:metric_sem_path_alignment}

\paragraph{Aligned-paper set.}
\textsc{Sem-Path} is computed on aligned paper pairs.
Let $D_a$ denote the set of aligned paper pairs $(d,\hat{d})$ between the expert
tree $T^*$ and the model tree $\hat{T}$. We construct $D_a$ by normalizing paper
titles and performing deterministic matching between expert papers and retrieved
model papers (details of normalization rules and tie-breaking are reported in this
section). This step is necessary to avoid inflated scores due to spurious title
overlaps (false positives) or minor title variants (false negatives).

\paragraph{Ancestor-path extraction.}
For each aligned paper $d\in D_a$, we extract the root-to-leaf chain in both trees
and keep only internal topic labels. Concretely, if the root-to-paper path in
$T^*$ is $(r^*, \ldots, u, d)$, we set $S_d=(r^*,\ldots,u)$, excluding the paper
node $d$. Likewise, if the root-to-paper path in $\hat{T}$ is $(\hat{r}, \ldots, \hat{u}, \hat{d})$, we set $\hat{S}_d=(\hat{r},\ldots,\hat{u})$.

\subsubsection{Distance function and monotone alignment}
\label{sec:metric_sem_path_dp}

\paragraph{Clipped cosine distance.}
We use the clipped cosine distance $\delta(x,y)\coloneqq 1-\mathrm{Sim}(x,y)\in[0,1]$, which equals the renaming cost defined in Section~\ref{sec:metric_similarity_calibration}. This keeps the distance on the same scale as the unit unmatched-node penalty.

\paragraph{Order-preserving alignment.}
Given two ancestor-label sequences $S_d$ and $\hat{S}_d$, we compute an
order-preserving minimum-cost alignment that tolerates different granularities
(i.e., different depths) while preserving relative order.
Without loss of generality, let $A=(a_1,\ldots,a_p)$ be the shorter sequence and
$B=(b_1,\ldots,b_q)$ be the longer sequence ($p\le q$), where $A,B$ correspond to
$S_d,\hat{S}_d$ up to swapping. We compute a subsequence-style dynamic program:
\begin{equation}
\mathrm{dp}[0,j]=0,
\end{equation}
\begin{equation}
\mathrm{dp}[i,j]=\min\!\bigl(\mathrm{dp}[i-1,j-1]+\delta(a_i,b_j),\ \mathrm{dp}[i,j-1]\bigr).
\end{equation}
Intuitively, $\mathrm{dp}[i,j]$ matches the first $i$ labels of the shorter chain
to an ordered subsequence within the first $j$ labels of the longer chain.

\paragraph{Per-paper cost and score.}
We define the per-paper alignment cost by adding a penalty for each unmatched
extra label in the longer chain:
\begin{equation}
J_d \coloneqq \mathrm{dp}[p,q] + \lambda\,(q-p),
\end{equation}
where $\lambda \ge 0$ is a penalty coefficient controlling the cost of each
unmatched extra node. In all experiments, we set $\lambda=1$.

The final metric maps cost to similarity and averages over aligned papers:
\begin{equation}
\textsc{Sem-Path}\coloneqq \frac{1}{|D_a|}\sum_{d\in D_a} \frac{1}{1+J_d}.
\end{equation}

\subsubsection{Basic Properties and Design Rationale}
\label{sec:metric_sem_path_props}

\paragraph{Range.}
Since $J_d\ge 0$, we have $\mathrm{Sem\text{-}Path}(d)\in(0,1]$ and therefore
$\mathrm{Sem\text{-}Path}\in(0,1]$.

\paragraph{Granularity tolerance with order sensitivity.}
\textsc{Sem-Path} allows different hierarchy granularities because matching is performed
between a shorter chain and an ordered subsequence of a longer chain. However, it
is sensitive to mis-ordered or semantically inconsistent ancestor placement: if
the model swaps high-level topics or inserts unrelated ancestors, the DP cost
increases via $\delta(\cdot,\cdot)$ and/or the unmatched penalty $\lambda\,(q-p)$.

\paragraph{Relation to soft-cardinality baselines.}
Soft-cardinality metrics such as NSR/NSP primarily measure semantic coverage of a
\emph{set} of node labels and can be insensitive to hierarchical placement errors
(e.g., correct topics but wrong parent--child relations). In contrast, US-TED/US-NTED
and Sem-Path encode explicit structural constraints: US-TED penalizes topology and
subtree-level edits, while \textsc{Sem-Path} evaluates whether each aligned paper is placed
under a semantically consistent ancestor chain. A detailed comparison, including
failure modes and illustrative cases, is provided in App.~\ref{sec:metric_nsr_nsp_discussion}.

\subsection{Paper-title Normalization and Alignment}
\label{sec:metric_paper_alignment}

To compute retrieval scores and paper-conditioned metrics (e.g., \textsc{Sem-Path}),
we align model-returned papers with expert-cited papers. We do not rely on exact title
match or a single identifier (e.g., \textsc{DOI}/arXiv ID), since identifiers can be missing
and the same work may appear in multiple versions (arXiv vs.\ venue).

\paragraph{Normalization.}
We apply deterministic text normalization (lowercasing and whitespace/punctuation cleanup)
to titles and other available textual fields.

\paragraph{Matching rule.}
For an expert paper $p$ and a model paper $\hat{p}$, we compute a semantic similarity
score $s(p,\hat{p})\in[0,1]$ on the normalized text representation.
We align $(p,\hat{p})$ if either (i) $s(p,\hat{p})=1$, or (ii) $0.6 \le s(p,\hat{p}) < 1$
and the normalized titles satisfy a strict containment check (one contains the other).
Otherwise, we treat them as different papers. If multiple candidates match an expert paper,
we keep the one with the highest $s(p,\hat{p})$.

\subsection{Why Soft-Cardinality Baselines (NSR/NSP) Are Insufficient for Taxonomy Structure}
\label{sec:metric_nsr_nsp_discussion}

\paragraph{What NSR/NSP measure.}
Node Soft Recall (\textsc{NSR}) and Node Soft Precision (\textsc{NSP}) are soft-cardinality
extensions of set-based Recall/Precision~\citep{franti2023soft}. They quantify \emph{semantic coverage}
between two \emph{collections of labels} by discounting near-duplicates via pairwise semantic similarity.
Given two node-label lists $A=(a_1,\dots,a_{|A|})$ and $B=(b_1,\dots,b_{|B|})$, define the soft cardinality
\begin{equation}
c(A) \;\coloneqq\; \sum_{i=1}^{|A|}\frac{1}{\sum_{j=1}^{|A|}\mathrm{Sim}(a_i,a_j)}.
\end{equation}
Following~\citep{franti2023soft}, define
\begin{align}
\textsc{NSR}(A,B) &\coloneqq \frac{c(A)+c(B)-c(A\uplus B)}{c(A)}, \\
\textsc{NSP}(A,B) &\coloneqq \frac{c(A)+c(B)-c(A\uplus B)}{c(B)}, \\
\textsc{Soft F1}(A,B) &\coloneqq \frac{2\,\textsc{NSP}(A,B)\,\textsc{NSR}(A,B)}{\textsc{NSP}(A,B)+\textsc{NSR}(A,B)}.
\end{align}
Here $A\uplus B$ denotes \emph{multiset union with multiplicities}, implemented as list concatenation.

\noindent\textbf{Our implementation.}
We instantiate $\mathrm{Sim}$ with embedding cosine similarity and clamp negatives:
\begin{equation}
\mathrm{Sim}(x,y)\;\coloneqq\;\max\!\bigl(0,\,\cos(\mathbf{e}(x),\mathbf{e}(y))\bigr),
\end{equation}
where $\mathbf{e}(\cdot)$ is produced by a SentenceTransformer encoder (e.g., \texttt{all-MiniLM-L6-v2})
with $\ell_2$-normalized embeddings. We collect \emph{hierarchy} node labels by preorder traversal
(including the root; excluding paper leaves); ordering is ignored and multiplicities are kept.
We report the soft-cardinality baselines \textsc{NSR}, \textsc{NSP}, and \textsc{Soft F1} in
Table~\ref{tab:soft_hierarchy_alignment}. They remain relatively high and
compress model differences even when structure-aware metrics (e.g., \textsc{US-NTED} and
\textsc{Sem-Path}) indicate substantial structural gaps.

\begin{table}[!htb]
\centering
\footnotesize 
\setlength{\tabcolsep}{10pt} 
\renewcommand{\arraystretch}{0.95} 
\begin{tabular}{lccc}
\toprule
\multirow{2}{*}{Model} & \multicolumn{3}{c}{Hierarchy-Level Metrics} \\
\cmidrule(lr){2-4}
& NSR$\uparrow$ & NSP$\uparrow$ & Soft F1$\uparrow$ \\
\midrule
\tablegroupc{tintDR}{4}{\grouplabel{Deep Research mode}{}} \\
\modelname{openai}{o3} & 0.76 & \textbf{0.92} & 0.83 \\
\modelname{doubao-color}{Doubao} & 0.82 & 0.80 & 0.81 \\
\modelname{deepseek-color}{DeepSeek} & 0.77 & 0.85 & 0.81 \\
\modelname{gemini-color}{Gemini} & 0.88 & 0.77 & 0.82 \\
\modelname{grok}{Grok} & 0.79 & 0.87 & 0.83 \\
\modelname{perplexity-color}{Perplexity} & 0.88 & 0.80 & \textbf{0.84} \\ 
\modelname{qwen-color}{Qwen} & \textbf{0.91} & 0.77 & 0.83 \\

\midrule
\tablegroupc{tintBU}{4}{\grouplabel{Bottom-Up mode}{}} \\
\tablegroup{4}{\textit{Non-thinking-based}} \\
\modelname{claude-color}{Claude-Sonnet-4.5} & 0.87 & 0.85 & \textbf{0.86} \\
\modelname{openai}{GPT-5} & 0.82 & 0.85 & 0.84 \\
\modelname{gemini-color}{Gemini-3-Pro*} & 0.87 & 0.83 & 0.85 \\
\modelname{deepseek-color}{DeepSeek-V3.2} & 0.82 & \textbf{0.87} & 0.85 \\
\modelname{qwen-color}{Qwen3-Max*} & 0.80 & 0.86 & 0.83 \\
\modelname{kimi-color}{Kimi-K2} & \textbf{0.89} & 0.82 & 0.85 \\ 
\midrule
\tablegroup{4}{\textit{Thinking-based}} \\
\modelname{claude-color}{Claude-Sonnet-4.5-Thinking} & 0.86 & 0.85 & \textbf{0.85} \\
\modelname{openai}{GPT-5-Thinking} & 0.83 & 0.85 & 0.84 \\
\modelname{gemini-color}{Gemini-3-Pro-Thinking*} & 0.87 & 0.83 & \textbf{0.85} \\
\modelname{deepseek-color}{DeepSeek-V3.2-Thinking} & 0.82 & \textbf{0.89} & \textbf{0.85} \\
\modelname{qwen-color}{Qwen3-Max-Thinking*} & 0.82 & 0.86 & 0.84 \\
\modelname{kimi-color}{Kimi-K2-Thinking} & \textbf{0.88} & 0.82 & \textbf{0.85} \\
\bottomrule
\end{tabular}
\caption{\textbf{Soft set-matching baselines} (NSR, NSP, Soft F1) for taxonomy evaluation.}
\label{tab:soft_hierarchy_alignment}
\end{table}

\paragraph{Why label coverage cannot evaluate taxonomy structure.}
A taxonomy \emph{structure} metric must depend on parent--child and ancestor relations, i.e.,
edge/ancestry errors should be penalized in an interpretable way aligned with hierarchical
organization. By construction, \textsc{NSR}/\textsc{NSP} depend only on the label inventory and
pairwise similarities and impose no edge or ancestry constraints; therefore, as suggested by
Table~\ref{tab:soft_hierarchy_alignment}, they primarily reflect semantic coverage/redundancy
rather than hierarchy correctness. The following proposition formalizes this limitation.

\paragraph{Proposition (Structure-blindness under label-preserving rewiring).}
Let $A=(a_1,\ldots,a_n)$ and $B=(b_1,\ldots,b_n)$ be two \emph{lists} of hierarchy-node labels,
and assume $B$ is a permutation of $A$ (equivalently, $A$ and $B$ induce the same label multiset).
Assume $\mathrm{Sim}(\cdot,\cdot)$ is deterministic, symmetric, and depends only on the label strings.
Define soft cardinality by
\[
c(A)\coloneqq \sum_{i=1}^{n}\frac{1}{\sum_{j=1}^{n}\mathrm{Sim}(a_i,a_j)},
\qquad
c(B)\ \text{analogously},
\]
and assume $\sum_{j=1}^{n}\mathrm{Sim}(a_i,a_j)>0$ for all $i$ (e.g., when $\mathrm{Sim}(x,x)=1$).
Let $A\uplus B$ denote multiset union with multiplicities, implemented as list concatenation.
Then $\textsc{NSR}(A,B)=\textsc{NSP}(A,B)=1$, regardless of any differences in the parent--child
relations of the underlying trees, as long as their hierarchy-node label multisets match.

\noindent\emph{Proof sketch.}
Because $B$ is a permutation of $A$ and $\mathrm{Sim}$ depends only on label strings, $c(A)=c(B)$.
Consider $A\uplus B$ (concatenation). For any $a_i$,
\[
\sum_{z\in A\uplus B}\mathrm{Sim}(a_i,z)
=
\sum_{j=1}^{n}\mathrm{Sim}(a_i,a_j)
+
\sum_{j=1}^{n}\mathrm{Sim}(a_i,b_j)
=
2\sum_{j=1}^{n}\mathrm{Sim}(a_i,a_j),
\]
since $(b_1,\ldots,b_n)$ is a reordering of $(a_1,\ldots,a_n)$. Hence each occurrence of $a_i$
in $A\uplus B$ has a denominator that is doubled, so each copy contributes half of its original
term; with two copies, their contributions sum to the original. Summing over all labels yields
$c(A\uplus B)=c(A)$. Substituting into the definitions of \textsc{NSR} and \textsc{NSP} gives
$\textsc{NSR}(A,B)=\textsc{NSP}(A,B)=1$.
\hfill$\square$

\paragraph{Counterexample 1.}
Consider two hierarchies with identical label multiset but different parent--child relations:
\[
T_1:\;\; R\bigl(A(B,C),\, D(E,F)\bigr)
\quad
T_2:\;\; R\bigl(A(B,E),\, D(C,F)\bigr),
\]
where the attachments of $C$ and $E$ are swapped. By the proposition, \textsc{NSR}/\textsc{NSP}
attain their maxima when computed from the hierarchy label lists, despite a non-trivial structural change.

In contrast, structure-aware metrics penalize such rewiring. Under \textsc{US-TED}/\textsc{US-NTED}
(App.~\ref{sec:metric_us_ted}), a parent change is not a primitive move. It must be realized
by deleting a (sub)tree and re-inserting it elsewhere. With insertion/deletion cost $1$ per node,
reattaching a subtree of size $s$ has positive edit cost, and under unit insertion/deletion it is
lower-bounded by deleting and re-inserting that subtree ($2s$) even if rename costs are $0$; hence
\textsc{US-TED} is strictly positive for this example.

\textsc{Sem-Path} (App.~\ref{sec:metric_sem_path}) is also affected: any paper under the swapped
branches experiences an altered ancestor chain. For such a paper, at least one aligned ancestor label
differs (e.g., matching $C$ against $E$ at the corresponding depth). Since the per-node distance is
$1-\max(0,\cos(\cdot,\cdot))$, the cumulative alignment cost is positive whenever $\mathrm{Sim}(C,E)<1$,
yielding a decreased path similarity $1/(1+\mathrm{cost})$.

\paragraph{Counterexample 2.}
Soft cardinality is not monotone: adding semantically similar labels can decrease $c(\cdot)$ because
row-sums increase. Consequently, \textsc{NSR}/\textsc{NSP} can exhibit non-intuitive scaling, including
Recall-like values above $1$, even under clamped similarities with $\mathrm{Sim}\in[0,1]$.

Let $A=(a)$ and $B=(b_1,b_2)$ and assume (after clamping) that
\[
\mathrm{Sim}(a,b_1)=\mathrm{Sim}(a,b_2)=1,\qquad \mathrm{Sim}(b_1,b_2)=0,
\]
with $\mathrm{Sim}(x,x)=1$. Using $A\uplus B$ as multiset union with multiplicities (concatenation),
we obtain
\[
\begin{aligned}
c(A) &= 1,\\
c(B) &= \tfrac{1}{1+0} + \tfrac{1}{0+1} = 2,\\
c(A\uplus B) &= \tfrac13 + \tfrac12 + \tfrac12 = \tfrac43,\\
\textsc{NSR}(A,B) &= \bigl(c(A)+c(B)-c(A\uplus B)\bigr)/c(A) = \tfrac53 > 1,\\
\textsc{NSP}(A,B) &= \bigl(c(A)+c(B)-c(A\uplus B)\bigr)/c(B) = \tfrac56.
\end{aligned}
\]
We therefore interpret \textsc{NSR}/\textsc{NSP} as diagnostics of semantic overlap and
redundancy under a particular soft-cardinality geometry, rather than stable Recall/Precision metrics for
hierarchical \emph{structure}.

\paragraph{Discussion.}
Because \textsc{NSR}/\textsc{NSP} operate on label inventories, they necessarily penalize any discrepancy
in intermediate-node labels, even when the model produces a coarser yet hierarchically consistent taxonomy.
For example, contracting an intermediate node changes the label multiset and reduces soft overlap. In contrast,
\textsc{US-NTED} and \textsc{Sem-Path} expose such differences as explicit, localized edit/alignment costs
(node deletions/insertions in \textsc{US-NTED}; unmatched ancestor steps in \textsc{Sem-Path}), which is more
directly tied to structural operations on trees and ancestor chains.

\paragraph{Summary.}
\textsc{NSR}/\textsc{NSP} quantify soft semantic overlap between \emph{label collections} and are useful
auxiliary diagnostics for coverage and redundancy. However, they do not encode parent--child or ancestor-chain
constraints and can exhibit non-intuitive scaling (including \textsc{NSR}$>1$). Therefore, they are insufficient
as primary measures of taxonomy \emph{structure} quality. We report \textsc{NSR}/\textsc{NSP} only as auxiliary
baselines/diagnostics, and rely on \textsc{US-TED}/\textsc{US-NTED} (global unordered semantic edit cost) and
\textsc{Sem-Path} (per-paper ancestor-chain consistency) as the main hierarchy-level evaluations.

\subsection{Algorithms}
\label{app:Algorithm}
\noindent
\begin{minipage}[t]{0.48\textwidth}
\begin{algorithm}[H]
   \caption{Semantic Path Alignment for \textsc{Sem-Path} Metric}
   \label{alg:sem_path}
\begin{algorithmic}
   \STATE {\bfseries Input:} Ancestor-label sequence $S \in \mathbb{R}^{m \times 1}$, Ancestor-label sequence $\hat{S} \in \mathbb{R}^{n \times 1}$, $m \le n$
   \STATE {\bfseries Parameter:} Unmatched penalty $\lambda \ge 0$ (set $\lambda=1$)
   \STATE Initialize cost matrix $D \in \mathbb{R}^{(m+1) \times (n+1)}$ with $\infty$
   \STATE $D[0,:] \gets 0$
   \FOR{$i=1$ {\bfseries to} $m$}
       \FOR{$j=i$ {\bfseries to} $n$}
           \STATE $cost_{\text{match}} \gets D[i-1,j-1] + \delta(S_i, \hat{S}_j)$
           \STATE $cost_{\text{skip}} \gets D[i,j-1]$
           \STATE $D[i,j] \gets \min(cost_{\text{match}}, cost_{\text{skip}})$
       \ENDFOR
   \ENDFOR
   \STATE \textbf{Return} $J(S, \hat{S}) \gets D[m,n] + \lambda \cdot (n-m)$
\end{algorithmic}
\end{algorithm}
\end{minipage}
\hfill
\begin{minipage}[t]{0.48\textwidth}
\begin{algorithm}[H]
\caption{Unordered Semantic Tree Edit Distance (\textsc{US-TED})}
\label{alg:us_ted}
\begin{algorithmic}
   \STATE {\bfseries Input:} Tree nodes $u, v$; Embedder $E$; Costs $c_{ins}, c_{del}$
   \STATE {\bfseries Procedure:} $\text{TED}(u, v)$
   \STATE $cost_{ren} \gets 1 - \max(0, \cos(E(u), E(v)))$
   \IF{$u, v$ are leaves}
       \STATE \textbf{Return} $cost_{ren}$
   \ENDIF
   \STATE Let $\{u_1, \dots, u_n\} = \text{Ch}(u)$ and $\{v_1, \dots, v_m\} = \text{Ch}(v)$
   \STATE Construct cost matrix $\mathbf{C} \in \mathbb{R}^{N \times N}$ where $N = \max(n, m)$:
   \STATE $\mathbf{C}_{i,j} \gets 
   \begin{cases} 
   \text{TED}(u_i, v_j) & \text{if } i \le n, j \le m \\
   |T_{u_i}| \cdot c_{del} & \text{if } i \le n, j > m \\
   |T_{v_j}| \cdot c_{ins} & \text{if } i > n, j \le m \\
   0 & \text{otherwise}
   \end{cases}$
   \STATE \textbf{Return} $cost_{ren} + \min \sum_{i=1}^N \mathbf{C}_{i, \sigma(i)}$ \COMMENT{Hungarian Algorithm}
\end{algorithmic}
\end{algorithm}
\end{minipage}

\FloatBarrier

\section{Metric Calibration: Floors, Ceilings, and Depth-Matched \textsc{Sem-Path}}
\label{app:metric_calibration}

A raw \textsc{Sem-Path} value is uninterpretable without knowing what a degenerate taxonomy scores. Unlike \textsc{ARI}, \textsc{Sem-Path} is \emph{not} chance-corrected, so we calibrate it directly by scoring taxonomies whose quality we know by construction. Table~\ref{tab:metric_calibration} reports the result, and it changes how Finding~5 must be read.

\begin{table*}[ht]
\centering
\small
\setlength{\tabcolsep}{5pt}
\begin{tabular}{lcl}
\toprule
Taxonomy & \textsc{Sem-Path} & Construction \\
\midrule
\textsc{Singleton} & 26.00\% & root $\to$ one category per paper \\
\textsc{Flat}      & 27.49\% & root $\to$ one category $\to$ every paper \\
\textsc{Random}    & 28.11\% & expert labels rewired into a random 3-level tree \\
\rowcolor{gray!12}
\emph{12 configs, prev.\ gen.} & \emph{28.13--29.16\%} & \emph{Bottom-Up, Title+Abstract} \\
\rowcolor{gray!12}
\emph{16 configs, current gen.} & \emph{28.12--30.03\%} & \emph{Bottom-Up, one generation later} \\
\midrule
\textsc{Truncated} & 47.30\% & expert tree collapsed to 3 levels, placement \emph{correct} \\
\rowcolor{gray!12}
\emph{Human baseline} & \emph{53.91\%} & \emph{two annotators, matched 10 surveys; 52.62 and 55.20\%} \\
\textsc{Shuffle}   & 49.45\% & expert tree, paper placement permuted at random \\
\textsc{Identity}  & 100.00\% & expert tree against itself \\
\bottomrule
\end{tabular}
\caption{\textbf{\textsc{Sem-Path} calibration.} \textsc{Random} and \textsc{Shuffle} are averaged over 3 seeds ($\pm 0.02$ and $\pm 0.27$\,pp). Two facts follow. (i)~The model bands sit 0.6--2.5\,pp above \textsc{Flat}, a taxonomy performing no organization at all. (ii)~Humans given the same Bottom-Up task clear the floor by 26.42\,pp (App.~\ref{app:human_baseline}), while \textsc{Shuffle} destroys placement yet still scores 49.45\%, just below the 50.56\% at which two independent annotators agree with each other.}
\label{tab:metric_calibration}
\end{table*}

\paragraph{What the floors imply for Finding~5.}
The convergence of all 12 previous-generation configurations into a 1\,pp band is largely a floor effect rather than a discovered property of models: in this shape regime \textsc{Sem-Path} assigns 26--28\% to essentially any taxonomy, so a band starting at 28.13\% carries little information. We therefore do not treat the narrowness of the band as evidence of shared organizational bias.

\paragraph{A natural experiment confirms the floor without synthetic baselines.}
An objection to the table above is that \textsc{Singleton}, \textsc{Flat}, \textsc{Random} and
\textsc{Shuffle} are constructions, so their scores might say more about our constructions than
about the metric. App.~\ref{app:prevgen} supplies the test that avoids this: re-running the
benchmark on the newest model generation raises best ARI by 3.68\,pp (31.24\%$\to$34.92\%) while
raw \textsc{Sem-Path} moves 0.87\,pp (29.16\%$\to$30.03\%), remaining within 2.54\,pp of
\textsc{Flat}. A leaf-partition alignment change large enough to be obvious on a chance-corrected
metric is therefore almost invisible to raw \textsc{Sem-Path}. This diagnoses sensitivity across
runs; it is not an absolute organization-quality claim from ARI. We regard it as strong evidence
because nothing about the comparison was designed by us.

\paragraph{What the ceilings imply for the human baseline.}
\textsc{Shuffle} retains expert labels and topology but randomizes the paper assigned to each leaf. It therefore removes the placement information that \textsc{Sem-Path} is intended to reward. Nevertheless, it scores 49.45\%, just below the 50.56\% at which two independent annotators agree with each other. Raw \textsc{Sem-Path} is therefore still sensitive to label and shape similarity. App.~\ref{app:human_baseline} applies the depth-matched control to the human trees: the human--model gap shrinks from $+$24.64\,pp to $+$13.27\,pp but remains large on every matched survey, so shape alone does not explain the advantage.

\paragraph{Depth-matched \textsc{Sem-Path}.}
The natural control is to score each model against a version of the expert tree collapsed to the model's own depth, so that the unmatched-ancestor penalty is not charged for a granularity choice. For the previous generation, whose trees are three levels deep in the median, we collapse to three levels; for the current generation we collapse to the depth of each predicted tree. Table~\ref{tab:depth_matched} reports the previous generation. Four things change.

\emph{Scores move well clear of the floor.} Models rise from 28.5--29.0\% to 38.3--42.3\%, against a \textsc{Flat} floor of 27.49\%.

\emph{The matched human gap remains.} On the same 10 surveys, humans average 54.09\% versus 38.71\% pooled across all 16 model configurations. Against the ARI-selected Kimi-K3 comparator (40.81\%), the paired advantage is $+$13.27\,pp (95\% CI $[10.69,16.06]$; humans higher on 10/10). This known-group separation is our primary evidence that the variant retains meaningful organization headroom after controlling for depth.

\emph{Across-model variation is restored.} The spread across the eight configurations widens from 0.49\,pp to 4.00\,pp, and the resulting order (Qwen3-Max $>$ Gemini-3-Pro $\approx$ GPT-5 $>$ Claude-Sonnet-4.5) matches the \textsc{ARI} ordering, which raw \textsc{Sem-Path} did not. App.~\ref{app:human_baseline} shows that ARI is saturated relative to human reproducibility, so this agreement cannot establish an absolute organization-quality gap. It indicates that the depth-matched variant recovers variation in leaf-partition alignment, not only tree shape. Because the first ordering was observed on the same eight configurations used to motivate the control, we re-test it on the 16 newest-generation configurations of App.~\ref{subsec:reasoning_control}, which played no part in its design: correlation with ARI rises from $r{=}0.294$ ($p{=}0.27$, not significant) to $r{=}0.813$ ($p{=}1{\times}10^{-4}$), and the spread from 1.91 to 9.45\,pp. Rank correlation improves more modestly ($\rho{=}0.368\to0.476$, $p{=}0.06$), so the relationship is strong in magnitude while the exact ordering still shifts locally.

\emph{Reference frames.} \textsc{Truncated} (47.30\%) is scored against the \emph{full} expert tree, whereas depth-matched model scores use the \emph{collapsed} tree, so the two are not directly comparable; in the depth-matched frame the ceiling is 100\% by construction.

To calibrate the variant properly, we recompute the floors inside its own frame in
Table~\ref{tab:depth_matched_floors}. We collapse the expert tree to three levels and randomize the paper assigned
to each leaf, which destroys placement information while retaining expert labels and topology.
This baseline still scores \textbf{77.29\%}. Placement error alone is therefore worth at most about
23\,pp, while models sit at 38--42\%, some 35\,pp \emph{below} even a randomly-placed
expert-labeled tree. The dominant term in the model deficit is thus divergence of category
\emph{vocabulary and topology}, not misplacement of papers: models do not put the right papers in
the wrong expert categories, they build a different set of categories. This is consistent with the
raw-frame pathology that \textsc{Shuffle} (49.43\% in our recomputation, 49.45\% as reported)
exceeds \textsc{Truncated}, and it means \textsc{Sem-Path} remains label-dominated even after
depth-matching. We therefore make no quantitative placement-error claim, and we flag separating
label divergence from placement error as a second open problem alongside chance correction.

\begin{table*}[ht]
\centering
\small
\setlength{\tabcolsep}{6pt}
\begin{tabular}{llc}
\toprule
Reference frame & Candidate & \textsc{Sem-Path} \\
\midrule
\multirow{2}{*}{Full expert tree}
 & collapsed to 3 levels, placement \emph{correct} (\textsc{Truncated}) & 47.30\% \\
 & full tree, placement \emph{random} (\textsc{Shuffle})                & 49.43\% \\
\midrule
\multirow{3}{*}{Collapsed expert tree}
 & itself (definitional ceiling)                          & 100.00\% \\
 & itself with placement \emph{random}                     & \textbf{77.29\%} \\
 & \emph{model trees} (Table~\ref{tab:depth_matched})      & \emph{38--42\%} \\
\bottomrule
\end{tabular}
\caption{\textbf{Calibrating the depth-matched variant in its own reference frame}, with the expert
tree collapsed to three levels, averaged over 3 seeds and all 72 surveys. The top block reproduces the published raw-frame values and confirms the
implementation. The bottom block tests the placement claim directly. Randomizing placement while keeping expert labels costs only 100\%$\to$77.29\%, yet models
score 38--42\%, so the model deficit is dominated by label and topology divergence rather than by
placement.}
\label{tab:depth_matched_floors}
\end{table*}

\begin{table*}[ht]
\centering
\small
\setlength{\tabcolsep}{6pt}
\begin{tabular}{lccc}
\toprule
Model & vs.\ full expert & vs.\ depth-matched expert & $\Delta$ \\
\midrule
\modelname{claude-color}{Claude-Sonnet-4.5} & 28.86\% & 38.34\% & $+9.49$ \\
\modelname{claude-color}{Claude-Sonnet-4.5-Thinking} & 28.51\% & 38.33\% & $+9.82$ \\
\modelname{openai}{GPT-5} & 28.92\% & 41.09\% & $+12.17$ \\
\modelname{openai}{GPT-5-Thinking} & 28.99\% & 40.61\% & $+11.61$ \\
\modelname{gemini-color}{Gemini-3-Pro} & 28.77\% & 41.10\% & $+12.33$ \\
\modelname{gemini-color}{Gemini-3-Pro-Thinking} & 28.80\% & 40.87\% & $+12.06$ \\
\modelname{qwen-color}{Qwen3-Max} & 28.91\% & \textbf{42.32\%} & $+13.41$ \\
\modelname{qwen-color}{Qwen3-Max-Thinking} & 28.58\% & 41.66\% & $+13.07$ \\
\midrule
\textbf{Spread across models} & \textbf{0.49\,pp} & \textbf{4.00\,pp} & \\
\midrule
\rowcolor{tintRef}
\multicolumn{2}{l}{\emph{Ceiling} in this frame (collapsed tree vs.\ itself)} & \emph{100.00\%} & \\
\rowcolor{tintRef}
\multicolumn{2}{l}{\emph{Placement-random floor} (expert labels, papers permuted)} & \emph{77.29\%} & \\
\bottomrule
\end{tabular}
\caption{\textbf{Depth-matched \textsc{Sem-Path}} on the previous generation (Bottom-Up, Title+Abstract; the 8 configurations with retained tree outputs covering 66--71 of the 72 surveys). Each model is additionally scored against the expert tree collapsed to 3 category levels, which removes the unmatched-ancestor penalty arising purely from depth. The two reference rows are computed in the same collapsed frame as the model column; models fall well below the placement-random floor, so the shortfall is not attributable to placement (Table~\ref{tab:depth_matched_floors}).}
\label{tab:depth_matched}
\end{table*}

\paragraph{Recommendation.}
We report raw \textsc{Sem-Path} for comparability with Table~\ref{tab:org_bottom_up_mode} but recommend that future work using \textsc{TaxoBench} report the depth-matched variant alongside it, together with the \textsc{Flat} floor for the survey set in question. The underlying issue is that \textsc{Sem-Path} lacks the chance correction that makes \textsc{ARI} interpretable; a properly chance-corrected hierarchy metric is the natural next step, and we regard it as the most valuable methodological extension of this work.

\section{Construct Validity Decomposition}
\label{app:construct_validity}

Table~\ref{tab:construct_validity_decomposition} maps each numerical claim in the paper to the expert information it uses (Section~\ref{subsec:alignment_vs_capability}).

\begin{table}[ht]
\centering
\small
\setlength{\tabcolsep}{4pt}
\begin{tabular}{p{0.40\linewidth}p{0.19\linewidth}p{0.29\linewidth}}
\toprule
Finding / Number & Level & What it consumes \\
\midrule
Best Recall 20.92\% (F1) & L3 & $\mathcal{P}^*$ only; no taxonomy \\
\midrule
Depth deficit 0.89--1.87 levels (Finding 6) & L2 & Expert depth (a scalar); no judge \\
Branching 3.16 vs.\ 2.52 & L2 & Expert branching (a scalar) \\
Over-segmentation, \#cat.\ and singletons & L2 & Expert category count (a scalar) \\
\midrule
Best ARI 34.92\% (\emph{saturated}; alignment only) & L1 & Full expert assignment $U^*$ \\
\textsc{Sem-Path} band 28--30\% (Finding 5) & L1 & Full expert hierarchy $T_h^*$ \\
\textsc{US-TED}/\textsc{US-NTED} rankings & L1 & Full expert hierarchy $T_h^*$ \\
Human baseline 53.91\% \textsc{Sem-Path} (10 surveys) & L1 & Same reference as models \\
Constrained-$K$ +7.49\,pp ARI & L1 & Full expert assignment $U^*$ \\
\midrule
\quad Sibling overlap 75.9\% & \emph{unusable} & Judge; expert scores 80.6\% \\
\quad Imbalance 83.4\%   & \emph{weak} & Judge; rubric cites reference tree \\
\quad MECE 51.2\%, criteria 66.0\% & \emph{weak} & Judge; no expert control \\
\quad Missing branches 77.5\% & \emph{weak} & Judge + domain knowledge \\
\bottomrule
\end{tabular}
\caption{\textbf{Expert information used by each finding} (L1--L3, Section~\ref{subsec:alignment_vs_capability}). L3 uses the expert paper set but no taxonomy; L2 uses only scalar shape statistics of the expert hierarchy; L1 uses the hierarchy or assignment in full. ARI is marked saturated because models exceed human--human agreement on it. The LLM-judged defect rates at the bottom have no working expert control (App.~\ref{app:reference_independent}).}
\label{tab:construct_validity_decomposition}
\end{table}

\section{Reference-Light Structural Evidence}
\label{app:reference_independent}

This appendix reports the reference-light evidence behind Finding~6. We separate two sources of evidence: \emph{structural statistics}, which are computed directly from trees and require no judge, and \emph{LLM-judged defect rates}, which are rubric scores produced by GPT-4o. We run the expert taxonomies through the same procedures as a control.

\paragraph{Structural statistics (no judge, no threshold).}
Table~\ref{tab:expert_paired_structure} compares the 72 expert taxonomies against model taxonomies on four statistics computed directly from the trees. Hierarchy depth counts category levels including the root and excluding paper nodes, matching the convention in Table~\ref{tab:stats}. One pattern is universal and one is not.

\emph{Depth is the most durable deficit.} We pool \textbf{70 Bottom-Up runs}
across four input granularities, two model generations, and both reasoning settings. They give a
depth of 2.99--3.97 levels against the expert's 4.86, a deficit of \textbf{0.89--1.87 levels with
zero exceptions}; per-paper ancestor chains are short by 1.18--1.48 labels. App.~\ref{app:depth_probe} shows that the deficit is not a capability ceiling and that removing it does not improve alignment.

We derive the metric consequence directly. For an aligned paper $d$ with model and expert ancestor-chain lengths $p$ and $q$, $J_d=\mathrm{dp}[p,q]+\lambda|q-p|\ge\lambda|q-p|$ because the alignment cost $\mathrm{dp}$ is non-negative. Hence
\begin{equation}
\begin{aligned}
\textsc{Sem-Path}\;&\le\;\frac{1}{|D_a|}\sum_{d\in D_a}\frac{1}{1+\lambda\,|q_d-p_d|}\\
&=:\;\textsc{Ceiling},
\end{aligned}
\label{eq:depth_ceiling}
\end{equation}
averaged \emph{per paper}; because $1/(1+x)$ is convex, plugging in the mean deficit would understate the ceiling by 6\,pp on our data.

We evaluate Equation~\ref{eq:depth_ceiling} over the 3{,}371--3{,}757 papers aligned per model and obtain a ceiling of \textbf{45.8--52.4\%}, with mean 48.6\%. The depth deficit alone costs models roughly 51\,pp of \textsc{Sem-Path}. Label semantics cost another 20\,pp because models score 28--29\% rather than the 48.6\% attainable with perfect label matching at their current depth. Depth therefore explains a large but incomplete part of the gap. Human annotators under the same Bottom-Up requirements score 53.91\% \textsc{Sem-Path} at depth 4.65 on a matched 10-survey subset in App.~\ref{app:human_baseline}, above the shallow-tree ceiling that models can attain even with perfect label matching.

\emph{Over-segmentation is model-specific.} Across the 70 pooled runs category counts span 11.0--28.3 around the expert's 14.0
and singleton rates 4.9--50.4\% around the expert's 17.0\%. In the previous generation
Claude-Sonnet-4.5 and the two GPT-5 variants emit 20.7--23.1 categories with 33--36\% singletons
while Gemini-3-Pro and Qwen3-Max match the expert count closely; in the current generation the
same split appears between GPT-5.6-sol (28.3 categories, 43.5\% singletons) and Gemini-3.1-Pro
(11.0, 5.1\%). Some models therefore over-segment by more than $2\times$ the expert rate while
others under-segment.

\begin{table*}[ht]
\centering
\small
\setlength{\tabcolsep}{6pt}
\begin{tabular}{lcccccc}
\toprule
Taxonomy source & Depth & Chain & Deficit & Ceiling & \#\,Cat. & Singleton \% \\
\midrule
\textbf{Expert reference} & \textbf{4.86} & \textbf{4.46} & --- & --- & \textbf{14.0} & \textbf{17.0} \\
\midrule
\modelname{claude-color}{Claude-Sonnet-4.5} & 3.62 & 3.28 & 1.18 & 52.4\% & 23.1 & 32.7 \\
\modelname{claude-color}{Claude-Sonnet-4.5-Thinking} & 3.58 & 3.22 & 1.24 & 51.5\% & 18.9 & 21.0 \\
\modelname{openai}{GPT-5} & 3.23 & 3.09 & 1.37 & 49.2\% & 20.7 & 35.9 \\
\modelname{openai}{GPT-5-Thinking} & 3.24 & 3.13 & 1.33 & 49.2\% & 21.2 & 32.9 \\
\modelname{gemini-color}{Gemini-3-Pro} & 3.04 & 3.00 & 1.46 & 47.0\% & 13.8 & \ \,7.1 \\
\modelname{gemini-color}{Gemini-3-Pro-Thinking} & 3.06 & 3.02 & 1.44 & 47.5\% & 13.7 & \ \,7.3 \\
\modelname{qwen-color}{Qwen3-Max} & 3.01 & 2.98 & 1.48 & 46.6\% & 12.9 & 12.0 \\
\modelname{qwen-color}{Qwen3-Max-Thinking} & 3.03 & 2.99 & 1.47 & 45.8\% & 13.3 & 12.3 \\
\midrule
Model mean                 & 3.23 & 3.09 & 1.37 & \textbf{48.6\%} & 17.2 & 20.1 \\
\textbf{$\Delta$ (model $-$ expert)} & \textbf{$-1.63$} & \textbf{$-1.37$} & --- & --- & $+3.2$ & $+3.2$ \\
\bottomrule
\end{tabular}
\caption{\textbf{Expert-paired structural statistics and the induced \textsc{Sem-Path} ceiling} (Bottom-Up, Title+Abstract; the 8 configurations whose tree outputs were fully retained). Nothing here uses an LLM judge or a tunable threshold. \textbf{Depth} counts category levels including the root and excluding paper nodes, matching Table~\ref{tab:stats}; \textbf{Chain} is the mean per-paper ancestor-chain length, which is the quantity \textsc{Sem-Path} actually operates on; \textbf{Deficit} is the mean of $q_d-p_d$ over aligned papers; \textbf{Ceiling} is Equation~\ref{eq:depth_ceiling}, averaged per paper. A singleton is a paper category holding exactly one paper. The depth deficit holds for every configuration without exception; the category-count and singleton gaps hold only for Claude-Sonnet-4.5 and the two GPT-5 variants.}
\label{tab:expert_paired_structure}
\end{table*}

\paragraph{Depth under a category-count constraint.}
The Constrained-$K$ probe in App.~\ref{app:constrained_k} fixes the number of paper categories. Claude-Sonnet-4.5 drops from 19.8 to 13.6 categories and GPT-5 from 17.4 to 10.8, bracketing the expert value of 14.0, yet depth stays far from 4.86: it falls from 3.40 to 3.10 for Claude-Sonnet-4.5 and from 3.24 to 3.12 for GPT-5 on paired sets of 20 and 17 surveys. App.~\ref{app:depth_probe} varies depth directly.

\paragraph{Shallow trees are also wide.}
Model trees are correspondingly \emph{wider}: mean branching factor is 3.16 across models against 2.52 for expert taxonomies, so each parent carries more siblings and mutual exclusivity is harder to maintain. A tree with 17.2 leaf categories over 3.2 levels also performs little progressive refinement, whereas the expert's 4.86 levels encode a chain of successive distinctions.

\paragraph{LLM-judged defect rates, with an expert control.}
Table~\ref{tab:reference_independent_defects} reports five defect rates over 1{,}000 model taxonomies together with the rate the same procedure assigns to the 72 expert taxonomies. All five come from the GPT-4o rubric of App.~\ref{sec:prompt_judge}, whose imbalance criterion is phrased relative to the reference tree. Expert taxonomies show comparable sibling overlap and imbalance under an embedding re-implementation of the same criteria, because sibling categories in a real survey are semantically related, so these two rates do not separate model trees from expert trees.

\begin{table}[ht]
\centering
\small
\setlength{\tabcolsep}{4pt}
\begin{tabular}{p{0.30\linewidth}ccp{0.30\linewidth}}
\toprule
Defect Category & Model & Expert & Basis \\
\midrule
Semantic overlap between siblings & 75.9\% & 80.6\% & GPT-4o rubric; cosine re-impl.\ at $\tau{=}0.6$ \\
Structural imbalance & 83.4\% & 63.9\% & GPT-4o rubric (criterion cites reference tree) \\
MECE violations & 51.2\% & --- & GPT-4o rubric \\
Inconsistent classification criteria & 66.0\% & --- & GPT-4o rubric \\
Missing core branches & 77.5\% & --- & GPT-4o rubric + domain knowledge \\
\bottomrule
\end{tabular}
\caption{\textbf{LLM-judged defect rates with an expert control.} All rates come from the GPT-4o rubric, which reaches $\kappa=0.89$ against human raters on the 5-point scale (App.~\ref{app:judge_co_with_human}). The \textbf{Expert} column applies an embedding re-implementation of the same criterion to the 72 expert taxonomies. Expert trees show higher sibling overlap than model trees, so this category does not discriminate. A dash marks criteria that we could not re-implement without a judge.}
\label{tab:reference_independent_defects}
\end{table}

We retain the 1{,}000 annotated model taxonomies in the private reproducibility archive. The public repository releases the dataset and scoring code but excludes raw model outputs and product logs.

\subsection{From Structural Defects to Downstream Survey Quality}
\label{app:downstream_utility}

The taxonomy is the section outline of a generated survey, so its defects carry over to the text. Overlapping sibling categories force the same subtopic to be narrated under several headings, and leaf categories that hold a single paper produce non-generalizing micro-headings that make the field hard to navigate. A reader study with taxonomies of controlled quality would measure this effect directly.

\section{\texorpdfstring{Intersection-Conditioned ARI$_\cap$ and V-Measure$_\cap$}{Intersection-Conditioned ARI-cap and V-Measure-cap}}
\label{app:ari_intersection}

This appendix specifies the intersection-conditioned metrics used in
Table~\ref{tab:org_deep_research_organization}. In Deep Research mode, an agent organizes only the
papers it retrieves, so we compute a conditioned view on
$\mathcal{P}^*\cap\hat{\mathcal{P}}$ in addition to the end-to-end view that assigns every
unretrieved expert paper to a sink label. We align expert and retrieved papers with the deterministic
title-normalization rules of App.~\ref{sec:metric_paper_alignment}, then compute ARI$_\cap$ and
V-Measure$_\cap$ on the aligned subset.

ARI is degenerate for a single point, so the intersection-conditioned columns are averaged only over surveys with at least two aligned papers, which leaves 29--67 surveys per agent. Intersection size also matters: Doubao averages only 1.7 retrieved expert papers and obtains the highest ARI$_\cap$, whereas o3 averages 11.1 and scores lower, so the table reports $|\mathcal{P}^*\cap\hat{\mathcal{P}}|$ alongside ARI$_\cap$.

\section{Depth-Controlled Probe}
\label{app:depth_probe}

Our generation prompts contain a three-level worked example. We vary the depth cue to test whether it drives the shallow output of Finding~6.

\paragraph{Design.}
We re-run Bottom-Up under three conditions that differ \emph{only} in the depth cue; the
prompt is edited in place on the released input, so every other byte is identical.

\begin{itemize}[topsep=2pt,itemsep=1pt,leftmargin=*]
\item \textbf{A (control)}: the shipped prompt, three-level worked example.
\item \textbf{B (exemplar)}: same prompt, worked example replaced by a five-level tree.
\item \textbf{C (instruction)}: three-level example plus an explicit instruction to build about
five levels.
\end{itemize}

Four models $\times$ 3 conditions $\times$ all 72 surveys $=$ 864 calls. Condition~A, the shipped
prompt, yields 3.01--3.41 levels, in line with the original runs of these models, confirming the apparatus.

\begin{table*}[ht]
\centering
\small
\setlength{\tabcolsep}{5pt}
\begin{tabular}{lccccccc}
\toprule
Model & Cond. & Depth & \#Cat & Singleton & \textsc{Sem-Path} & ARI & Hom. \\
\midrule
\modelname{openai}{GPT-5} & A & 3.18 & 19.5 & 32.7\% & 28.67\% & 29.02\% & 78.30\% \\
\modelname{openai}{GPT-5} & B & 3.38 & 20.4 & 33.9\% & 28.96\% & 29.30\% & 79.54\% \\
\modelname{openai}{GPT-5} & \textbf{C} & \textbf{4.86} & \textbf{35.1} & \textbf{66.7\%} & \textbf{25.90\%} & \textbf{14.65\%} & 91.55\% \\
\midrule
\modelname{claude-color}{Claude-Sonnet-4.5} & A & 3.25 & 19.2 & 24.4\% & 29.05\% & 30.39\% & 79.65\% \\
\modelname{claude-color}{Claude-Sonnet-4.5} & B & 4.01 & 27.2 & 41.2\% & 29.17\% & 23.08\% & 87.15\% \\
\modelname{claude-color}{Claude-Sonnet-4.5} & C & 4.08 & 28.6 & 44.6\% & 28.81\% & 20.12\% & 87.79\% \\
\midrule
\modelname{deepseek-color}{DeepSeek-V3.2} & A & 3.41 & 16.7 & 19.1\% & 28.70\% & 27.45\% & 73.64\% \\
\modelname{deepseek-color}{DeepSeek-V3.2} & B & 3.90 & 21.1 & 27.5\% & 28.38\% & 23.34\% & 79.26\% \\
\modelname{deepseek-color}{DeepSeek-V3.2} & C & 4.00 & 20.2 & 24.4\% & 28.19\% & 24.57\% & 78.70\% \\
\midrule
\modelname{qwen-color}{Qwen3-Max} & A & 3.01 & 14.5 & 15.7\% & 29.01\% & 30.79\% & 71.70\% \\
\modelname{qwen-color}{Qwen3-Max} & B & 3.18 & 15.2 & 16.7\% & 29.30\% & 30.04\% & 72.40\% \\
\modelname{qwen-color}{Qwen3-Max} & C & 3.11 & 13.5 & 9.7\% & 29.24\% & 29.63\% & 70.45\% \\
\midrule
\emph{Expert reference} & --- & \emph{4.86} & \emph{14.0} & \emph{17.0\%} & --- & --- & --- \\
\bottomrule
\end{tabular}
\caption{\textbf{Depth-controlled probe} with 864 calls and 857 parseable trees. Condition~C makes
GPT-5 match the expert's depth exactly but costs it 2.77\,pp \textsc{Sem-Path} and half its
ARI, because the extra levels are produced by fragmentation. Rising homogeneity under~C
reflects splitting (smaller clusters are purer).}
\label{tab:depth_probe}
\end{table*}

\paragraph{Result 1: exemplar shape matters little; explicit instruction works.}
Table~\ref{tab:depth_probe} reports the three conditions. Swapping in a five-level exemplar (B) moves depth by only $+0.17$ to $+0.76$ levels. An
explicit target (C) is far more effective: GPT-5 reaches 4.86, exactly the expert mean, with
Claude at 4.08 and DeepSeek at 4.00. Three of four models can therefore build five-level
taxonomies and simply do not do so unprompted. Qwen3-Max is the exception, barely
responding to either manipulation ($3.01\to3.11$); for that model the shallow output does look
like a limit.

\paragraph{Result 2: depth compliance does not buy expert-like structure.}
At matched depth GPT-5's category count rises from 19.5 to 35.1
($2.5\times$ the expert's 14.0) and its singleton-leaf rate doubles from 32.7\% to 66.7\%
(expert: 17.0\%), while \textsc{Sem-Path} \emph{falls} 2.77\,pp and ARI halves. Claude and
DeepSeek move the same direction. Across all 12 model--condition pairs, none improves
\textsc{Sem-Path} by more than 0.3\,pp.

\paragraph{Implication.}
Models can increase depth when instructed, but alignment does not improve: \textsc{Sem-Path} never gains more than 0.3\,pp, and ARI falls in every pair whose depth rises by at least 0.4 levels, because the added levels fragment the taxonomy rather than refine it. The depth-based bound on \textsc{Sem-Path} (Equation~\ref{eq:depth_ceiling}) is derived under the standard prompt, where the deficit is universal.

\section{Decomposition Analysis and Mechanistic Probe}
\label{app:decomposition_analysis}

\subsection{Decomposition by Depth and Size}
\label{app:decomposition}

We decompose Bottom-Up results by paper count and hierarchy depth in Table~\ref{tab:decomposition_depth_size} to separate scale from deep hierarchical reasoning. \textsc{US-TED} grows monotonically with paper count, so larger paper sets increase structural divergence. \textsc{Sem-Path} falls sharply with depth. For Qwen3-Max, it drops from $43.5$ to $34.7$ to $25.6\%$ as depth increases from 3 to at least 5. This pattern localizes the bottleneck to deep hierarchy rather than shallow categorization.

\begin{table}[ht]
\centering
\small
\setlength{\tabcolsep}{3pt}
\begin{tabular}{lcccccc}
\toprule
\multirow{2}{*}{Model} & \multicolumn{3}{c}{\textsc{US-TED} ($\downarrow$)} & \multicolumn{3}{c}{\textsc{Sem-Path} ($\uparrow$)} \\
\cmidrule(lr){2-4}\cmidrule(lr){5-7}
 & $<40$ & 40--80 & $>80$ & $d{=}3$ & $d{=}4$ & $d{\ge}5$ \\
\midrule
\modelname{claude-color}{Claude-Sonnet-4.5} & 30.3 & 45.9 & 59.3 & 37.5 & 35.2 & 25.8 \\
\modelname{openai}{GPT-5} & 28.7 & 44.8 & 50.5 & 41.9 & 33.8 & 26.0 \\
\modelname{gemini-color}{Gemini-3} & 24.2 & 34.7 & 40.0 & 39.8 & 33.6 & 25.0 \\
\modelname{qwen-color}{Qwen3-Max} & 25.5 & 35.3 & 37.9 & 43.5 & 34.7 & 25.6 \\
\modelname{kimi-color}{Kimi-K2} & 32.5 & 44.8 & 44.8 & 42.8 & 33.5 & 25.2 \\
\bottomrule
\end{tabular}
\caption{\textbf{Depth and size decomposition} (Bottom-Up). Each cell is the average over the corresponding bin.}
\label{tab:decomposition_depth_size}
\end{table}

\subsection{\texorpdfstring{Constrained-$K$}{Constrained-K} Mechanistic Probe}
\label{app:constrained_k}

To causally test whether granularity selection drives the over-segmentation pattern observed in Finding~4, we ran a Constrained-$K$ experiment on 15 sampled surveys: GPT-5 was prompted to produce exactly $K$ paper categories (where $K$ is the expert category count). This isolates granularity from semantic-assignment ability.

Two effects are observed (Table~\ref{tab:constrained_k}):
(i)~ARI improves substantially (13.05\% $\to$ 20.54\%, +7.49\,pp), confirming that granularity selection is a major source of error;
(ii)~however, ARI plateaus at 20.54\%, far below the human ceiling, and homogeneity \emph{drops} (73.84\% $\to$ 62.87\%), revealing that the previously high homogeneity was largely the by-product of over-segmentation rather than genuine semantic understanding.

\begin{table}[ht]
\centering
\small
\setlength{\tabcolsep}{6pt}
\begin{tabular}{lcc}
\toprule
Metric & Free-$K$ & Constrained-$K$ \\
\midrule
ARI            & 13.05\% & \textbf{20.54\%} (+7.49\,pp) \\
Homogeneity    & 73.84\% & 62.87\% ($-$10.97\,pp) \\
\bottomrule
\end{tabular}
\caption{\textbf{Constrained-$K$ probe} (15 surveys, GPT-5). Forcing the model to use exactly $K$ paper categories (the expert count) raises ARI by 7.49\,pp but lowers homogeneity, showing both granularity and semantic assignment as bottlenecks.}
\label{tab:constrained_k}
\end{table}

We also report the average paper-category counts produced under the three input settings (Bottom-Up): richer input causes models to produce more fine-grained categories, amplifying over-segmentation (Table~\ref{tab:avg_categories_per_input}).

\begin{table*}[ht]
\centering
\small
\setlength{\tabcolsep}{6pt}
\begin{tabular}{lccc}
\toprule
Model & Title+Abs & +Core-task & +Summary \\
\midrule
\modelname{claude-color}{Claude-Sonnet-4.5} & 22.1 & 24.4 & 25.3 \\
\modelname{deepseek-color}{DeepSeek-V3.2} & 16.9 & 23.7 & 21.1 \\
\modelname{gemini-color}{Gemini-3-Pro} & 12.8 & 14.3 & 14.0 \\
\bottomrule
\end{tabular}
\caption{\textbf{Average paper categories} per input setting. Richer input increases the number of fine-grained categories, consistent with the over-segmentation explanation.}
\label{tab:avg_categories_per_input}
\end{table*}

\subsection{Metric--Human Correlation}
\label{app:metric_human_correlation}

We compute the Spearman correlation between \textsc{Sem-Path} and the LLM-as-Judge scores, which are themselves validated against human evaluators with Cohen's $\kappa = 0.89$; details are in App.~\ref{app:judge_co_with_human}. In Deep Research mode, where the 7 agents exhibit meaningful variance in performance, \textsc{Sem-Path} is strongly correlated with the average judge score (Table~\ref{tab:metric_human_correlation}). In Bottom-Up mode, all 12 models converge into a 1\,pp band on \textsc{Sem-Path}, so rank-based correlations become unstable with only 6 unique data points.

\begin{table}[ht]
\centering
\small
\setlength{\tabcolsep}{6pt}
\begin{tabular}{lccccc}
\toprule
 & Cov.\ & Org.\ & Logic & Topo.\ & Avg \\
\midrule
\textsc{Sem-Path} & 0.949 & 0.632 & 0.774 & 0.964 & \textbf{1.000} \\
\bottomrule
\end{tabular}
\caption{\textbf{Spearman correlation} of \textsc{Sem-Path} with LLM-as-Judge scores (Deep Research mode, $n{=}7$).}
\label{tab:metric_human_correlation}
\end{table}

\section{Reasoning-Effort Control}
\label{subsec:reasoning_control}

We run every model in Table~\ref{tab:org_bottom_up_mode} twice, at the lowest and highest
reasoning effort accepted by its endpoint. Because this control is not portable across vendors,
we report in Table~\ref{tab:reasoning_effort} the parameters sent and tokens returned instead of
labeling the rows as thinking and non-thinking.

\begin{table*}[ht]
\centering
\small
\setlength{\tabcolsep}{5pt}
\begin{tabular}{llcccc}
\toprule
\multirow{2}{*}{Model} & \multirow{2}{*}{Endpoint} & \multicolumn{2}{c}{\texttt{reasoning\_effort} sent} & \multicolumn{2}{c}{mean reasoning tokens} \\
\cmidrule(lr){3-4}\cmidrule(lr){5-6}
& & low & high & low & high \\
\midrule
\modelname{claude-color}{Claude-Sonnet-5} & \texttt{claude-sonnet-5}        & \texttt{minimal} & \texttt{high} & n/a & n/a \\
\modelname{openai}{GPT-5.5} & \texttt{gpt-5.5}                & \texttt{none}    & \texttt{high} & 0 & 4{,}091 \\
\modelname{openai}{GPT-5.6-sol} & \texttt{gpt-5.6-sol}            & \texttt{none}    & \texttt{high} & 0 & 2{,}541 \\
\modelname{gemini-color}{Gemini-3.1-Pro} & \texttt{gemini-3.1-pro-preview} & \texttt{minimal} & \texttt{high} & 2{,}717 & 7{,}771 \\
\modelname{deepseek-color}{DeepSeek-V4-Pro} & \texttt{deepseek-v4-pro}        & \texttt{minimal} & \texttt{high} & 10{,}182 & 10{,}164 \\
\modelname{qwen-color}{Qwen3.8-Max} & \texttt{qwen3.8-max-preview}    & \texttt{minimal} & \texttt{high} & 3{,}791 & 12{,}797 \\
\modelname{kimi-color}{Kimi-K3} & \texttt{kimi-k3}                & \texttt{minimal} & \texttt{high} & 1{,}190 & 4{,}187 \\
\modelname{grok}{Grok-4} & \texttt{grok-4-latest}          & \texttt{minimal} & \texttt{high} & 970 & 924 \\
\bottomrule
\end{tabular}
\caption{\textbf{Reasoning-effort settings, as sent and as observed.} The low setting is the
weakest value each endpoint accepts, found by probing: GPT accepts \texttt{none} while the
others reject it and require \texttt{minimal}. \texttt{max\_tokens} is 65{,}536 throughout,
temperature is left at the endpoint default, and the prompt is byte-identical across both
rows. Three cases show why the parameter alone is not evidence of a contrast:
\texttt{minimal} does \emph{not} disable reasoning (Gemini still spends 2{,}717 tokens);
DeepSeek-V4-Pro appears to reason at capacity regardless of the setting; and Claude-Sonnet-5
reports no reasoning tokens at all, so its contrast is unverified.}
\label{tab:reasoning_effort}
\end{table*}

\paragraph{Verifying reasoning settings.}
The previous-generation Gemini-3-Pro pair in App.~\ref{app:prevgen} produced byte-identical outputs on
all 72 surveys because the endpoint accepted a thinking flag and ignored it. That failure is
invisible unless the returned reasoning-token count is recorded, so every call in
Table~\ref{tab:org_bottom_up_mode} logs both the parameter sent and the tokens returned, and we
mark the one configuration (Claude-Sonnet-5) where the endpoint does not report them.

\paragraph{Two practical consequences.}
First, the lowest effort setting does not always disable reasoning. Our settings bracket a range of
reasoning effort and never switch reasoning off.
Second, effort settings are not comparable across vendors. The \texttt{high} setting uses 924
tokens on Grok-4 and 12{,}797 on Qwen3.8-Max, so we compare each model only with itself.

\paragraph{Results at the highest setting.}
Table~\ref{tab:org_bottom_up_mode} reports the lowest-effort rows; Table~\ref{tab:newgen_high}
gives the matching highest-effort rows on the same 72 surveys and the same prompts. Of the 8
models run, 3 gain ARI at the higher setting (Gemini-3.1-Pro $+2.57$, DeepSeek-V4-Pro $+2.09$,
GPT-5.5 $+0.83$) and 5 lose (Grok-4 $-0.17$, GPT-5.6-sol $-0.32$, Qwen3.8-Max $-1.36$,
Kimi-K3 $-1.49$, Claude-Sonnet-5 $-1.65$). Neither the depth deficit nor the \textsc{Sem-Path} floor is affected:
depth stays within 2.99--3.97 and \textsc{Sem-Path} within 28.12--30.03\% across all 16
configurations.

\begin{table*}[ht]
\centering
\small
\setlength{\tabcolsep}{3.5pt}
\adjustbox{max width=\textwidth}{%
\begin{tabular}{lcccccccc}
\toprule
Model (effort \texttt{high}) & ARI$\uparrow$ & Hom.$\uparrow$ & Comp.$\uparrow$ & \textsc{US-NTED}$\downarrow$ & \textsc{Sem-Path}$\uparrow$ & Depth & \#Cat & Singl. \\
\midrule
\modelname{claude-color}{Claude-Sonnet-5$^\dagger$} & \textbf{33.27\%} & 74.99\% & \textbf{69.14\%} & 76.08\% & \textbf{29.51\%} & 3.31 & 16.3 & 22.3\% \\
\modelname{gemini-color}{Gemini-3.1-Pro} & 32.24\% & 70.93\% & 68.30\% & 77.60\% & 28.80\% & 3.04 & 13.3 & 5.6\% \\
\modelname{deepseek-color}{DeepSeek-V4-Pro} & 31.91\% & 76.22\% & 68.16\% & 77.45\% & 28.76\% & 3.28 & 18.3 & 28.2\% \\
\modelname{kimi-color}{Kimi-K3} & 31.73\% & 77.86\% & 67.85\% & 78.75\% & 28.78\% & 3.10 & 17.8 & 14.6\% \\
\modelname{grok}{Grok-4} & 31.27\% & 70.60\% & 68.65\% & 78.31\% & 28.12\% & 3.07 & 13.5 & 11.6\% \\
\modelname{qwen-color}{Qwen3.8-Max} & 28.45\% & 77.13\% & 66.23\% & 78.85\% & 28.66\% & 3.17 & 18.6 & 17.5\% \\
\modelname{openai}{GPT-5.6-sol} & 21.66\% & \textbf{86.18\%} & 64.91\% & \textbf{73.81\%} & 28.74\% & 3.97 & 27.6 & 38.8\% \\
\midrule
\tablegroup{9}{\textit{Within-family control: the preceding GPT tier}} \\
\modelname{openai}{GPT-5.5, effort \texttt{none}} & 28.30\% & 82.69\% & 66.96\% & 79.32\% & 28.96\% & 3.22 & 22.1 & 32.4\% \\
\modelname{openai}{GPT-5.5, effort \texttt{high}} & 29.13\% & 81.56\% & 67.03\% & 77.38\% & 28.77\% & 3.35 & 20.9 & 23.0\% \\
\midrule
\emph{Expert reference} & --- & --- & --- & --- & --- & \emph{4.86} & \emph{14.0} & \emph{17.0\%} \\
\bottomrule
\end{tabular}%
}
\caption{\textbf{Highest reasoning effort}, same prompts and surveys as
Table~\ref{tab:org_bottom_up_mode} ($n{=}72$, except Grok-4 at 71), plus both settings of the
preceding GPT tier. $^\dagger$Claude-Sonnet-5 reports zero reasoning tokens in both settings, so
this is the one pair we could \emph{not} verify as two distinct conditions and it should not be
counted as evidence about reasoning.}
\label{tab:newgen_high}
\end{table*}

\paragraph{A newer tier is not automatically better here.}
GPT-5.5 outscores its successor GPT-5.6-sol by 6.3\,pp ARI (28.30\% vs.\ 21.98\% at lowest
effort) while emitting 22.1 categories against 28.3 and 32.4\% singleton leaves against 43.5\%.
The two tiers are otherwise the same family and the same prompt, so this is a clean illustration
of the mechanism in Finding~6: GPT-5.6-sol builds the deepest trees of any model we tested (3.97
levels at high effort, closest to the expert's 4.86) and is simultaneously the worst on ARI,
because it reaches that depth by fragmenting. Depth alone is not progress.

\section{Previous-Generation Control}
\label{app:prevgen}

Table~\ref{tab:org_bottom_up_mode} reports the newest model in each family. We additionally
evaluate the preceding generation as a longitudinal control of whether the metrics respond to
model improvement.

\begin{table*}[ht]
\centering
\small
\setlength{\tabcolsep}{3.5pt}
\adjustbox{max width=\textwidth}{%
\begin{tabular}{lccccccc}
\toprule
\multirow{2}{*}{Model} & \multicolumn{4}{c}{Leaf-Level} & \multicolumn{3}{c}{Hierarchy-Level} \\
\cmidrule(lr){2-5} \cmidrule(lr){6-8}
& ARI$\uparrow$ & Hom.$\uparrow$ & Comp.$\uparrow$ & V-Meas.$\uparrow$ & \textsc{US-TED}$\downarrow$ & \textsc{US-NTED}$\downarrow$ & \textsc{Sem-Path}$\uparrow$ \\
\midrule
\tablegroup{8}{\textit{Non-thinking-based}} \\
\modelname{claude-color}{Claude-Sonnet-4.5} & 27.25\% & \textbf{82.27\%} & 66.61\% & \textbf{72.84\%} & 42.89 & 77.73\% & \textbf{29.16\%} \\
\modelname{openai}{GPT-5} & 28.17\% & 77.79\% & 66.45\% & 70.99\% & 40.84 & 78.65\% & 28.97\% \\
\modelname{gemini-color}{Gemini-3-Pro*$^\dagger$} & 29.86\% & 68.25\% & 67.15\% & 67.00\% & \textbf{32.32} & 76.35\% & 28.13\% \\
\modelname{deepseek-color}{DeepSeek-V3.2} & 27.15\% & 74.33\% & 66.15\% & 69.28\% & 37.04 & \textbf{75.71\%} & 28.63\% \\
\modelname{qwen-color}{Qwen3-Max*} & \textbf{31.24\%} & 68.48\% & \textbf{68.63\%} & 67.52\% & 32.80 & 78.17\% & 29.00\% \\
\modelname{kimi-color}{Kimi-K2} & 23.69\% & 79.32\% & 64.88\% & 70.48\% & 41.35 & 79.58\% & 28.33\% \\
\midrule
\tablegroup{8}{\textit{Thinking-based}} \\
\modelname{claude-color}{Claude-Sonnet-4.5-Thinking} & 29.28\% & 76.35\% & 66.96\% & 70.58\% & 37.87 & 76.56\% & 28.66\% \\
\modelname{openai}{GPT-5-Thinking} & 25.97\% & \textbf{80.89\%} & 66.05\% & \textbf{72.07\%} & 43.05 & 79.86\% & 28.56\% \\
\modelname{gemini-color}{Gemini-3-Pro-Thinking*$^\dagger$} & 28.84\% & 67.26\% & 67.00\% & 66.45\% & \textbf{32.40} & 76.47\% & 28.21\% \\
\modelname{deepseek-color}{DeepSeek-V3.2-Thinking} & 27.71\% & 70.74\% & 67.14\% & 67.68\% & 34.51 & \textbf{76.02\%} & 28.67\% \\
\modelname{qwen-color}{Qwen3-Max-Thinking*} & \textbf{30.32\%} & 68.70\% & \textbf{68.10\%} & 67.38\% & 33.62 & 78.24\% & \textbf{28.95\%} \\
\modelname{kimi-color}{Kimi-K2-Thinking} & 22.59\% & 78.98\% & 64.64\% & 70.18\% & 44.07 & 80.85\% & 28.21\% \\
\bottomrule
\end{tabular}%
}
\caption{\textbf{Previous generation}, Bottom-Up, Title+Abstract. * denotes Preview versions.
$^\dagger$\textbf{The two Gemini-3-Pro rows are not two distinct conditions}: their outputs are
byte-identical on all 72 surveys because the endpoint silently ignored the thinking flag, so any
difference between these rows is not attributable to reasoning mode. The successor-model
comparison uses per-call token verification (App.~\ref{subsec:reasoning_control}); the three other
families differ on 72/72 surveys.}
\label{tab:prevgen}
\end{table*}

\paragraph{The depth deficit survives a model generation.}
Depth spans 2.99--3.97 in the current generation against the expert's 4.86; the deepest
configuration is still 0.89 levels short and none of the 16 reaches parity. Pooling with the 54
previous-generation runs gives 70 runs across four input granularities and two generations with no
exception. This is the most durable result in the paper: a year of model progress did not move
it.

\paragraph{The \textsc{Sem-Path} floor also persists across generations.}
Best ARI rises from 31.24\% (Qwen3-Max, Table~\ref{tab:prevgen}) to 34.92\% (Claude-Sonnet-5), a
3.68\,pp gain in expert-reference leaf-partition alignment. Over the same comparison raw
\textsc{Sem-Path} moves 0.87\,pp (29.16\%$\to$30.03\%) and stays within 2.54\,pp of the
\textsc{Flat} floor. This independently measured alignment change is thus nearly invisible to the
raw metric. Unlike the
synthetic baselines of App.~\ref{app:metric_calibration}, this is a natural experiment:
nothing about it was designed by us.

\paragraph{Depth-matching recovers model-level variation on held-out configurations.}
Table~\ref{tab:depth_matched} found wider spread and an ARI-consistent ordering using the same
eight configurations that motivated the control. The 16 current-generation configurations test
that sensitivity claim on data which played no part in its design: correlation with ARI
rises from $r{=}0.294$ ($p{=}0.27$, not significant) for raw \textsc{Sem-Path} to $r{=}0.813$
($p{=}1{\times}10^{-4}$) depth-matched, and the spread widens from 1.91 to 9.45\,pp. Rank
correlation improves more modestly ($\rho{=}0.37\to0.48$, $p{=}0.06$), so the agreement is strong
in magnitude but not yet a perfect ordering. Because ARI is saturated relative to human
reproducibility, we treat this as auxiliary evidence of recovered alignment variation, not as
validation of absolute organization quality.

\paragraph{Reasoning effort remains inconsistent, now on verified data.}
Of the 8 models run, 3 gain ARI at the higher effort setting (Gemini-3.1-Pro
$+2.57$, DeepSeek-V4-Pro $+2.09$, GPT-5.5 $+0.83$) and 5 lose (Grok-4 $-0.17$, GPT-5.6-sol
$-0.32$, Qwen3.8-Max $-1.36$, Kimi-K3 $-1.49$, Claude-Sonnet-5 $-1.65$). Gemini-3.1-Pro's two settings
differ in reasoning tokens (2{,}717 vs.\ 7{,}771) and in output. Claude-Sonnet-5 reports zero
reasoning tokens in both modes, so its contrast is \emph{unverified} and
should not be counted as evidence.

\paragraph{Over-segmentation reappears without any prompt manipulation.}
GPT-5.6-sol is at once the deepest configuration (3.97) and the worst on ARI (21.66\%), with 28.3
categories and 43.5\% singleton leaves. It also scores lowest on depth-matched \textsc{Sem-Path}
(32.49\%). Across the 16
configurations, depth correlates with singleton rate at $\rho=0.829$
($p{=}1{\times}10^{-4}$) while predicting nothing about ARI ($\rho=-0.266$, $p{=}0.32$);
singleton rate in turn correlates negatively with ARI ($\rho=-0.482$, $p{=}0.059$).
App.~\ref{app:depth_probe} reaches the same conclusion by forcing depth within a fixed model;
here it emerges from natural variation across models. Two independent designs, one mechanism.

\section{LLM-as-Judge Validation}
\label{app:judge_co_with_human}
We validate GPT-4o as an LLM-as-Judge by measuring its agreement with human expert ratings on the same trees, using Cohen's $\kappa$ as the agreement statistic. Per-dimension and average $\kappa$ values are reported in Table~\ref{tab:human_gpt4o_kappa}.

We sampled 10 trees per model via random sampling without replacement, covering 7 models from Deep Research mode and 12 models from Bottom-Up mode (Title+Abstract input setting), for a total of 190 human-evaluated trees.

\begin{table*}[t]
\centering
\small
\setlength{\tabcolsep}{4pt}
\begin{tabular}{lccccc}
\toprule
Agreement & Cov. & Org. & Log. & Topo. & Avg. \\
\midrule
\modelname{openai}{GPT-4o vs Human} & 0.9295 & 0.8905 & 0.8807 & 0.8627 & 0.8909 \\
\bottomrule
\end{tabular}
\caption{\textbf{Cohen's Kappa coefficient} between human and GPT-4o evaluations.}
\label{tab:human_gpt4o_kappa}
\end{table*}
Human evaluators are domain experts, paid 100 RMB per hour, who follow the same rubric as GPT-4o and independently rate each tree on a 5-point integer scale along four dimensions: Semantic Coverage, Organization Quality, Logical Consistency, and Topological Similarity. Average $\kappa = 0.8909$, which corresponds to almost-perfect agreement under the standard Landis--Koch interpretation, supporting the use of GPT-4o as a reliable LLM-as-Judge for the taxonomy-evaluation task. Table~\ref{tab:llm_judge_only} reports the resulting judge scores for every model in both modes.

\begin{table*}[t]
\centering
\small
\setlength{\tabcolsep}{6pt}
\begin{tabular}{lccccc}
\toprule
\multirow{2}{*}{Model} & \multicolumn{5}{c}{LLM-as-Judge} \\
\cmidrule(lr){2-6}
& Cov.$\uparrow$ & Org.$\uparrow$ & Log.$\uparrow$ & Topo.$\uparrow$ & Avg.$\uparrow$ \\
\midrule
\tablegroupc{tintDR}{6}{\grouplabel{Deep Research mode}{}} \\
\modelname{openai}{o3} & \textbf{2.14} & 2.71 & \textbf{3.06} & \textbf{2.28} & \textbf{2.55} \\
\modelname{doubao-color}{Doubao} & 1.90 & 2.57 & 2.86 & 2.03 & 2.34 \\
\modelname{deepseek-color}{DeepSeek} & 1.90 & 2.57 & 2.75 & 1.96 & 2.30 \\
\modelname{gemini-color}{Gemini} & 2.03 & \textbf{2.76} & 2.78 & 2.25 & 2.46 \\
\modelname{grok}{Grok} & 2.08 & 2.58 & \textbf{3.06} & 2.19 & 2.48 \\
\modelname{perplexity-color}{Perplexity} & 2.07 & 2.60 & 2.78 & 2.10 & 2.39 \\
\modelname{qwen-color}{Qwen} & 2.01 & 2.62 & 2.76 & 2.08 & 2.37 \\

\midrule
\tablegroupc{tintBU}{6}{\grouplabel{Bottom-Up mode}{}} \\
\tablegroup{6}{\textit{Non-thinking-based}} \\
\modelname{claude-color}{Claude-Sonnet-4.5} & \textbf{2.17} & \textbf{2.74} & 2.90 & 2.32 & 2.53 \\
\modelname{openai}{GPT-5} & 2.11 & 2.71 & 2.83 & 2.43 & 2.52 \\
\modelname{gemini-color}{Gemini-3-Pro*} & 2.18 & 2.57 & 2.97 & 2.46 & 2.55 \\
\modelname{deepseek-color}{DeepSeek-V3.2} & \textbf{2.17} & 2.65 & \textbf{3.01} & \textbf{2.53} & \textbf{2.59} \\
\modelname{qwen-color}{Qwen3-Max*} & 2.14 & 2.54 & 2.85 & 2.29 & 2.46 \\
\modelname{kimi-color}{Kimi-K2} & 2.04 & 2.57 & 2.85 & 2.36 & 2.46 \\
\midrule
\tablegroup{6}{\textit{Thinking-based}} \\
\modelname{claude-color}{Claude-Sonnet-4.5-Thinking} & \textbf{2.17} & 2.68 & 2.83 & 2.32 & \textbf{2.50} \\
\modelname{openai}{GPT-5-Thinking} & 2.14 & 2.68 & 2.83 & 2.29 & 2.49 \\
\modelname{gemini-color}{Gemini-3-Pro-Thinking*} & 2.03 & \textbf{2.71} & 2.81 & 2.28 & 2.46 \\
\modelname{deepseek-color}{DeepSeek-V3.2-Thinking} & 2.14 & 2.54 & \textbf{2.94} & 2.25 & 2.47 \\
\modelname{qwen-color}{Qwen3-Max-Thinking*} & 2.11 & 2.60 & 2.88 & \textbf{2.43} & \textbf{2.50} \\
\modelname{kimi-color}{Kimi-K2-Thinking} & 2.10 & 2.50 & 2.64 & 2.39 & 2.41 \\
\bottomrule
\end{tabular}%

\caption{\textbf{LLM-as-Judge evaluation} results across Deep Research and Bottom-Up modes.}
\label{tab:llm_judge_only}
\end{table*}

\begin{table*}[t]
\centering
\adjustbox{max width=\textwidth}{%
\begin{tabular}{llccccccc}
\toprule
\multirow{2}{*}{Model} & \multirow{2}{*}{Input Setting} &
\multicolumn{4}{c}{Leaf-Level} & \multicolumn{3}{c}{Hierarchy-Level} \\
\cmidrule(lr){3-6} \cmidrule(lr){7-9} 
& & ARI$\uparrow$ & Hom.$\uparrow$ & Comp.$\uparrow$ & V-Meas.$\uparrow$ &
US-TED$\downarrow$ & US-NTED$\downarrow$ & \textsc{Sem-Path}$\uparrow$ \\
\midrule

\multirow{3}{*}{\modelname{claude-color}{Claude-Sonnet-4.5}}
& Title + Abs & 27.25\% & 82.27\% & 66.61\% & 72.84\% & 42.89 & 77.73\% & 29.16\% \\
& + Summary & 24.20\% & \textbf{85.81\%} & 66.16\% & \textbf{73.88\%} & 45.75 & 76.52\% & 29.09\% \\
& + Core-task \& Contrib. & 24.95\% & 84.98\% & 66.03\% & 73.57\% & 44.01 & 75.41\% & \textbf{29.18\%} \\
\midrule
\multirow{3}{*}{\modelname{claude-color}{Claude-Sonnet-4.5-Thinking}}
& Title + Abs & 29.28\% & 76.35\% & 66.96\% & 70.58\% & 37.87 & 76.56\% & 28.66\% \\
& + Summary & 28.23\% & 79.49\% & 66.83\% & 71.92\% & 39.39 & \textbf{74.35\%} & 28.69\% \\
& + Core-task \& Contrib. & 29.39\% & 79.23\% & 66.77\% & 71.75\% & 39.73 & 76.00\% & 28.69\% \\
\midrule

\multirow{3}{*}{\modelname{openai}{GPT-5}}
& Title + Abs & 28.17\% & 77.79\% & 66.45\% & 70.99\% & 40.84 & 78.65\% & 28.97\% \\
& + Summary & 28.27\% & 80.13\% & 66.85\% & 72.13\% & 41.44 & 79.17\% & 28.72\% \\
& + Core-task \& Contrib. & 28.51\% & 79.94\% & 66.84\% & 72.10\% & 41.94 & 79.45\% & 28.64\% \\
\midrule
\multirow{3}{*}{\modelname{openai}{GPT-5-Thinking}}
& Title + Abs & 25.97\% & 80.89\% & 66.05\% & 72.07\% & 43.05 & 79.86\% & 28.56\% \\
& + Summary & 28.36\% & 81.66\% & 66.98\% & 72.90\% & 42.52 & 79.66\% & 28.84\% \\
& + Core-task \& Contrib. & 26.95\% & 80.46\% & 66.37\% & 71.91\% & 42.56 & 79.56\% & 28.85\% \\
\midrule

\multirow{3}{*}{\modelname{gemini-color}{Gemini-3-Pro*}}
& Title + Abs & 29.86\% & 68.25\% & 67.15\% & 67.00\% & \textbf{32.32} & 76.35\% & 28.13\% \\
& + Summary & 26.90\% & 68.83\% & 65.66\% & 66.43\% & 33.53 & 75.82\% & 28.16\% \\
& + Core-task \& Contrib. & 27.84\% & 69.89\% & 65.91\% & 67.15\% & 33.66 & 76.09\% & 28.23\% \\
\midrule
\multirow{3}{*}{\modelname{gemini-color}{Gemini-3-Pro-Thinking*}}
& Title + Abs & 28.84\% & 67.26\% & 67.00\% & 66.45\% &32.40 & 76.47\% & 28.21\% \\
& + Summary & 26.17\% & 67.77\% & 65.09\% & 65.70\% & 33.28 & 75.41\% & 27.91\% \\
& + Core-task \& Contrib. & 29.17\% & 69.26\% & 66.72\% & 67.31\% & 32.65 & 75.28\% & 28.26\% \\
\midrule

\multirow{3}{*}{\modelname{deepseek-color}{DeepSeek-V3.2}}
& Title + Abs & 27.15\% & 74.33\% & 66.15\% & 69.28\% & 37.04 & 75.71\% & 28.63\% \\
& + Summary & 24.95\% & 79.52\% & 65.21\% & 71.11\% & 41.84 & 74.60\% & 28.99\% \\
& + Core-task \& Contrib. & 21.76\% & 81.39\% & 64.26\% & 71.26\% & 45.56 & 75.42\% & 28.69\% \\
\midrule
\multirow{3}{*}{\modelname{deepseek-color}{DeepSeek-V3.2-Thinking}}
& Title + Abs & 27.71\% & 70.74\% & 67.14\% & 67.68\% & 34.51 & 76.02\% & 29.07\% \\
& + Summary & 28.07\% & 72.63\% & 66.80\% & 68.68\% & 37.18 & 77.98\% & 28.44\% \\
& + Core-task \& Contrib. & 27.76\% & 71.30\% & 67.06\% & 67.90\% & 35.98 & 76.94\% & 28.67\% \\
\midrule

\multirow{3}{*}{\modelname{qwen-color}{Qwen3-Max*}}
& Title + Abs & \textbf{31.24\%} & 68.48\% & \textbf{68.63\%} & 67.52\% & 32.80 & 78.17\% & 29.00\% \\
& + Summary & 28.43\% & 72.85\% & 67.39\% & 69.19\% & 35.05 & 78.55\% & 29.02\% \\
& + Core-task \& Contrib. & 28.37\% & 72.41\% & 66.61\% & 68.53\% & 35.60 & 78.94\% & 28.66\% \\
\midrule
\multirow{3}{*}{\modelname{qwen-color}{Qwen3-Max-Thinking*}}
& Title + Abs & 30.32\% & 68.70\% & 68.10\% & 67.38\% & 33.62 & 78.24\% & 28.95\% \\
& + Summary & 29.24\% & 73.64\% & 67.39\% & 69.55\% & 35.85 & 78.44\% & 28.93\% \\
& + Core-task \& Contrib. & 27.54\% & 73.58\% & 66.54\% & 69.06\% & 35.54 & 78.77\% & 28.78\% \\
\midrule

\multirow{3}{*}{\modelname{kimi-color}{Kimi-K2}}
& Title + Abs & 23.69\% & 79.32\% & 64.88\% & 70.48\% & 41.35 & 79.58\% & 28.33\% \\
& + Summary & 20.74\% & 82.81\% & 64.11\% & 71.47\% & 41.96 & 78.96\% & 28.38\% \\
& + Core-task \& Contrib. & 21.58\% & 82.79\% & 64.45\% & 71.65\% & 42.61 & 78.31\% & 28.50\% \\
\midrule
\multirow{3}{*}{\modelname{kimi-color}{Kimi-K2-Thinking}}
& Title + Abs & 22.59\% & 78.98\% & 64.64\% & 70.18\% & 44.07 & 80.85\% & 28.21\% \\
& + Summary & 20.95\% & 81.20\% & 64.37\% & 70.96\% & 44.95 & 79.19\% & 28.23\% \\
& + Core-task \& Contrib. & 21.76\% & 80.27\% & 64.82\% & 70.91\% & 44.13 & 78.87\% & 28.81\% \\

\bottomrule
\end{tabular}%
}
\caption{\textbf{Ablation on input granularity} (Bottom-Up mode). Models marked with * denote the Preview versions. We incrementally add information to the input: \textbf{Title + Abs} uses only titles and abstracts; \textbf{+ Summary} further includes machine-generated summaries; \textbf{+ Core-task \& Contrib.} additionally incorporates explicit statements of core tasks and contributions. Best results within each model group are \textbf{bold}.}
\label{tab:bottom_up_ablation}
\end{table*}

\section{Error Analysis and Case Studies}
\label{app:error_and_case_study}
\subsection{Summary of Four-Dimensional Error Analysis}
\label{subsec:summary_error_four_dimensional}

\begin{figure*}[t]
  \centering
  \includegraphics[width=\linewidth]{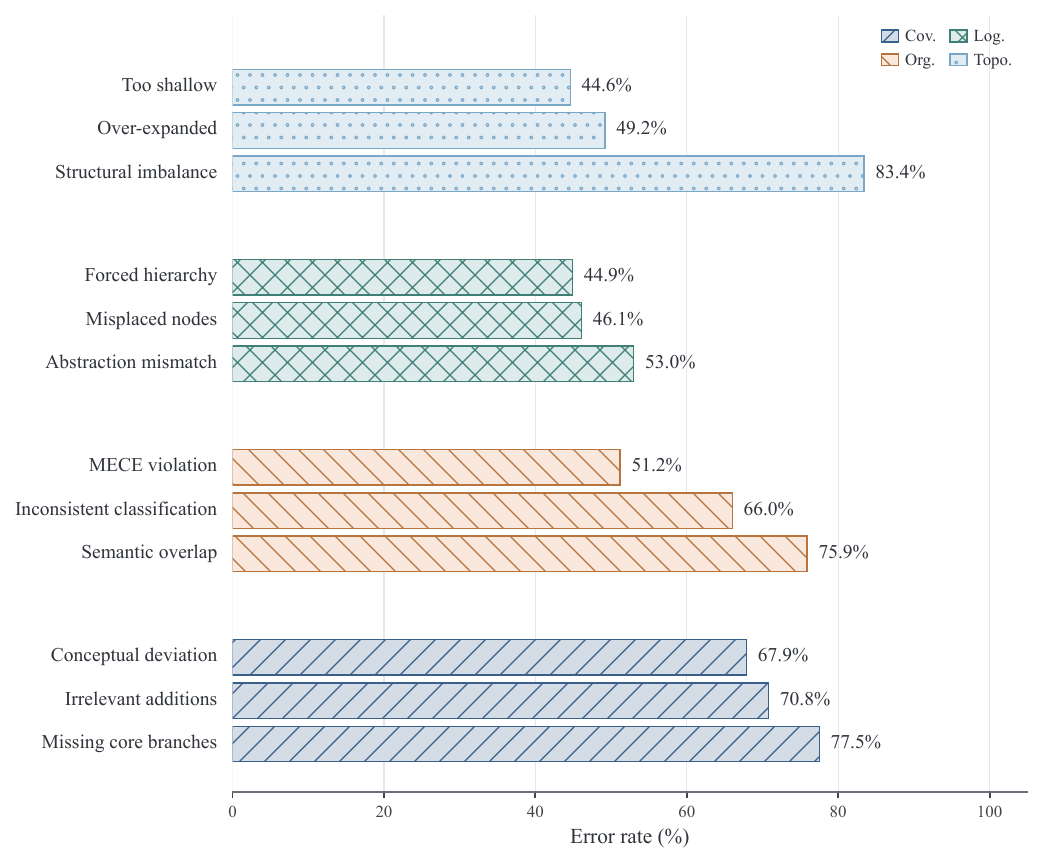}
    \caption{Top 3 error types by evaluation category. Error rates computed from qualitative analysis of 1000 model-generated taxonomies.}
    \label{fig:error_analysis_top}
\end{figure*}

\begin{figure*}[t]
  \centering
  \includegraphics[width=\textwidth]{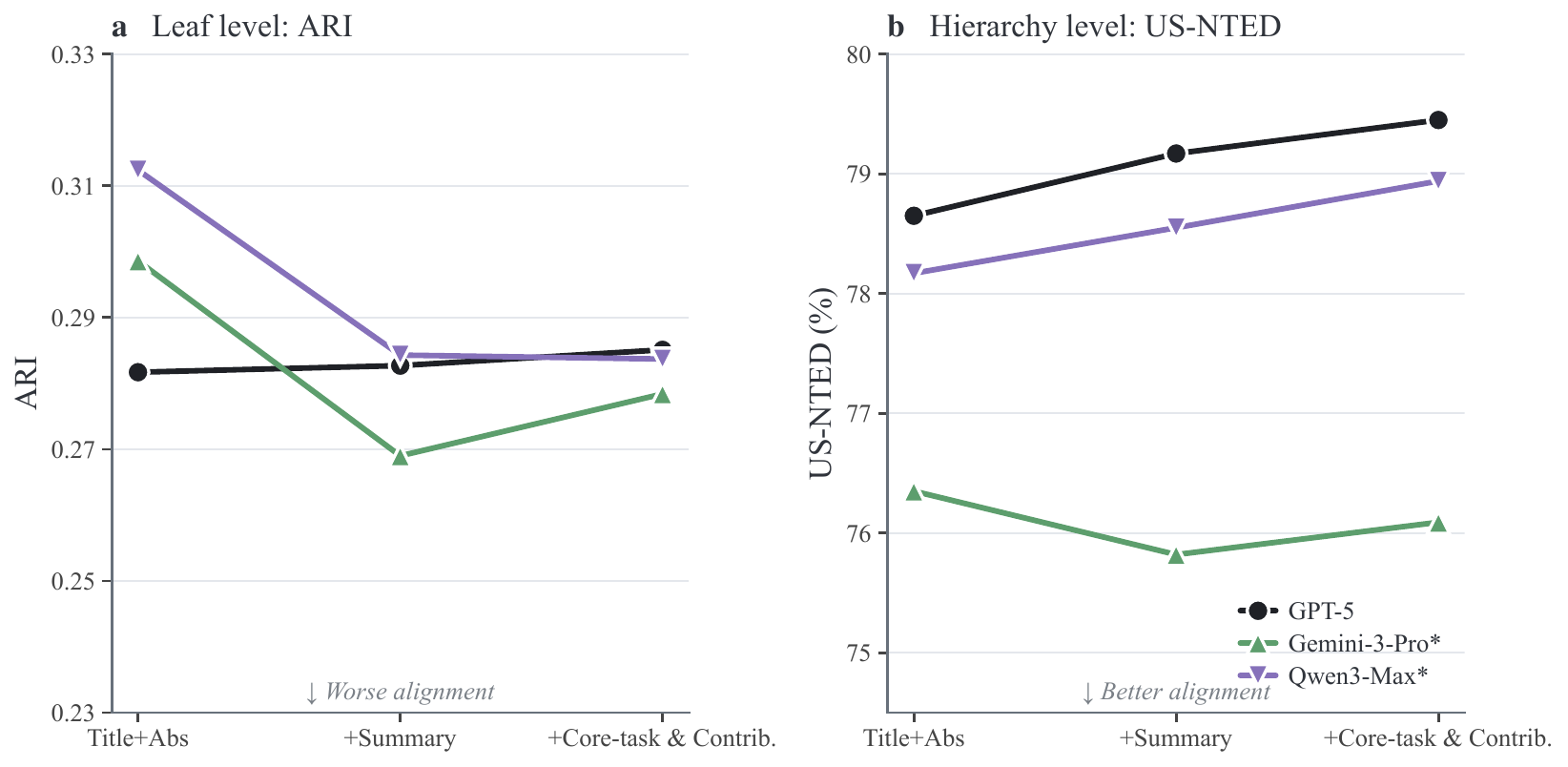}
    \caption{Ablation study on input settings in Bottom-Up mode. (a) ARI scores decrease as more information is provided, indicating reduced alignment with expert classifications. (b) US-NTED shows mixed or improving trends, suggesting that richer input does not necessarily help models generate taxonomies more aligned with expert standards.}
    \label{fig:ablation_diff_input_th}
\end{figure*}

To gain deeper insights into the limitations of current LLMs in taxonomy generation, we conducted a fine-grained error analysis on 1,000 sampled taxonomies. We categorized structural and semantic failures into four distinct dimensions:
\begin{itemize}
    \item \textbf{Semantic Coverage:} Evaluates the completeness and relevance of the taxonomy. Errors include omitting core subfields recognized by experts or including hallucinatory/irrelevant branches.
    \item \textbf{Sibling Organization:} Assesses the quality of nodes within the same hierarchical level. Common failures involve semantic redundancy (violating the MECE principle) or inconsistent classification criteria.
    \item \textbf{Hierarchical Logic:} Examines the validity of parent-child relationships. Errors include abstraction mismatches (e.g., placing a high-level concept under a specific method) or misclassification of paper categories.
    \item \textbf{Structural Topology:} Analyzes the overall shape and balance of the tree. This dimension captures issues such as structural imbalance, excessive depth, or insufficient granularity compared to human-curated benchmarks.
\end{itemize}

Figure~\ref{fig:error_analysis_top} summarizes the top 3 error types for each category, and Table~\ref{tab:error_analysis_detailed} lists the full per-dimension breakdown. The data reveals that \textit{Structural Imbalance} (83.4\%) and \textit{Missing Core Branches} (77.5\%) are the most prevalent issues, highlighting the challenge models face in maintaining a global perspective on domain structures.

\begin{table*}[t]
\centering
\small
\setlength{\tabcolsep}{8pt} 
\begin{tabular}{ll r} 
\toprule
Dimension & Error Type & Freq. (\%) \\ 
\midrule

\multirow{5}{*}{\textbf{Semantic Coverage}}
& Missing Core Branches & 77.5\% \\
& Irrelevant Additions & 70.8\% \\
& Conceptual Deviation & 67.9\% \\
& Missing Subfields & 58.0\% \\
& Incomplete Recall & 16.6\% \\
\midrule

\multirow{6}{*}{\textbf{Sibling Organization}}
& Semantic Overlap & 75.9\% \\
& Inconsistent Classification & 66.0\% \\
& MECE Violation & 51.2\% \\
& Blurred Boundaries & 40.3\% \\
& Over-fragmentation & 11.8\% \\
& Under-fragmentation & 3.2\% \\
\midrule

\multirow{6}{*}{\textbf{Hierarchical Logic}}
& Abstraction Mismatch & 53.0\% \\
& Misplaced Nodes & 46.1\% \\
& Forced Hierarchy & 44.9\% \\
& Logical Inconsistency & 28.5\% \\
& Missing Intermediate & 5.6\% \\
& Inverted Relationships & 0.6\% \\
\midrule

\multirow{6}{*}{\textbf{Structural Topology}}
& Structural Imbalance & 83.4\% \\
& Over-expanded & 49.2\% \\
& Too Shallow & 44.6\% \\
& Under-expanded & 42.4\% \\
& Granularity Mismatch & 36.3\% \\
& Too Deep & 19.3\% \\
\bottomrule
\end{tabular}
\caption{\textbf{Error-type statistics} on 1{,}000 model-generated taxonomies. Errors are sorted by frequency within each dimension. Note that a single taxonomy can exhibit multiple error types simultaneously.}
\label{tab:error_analysis_detailed}
\end{table*}

\subsection{Analysis of Bottom-Up Model Results under Different Input Forms}
\label{subsec:bottomup_results_input_forms}

Table~\ref{tab:bottom_up_ablation} ablates the granularity of the input description in Bottom-Up mode. Richer text does not improve alignment with the expert reference at either the leaf or the hierarchy level; in many cases it makes alignment worse.

At the leaf level (Figure~\ref{fig:ablation_diff_input_th}), adding machine-generated summaries (\textbf{+ Summary}) or explicit core-task descriptions (\textbf{+ Core-task \& Contrib.}) consistently lowers ARI across the high-performing models, even though internal clustering purity (Homogeneity) often goes up. From \textit{Title + Abs} to \textit{+ Summary}, ARI drops for \textbf{Qwen3-Max*} (31.24\% $\to$ 28.43\%) and \textbf{Gemini-3-Pro*} (29.86\% $\to$ 26.90\%); the same direction holds at the hierarchy level, where \textsc{US-NTED} (lower is better) rises for \textbf{GPT-5} (78.65\% $\to$ 79.17\% $\to$ 79.45\%) and \textbf{Qwen3-Max*} (78.17\% $\to$ 78.55\% $\to$ 78.94\%), while \textsc{Sem-Path} stays essentially flat.

Read together with the Constrained-$K$ probe (App.~\ref{app:constrained_k}), the pattern is consistent with over-segmentation rather than with a content gap: more input text encourages the model to invent additional fine-grained categories that drive Homogeneity up but move the partition further from the expert assignment. Providing more text does not, on its own, close the alignment gap in Findings~4--5.

\subsection{Human Baseline Study}
\label{subsec:human_vs_model_study}
\label{app:human_baseline}

To establish a human reference under the \emph{same} Bottom-Up requirements given to models, we asked \emph{non-author} CS graduate students familiar with AI/ML literature to organize each expert paper set into a hierarchical taxonomy. Participants never saw the published expert taxonomy, so the study is a \emph{lower bound} on the performance of the domain experts who wrote the surveys. Two annotators independently completed all 10 surveys (protocol in App.~\ref{app:human_protocol}).

\paragraph{Task setup.}
Humans and models receive the same inputs and organization target: the Title+Abstract paper set $\mathcal{P}^*$ for a survey, with instructions to build a hierarchical taxonomy that covers every paper. Neither side sees the expert reference, and neither is given a target depth, category count, or score. We score both with the identical metric implementation used throughout the paper (\textsc{ARI}, \textsc{Sem-Path}, depth-matched \textsc{Sem-Path}, \textsc{US-NTED}, and the judge-free shape statistics). All human--model contrasts use the 10 annotated surveys.

\paragraph{Coverage and independence.}
The study covers \textbf{10 of the 72 surveys}, each annotated independently by both annotators. Every human taxonomy assigns every expert paper exactly once (full coverage, no duplicates), and the transcribed expert trees match the benchmark \texttt{data.jsonl} entries for those IDs. We check independence by comparing the two annotators' per-paper depth vectors, their level-wise ancestor partitions, and whether either leaf partition is a merge-only refinement of the other; a third submission that failed these checks was excluded.

\paragraph{Matched-subset comparison.}
Table~\ref{tab:human_baseline_matched} summarizes the result. The two annotators average \textbf{53.91\%} \textsc{Sem-Path} and reach 55.20\% and 52.62\% individually. On the same 10 surveys, every current-generation Bottom-Up configuration remains in a 27.07--29.68\% \textsc{Sem-Path} band, within 2.2\,pp of the 27.49\% \textsc{Flat} floor from App.~\ref{app:metric_calibration}. Relative to the strongest matched configuration on ARI (Kimi-K3, lowest reasoning effort: 40.64\% ARI, 29.27\% \textsc{Sem-Path}), the paired human advantage is \textbf{$+$24.64\,pp} \textsc{Sem-Path} (bootstrap 95\% CI $[20.20, 28.98]$; humans higher on \textbf{10/10} surveys) and \textbf{$+$1.55} levels of depth (CI $[1.05, 2.05]$; 9/10). Humans match expert depth on this subset (4.60 and 4.70 versus the 4.70 expert mean; full-set expert mean 4.86), whereas the matched Kimi-K3 trees average only 3.10.

ARI behaves differently. The paired ARI advantage is only \textbf{$+$5.63\,pp} with a confidence interval spanning zero (CI $[-1.77, 12.79]$; humans higher on 6/10), and the two annotators differ sharply from each other on this metric (69.55\% versus 23.00\% against the expert) while differing by less than 3\,pp on \textsc{Sem-Path}. Flat assignment agreement against a single reference is thus unstable across human taxonomies, whereas the hierarchy-aware metrics are not.

\begin{table*}[t]
\centering
\small
\setlength{\tabcolsep}{4pt}
\begin{tabular}{lccccccc}
\toprule
System & ARI$\uparrow$ & \textsc{Sem-Path}$\uparrow$ & SP$_{\mathrm{dm}}\uparrow$ & \textsc{US-NTED}$\downarrow$ & Depth & \#Cat & Singl. \\
\midrule
Annotator A1 & 69.55\% & 55.20\% & 55.55\% & 26.77\% & 4.60 & 13.2 & 11.1\% \\
Annotator A2 & 23.00\% & 52.62\% & 52.62\% & 32.83\% & 4.70 & 11.6 & 13.0\% \\
Human mean & \textbf{46.27\%} & \textbf{53.91\%} & \textbf{54.09\%} & \textbf{29.80\%} & 4.65 & 12.4 & 12.0\% \\
\midrule
\modelname{kimi-color}{Kimi-K3 (lowest effort)} & 40.64\% & 29.27\% & 40.81\% & 78.18\% & 3.10 & 14.9 & 21.2\% \\
\modelname{claude-color}{Claude-Sonnet-5 (lowest effort)} & 39.72\% & 29.68\% & 37.83\% & 74.93\% & 3.50 & 14.4 & 17.4\% \\
\midrule
\emph{Expert reference (matched 10)} & --- & --- & --- & --- & \emph{4.70} & --- & --- \\
\emph{Expert reference (all 72)} & --- & --- & --- & --- & \emph{4.86} & \emph{14.0} & \emph{17.0\%} \\
\bottomrule
\end{tabular}
\caption{\textbf{Human baseline on a matched 10-survey subset.} Humans and models receive the same Title+Abstract Bottom-Up task and are scored with the same code. SP$_{\mathrm{dm}}$ is depth-matched \textsc{Sem-Path}; bold marks the better of the human mean and the models on each metric. Model rows are macro-averages over the same 10 surveys; Kimi-K3 is the strongest ARI configuration on this subset and Claude-Sonnet-5 the strongest raw \textsc{Sem-Path}. The two annotators separate from every model on \textsc{Sem-Path}, \textsc{US-NTED}, and depth, but not on ARI, where they disagree with each other more than they disagree with the models.}
\label{tab:human_baseline_matched}
\end{table*}

\paragraph{Inter-annotator agreement and metric headroom.}
Because both annotators covered all 10 surveys, we can measure agreement without any expert reference and use it to calibrate the metrics. Table~\ref{tab:iaa_fourway} reports four comparisons under one scoring pipeline: annotator versus annotator, annotator versus expert, model versus expert, and model versus model. For the symmetric human--human and model--model cases we average the metric in both directions, since \textsc{Sem-Path} is defined against a reference. Three readings follow.

First, ARI and \textsc{V-Measure} have no usable headroom left. Models agree with the expert at 35.76\% ARI and 75.91\% \textsc{V-Measure}, \emph{above} the 22.92\% and 63.54\% at which two independent annotators agree with each other. A system can therefore exceed inter-annotator agreement on flat assignment metrics, so a score gap on them against a single reference is not a quality gap.

Second, the hierarchy-aware metrics do not have this property, and the model gap on them is far larger than annotator disagreement. Annotators agree with each other at 50.56\% \textsc{Sem-Path} while models reach 28.21\% against the expert, that is 0.56$\times$ the annotator-agreement level and within 0.7\,pp of the \textsc{Flat} floor. On \textsc{US-NTED} the two annotators sit 31.99\% apart and 29.80\% from the expert, while models sit 77.04\% from the expert, 2.4$\times$ further. The headroom these metrics expose therefore reflects a capability gap rather than subjectivity in the choice of reference.

Third, models are more homogeneous than humans. Model--model agreement (46.04\% ARI, 81.83\% \textsc{V-Measure}) exceeds annotator--annotator agreement (22.92\%, 63.54\%), so eight models from seven families converge on each other more than two people do, while all of them land far from the expert on hierarchy. Model taxonomies are variations on one shallow, wide shape that neither annotator produced.

\begin{table*}[t]
\centering
\small
\setlength{\tabcolsep}{5pt}
\begin{tabular}{lccccc}
\toprule
Comparison & ARI$\uparrow$ & \textsc{V-Meas.}$\uparrow$ & \textsc{Sem-Path}$\uparrow$ & SP$_{\mathrm{dm}}\uparrow$ & \textsc{US-NTED}$\downarrow$ \\
\midrule
Annotator vs.\ annotator & 22.92\% & 63.54\% & 50.56\% & 50.81\% & 31.99\% \\
Annotator vs.\ expert & 46.27\% & 77.09\% & 53.91\% & 54.09\% & 29.80\% \\
\midrule
Model vs.\ expert & 35.76\% & 75.91\% & \textbf{28.21\%} & 38.71\% & \textbf{77.04\%} \\
Model vs.\ model & 46.04\% & 81.83\% & 47.82\% & 49.95\% & 44.35\% \\
\bottomrule
\end{tabular}
\caption{\textbf{Four-way agreement on the matched 10 surveys}, one scoring pipeline throughout. Model rows pool all 16 current-generation Bottom-Up configurations. Models agree with the expert \emph{more} than two independent annotators agree with each other on ARI and \textsc{V-Measure}, so those metrics are saturated relative to the reproducibility of a single reference. On \textsc{Sem-Path} and \textsc{US-NTED} the ordering reverses: models reach 0.56$\times$ the annotator-agreement level on \textsc{Sem-Path} and sit 2.4$\times$ further from the expert on \textsc{US-NTED} than the annotators sit from each other. Model--model agreement above annotator--annotator agreement shows that model taxonomies are more homogeneous than human ones. Substituting the excluded third delivery for annotator A2 changes no entry by more than 2.2\,pp and preserves every ordering in this table.}
\label{tab:iaa_fourway}
\end{table*}

\paragraph{Floor effect.}
Human scores clear the \textsc{Flat} floor by 26.42\,pp on the same surveys and scoring pipeline, so \textsc{Sem-Path} rewards deep, coherent taxonomies; the narrow cross-model spread near 28--30\% is compressed by the floor.

\paragraph{Depth matching does not erase the gap.}
App.~\ref{app:metric_calibration} notes that raw \textsc{Sem-Path} is sensitive to tree shape, and \textsc{Shuffle} scores 49.45\% while destroying placement. We therefore also report depth-matched \textsc{Sem-Path}: humans average 54.09\% versus 40.81\% for matched Kimi-K3, a paired advantage of \textbf{$+$13.27\,pp} (CI $[10.69, 16.06]$; humans higher on 10/10). Shape explains part of the raw gap, but a large human advantage remains after the depth control. Combined with the near-expert human depth of 4.65, this supports reading Finding~6's depth deficit as largely model-specific under the shared task requirements.

Figure~\ref{fig:human_baseline} summarizes both results visually, and Figure~\ref{fig:taxonomy_comparison} illustrates a side-by-side comparison between an expert taxonomy and a human-constructed alternative on the same paper set.

\subsection{Human Annotation Protocol}
\label{app:human_protocol}

This section records the task specification the annotators worked from and the checks each submission had to pass.

\paragraph{Annotators.}
Each annotator is a graduate student or researcher in machine learning or NLP who reads survey papers in the area. Annotator identifiers are pseudonymous (A1, A2), and no participant-level material is released. Annotators were paid 100 RMB per hour and were told before starting that their trees would be scored and reported in aggregate, and that they could withdraw.

\paragraph{What the annotator receives.}
One packet per survey containing the survey topic, the fixed root name, and every scored paper with its title and full abstract, matching the model prompt.

\paragraph{What the annotator does not receive.}
The packet contains no expert taxonomy, no source survey or its section structure, no other annotator's tree, no model-generated taxonomy, and no metric, score, or target range. Annotators were asked not to look up the survey and to flag any survey whose structure they already remembered.

\paragraph{Task.}
Organize every paper into a hierarchical topic tree that would be defensible in a survey paper, under three constraints: every paper appears exactly once, a category holds either papers or sub-categories but never both, and the root is the name given in the packet. Depth and the number of categories are the annotator's choice. Annotators group by what the papers are about rather than by the surface wording of titles.

\paragraph{Independence.}
Annotators work alone in separate files with no shared folder, and discuss no survey until every submission is in. Each annotator receives a different survey order so that fatigue and learning effects do not align across annotators.

\paragraph{Submission format.}
Trees are submitted as indented outlines in which heading depth gives category depth and a line \texttt{- N} places paper \texttt{N} into the category above it. Annotators cite paper numbers rather than retyping titles, and the ingest step maps numbers back to exact titles; retyped titles are the main source of coverage errors at this scale.

\paragraph{Validation on receipt.}
Every submission passes through two gates before any number is quoted. The ingest step rejects a tree that omits a paper, places one twice, uses an out-of-range paper number, skips an outline level, leaves a category empty, or hangs papers off a node that also has sub-categories. The independence step then compares annotators on per-paper depth vectors, level-wise ancestor partitions, and whether either leaf partition is a merge-only refinement of the other, and must report a pass before agreement is computed. These checks excluded the third submission mentioned in App.~\ref{app:human_baseline}.

\subsection{Over-segmentation in Model-generated Trees}
\label{subsec:over_segmentation_model_generated_trees}
Figure~\ref{fig:taxonomy_color_comparison} illustrates a key distinction between the taxonomy generated by the Kimi-thinking model and the expert taxonomy. The model-generated tree exhibits a tendency to create singleton clusters, isolating specific papers into narrow, fine-grained subcategories of their own. This pattern illustrates one mechanism behind Finding~4: the model prioritizes precise, local semantic descriptions over broader, structural aggregation.

\begin{figure*}[p]
    \centering
    \begin{minipage}[t]{0.49\textwidth}
        \begin{tcolorbox}[
            enhanced,
            colback=humanBg,
            colframe=humanBg!80!black,
            arc=2mm,
            boxrule=0.5pt,
            title={\centering\bfseries\scriptsize (a) Human Expert Taxonomy},
            coltitle=black,
            fonttitle=\sffamily,
            height=15cm,
            valign=top,
            before upper={\setlength{\parskip}{0.05em}\setlength{\baselineskip}{0.95em}}
        ]
            
            \Tr{ROOT: Large Language Model Agent in Financial Trading: A Survey-Finance Trading Agent}
            \treegap \treeline
            
            \treegap \treebranch \Tone{LLM as a Trader}
            \treegap \treeline \treegap \treeline
            
            \treegap \treeline \treegap \treebranch \Ttwo{News-Driven}
            \treegap \treeline \treegap \treeline \treegap \treebranch \Tp{Unveiling the Potential of Sentiment...}
            \treegap \treeline \treegap \treeline \treegap \treebranch \Tp{LLMFactor: Extracting Profitable Factors...}
            \treegap \treeline \treegap \treeline \treegap \treebranch \Tp{Can ChatGPT Forecast Stock Price...}
            \treegap \treeline \treegap \treeline \treegap \treebranch \Tp{Sentiment trading with LLMs...}
            \treegap \treeline \treegap \treeline \treegap \treebranch \Tp{Modeling asset allocation strategies...}
            \treegap \treeline \treegap \treeline \treegap \treelast  \Tp{Can LLMs Beat Wall Street?}
            
            \treegap \treeline \treegap \treeline
            
            \treegap \treeline \treegap \treebranch \Ttwo{Debate-Driven}
            \treegap \treeline \treegap \treeline \treegap \treebranch \Tp{Designing Heterogeneous LLM Agents...}
            \treegap \treeline \treegap \treeline \treegap \treelast  \Tp{TradingGPT: Multi-Agent System...}
            
            \treegap \treeline \treegap \treeline
            
            \treegap \treeline \treegap \treelast  \Ttwo{Reflection-Driven}
            \treegap \treeline \treegap \treegap \treegap \treebranch \Tp{A Multimodal Foundation Agent...}
            \treegap \treeline \treegap \treegap \treegap \treelast  \Tp{FinMem: A Performance-Enhanced LLM...}
            \treegap \treebranch \Tone{LLM as an Alpha Miner}
            \treegap \treeline \treegap \treebranch \Ttwo{AlphaGPT}
            \treegap \treeline \treegap \treeline \treegap \treebranch \Tp{Alpha-GPT: Human-AI...}
            
            \treegap \treeline \treegap \treeline
            
            \treegap \treeline \treegap \treelast  \Ttwo{QuantAgent}
            \treegap \treeline \treegap \treegap \treegap \treebranch \Tp{QuantAgent: Seeking Holy Grail...}

        \end{tcolorbox}
    \end{minipage}%
    \hfill
    \begin{minipage}[t]{0.49\textwidth}
        \begin{tcolorbox}[
            enhanced,
            colback=modelBg,
            colframe=modelBg!80!black,
            arc=2mm,
            boxrule=0.5pt,
            title={\centering\bfseries\scriptsize (b) Model Generated Taxonomy},
            coltitle=black,
            fonttitle=\sffamily,
            height=15cm,
            valign=top,
            before upper={\setlength{\parskip}{0.05em}\setlength{\baselineskip}{0.95em}}
        ]
            
            \Tr{ROOT: Large Language Model Agent in Financial Trading: A Survey-Finance Trading Agent}
            \treegap \treeline
            
            \treegap \treebranch \Tone{Market Intelligence \& Sentiment}
            \treegap \treeline \treegap \treebranch \Ttwo{Heterogeneous Frameworks...}
            \treegap \treeline \treegap \treeline \treegap \treelast \Tp{Designing Heterogeneous Agents...}
            
            \treegap \treeline \treegap \treebranch \Ttwo{Cross-Lingual Sentiment... \Sgl}
            \treegap \treeline \treegap \treeline \treegap \treelast \Tp{Unveiling the Potential...}
            
            \treegap \treeline \treegap \treelast \Ttwo{Comparative Assessment... \Sgl}
            \treegap \treeline \treegap \treegap \treegap \treelast \Tp{Sentiment trading with LLMs...}
            
            \treegap \treeline
            
            \treegap \treebranch \Tone{Trading Strategy Formulation}
            \treegap \treeline \treegap \treebranch \Ttwo{News-Driven Forecasting... \Sgl}
            \treegap \treeline \treegap \treeline \treegap \treelast \Tp{Can ChatGPT Forecast Stock Price...}
            
            \treegap \treeline \treegap \treelast \Ttwo{Integrated Stock Selection... \Sgl}
            \treegap \treeline \treegap \treegap \treegap \treelast \Tp{Can LLMs Beat Wall Street?}

            \treegap \treeline \treegap \treebranch \Ttwo{Feature-Enhanced Return Prediction Models \Sgl}
            \treegap \treeline \treegap \treeline \treegap \treelast \Tp{Integrating Stock Features and...}
            
            \treegap \treeline
            
            \treegap \treelast \Tone{Agent Cognitive Architecture}
            \treegap \treegap \treegap \treebranch \Ttwo{Layered Memory Systems... \Sgl}
            \treegap \treegap \treegap \treeline \treegap \treelast \Tp{FinMem: Performance-Enhanced...}
            
            \treegap \treegap \treegap \treebranch \Ttwo{Multi-Agent Collaboration... \Sgl}
            \treegap \treegap \treegap \treeline \treegap \treelast \Tp{TradingGPT: Multi-Agent System...}
            
            \treegap \treegap \treegap \treelast \Ttwo{Multimodal Foundation... \Sgl}
            \treegap \treegap \treegap \treeline \treegap \treelast \Tp{A Multimodal Foundation Agent...}

            \treegap \treelast \Tone{...}

        \end{tcolorbox}
    \end{minipage}
    \caption{Comparison between \textbf{human expert taxonomy (left)} and \textbf{Kimi-K2-Thinking generated taxonomy (right)}. The model-generated tree exhibits over-segmentation with many singleton clusters (marked as \Sgl), while the expert taxonomy uses broader thematic groupings.}
    \label{fig:taxonomy_color_comparison}
\end{figure*}

\begin{figure*}[p]
\centering
\begin{minipage}{\textwidth}
\begin{tcolorbox}[
    enhanced,
    colback=humanBg,
    colframe=humanBg!80!black,
    arc=2mm,
    boxrule=0.5pt,
    title={\centering\bfseries\scriptsize (a) Human-Expert Taxonomy}, 
    coltitle=black,
    fonttitle=\sffamily,       
    width=\textwidth,         
    valign=top,
    top=5pt, bottom=5pt, left=5pt, right=5pt, 
    before upper={\setlength{\parskip}{0.05em}\setlength{\baselineskip}{0.95em}} 
]

\Tr{Exploring Large Language Model based Intelligent Agents}

\treegap \treebranch \Tone{Actions of LLM-based Agents}
    
    \treegap \treeline \treegap \treebranch \Ttwo{Tool Creation}
        \treegap \treeline \treegap \treeline \treegap \treebranch \Tp{CRAFT: Customizing LLMs by Creating and Retrieving from Specialized Toolsets}
        \treegap \treeline \treegap \treeline \treegap \treelast  \Tp{Large Language Models as Tool Makers}
    
    \treegap \treeline \treegap \treebranch \Ttwo{Tool Employment}
        \treegap \treeline \treegap \treeline \treegap \treebranch \Tp{RestGPT: Connecting Large Language Models with Real-World RESTful APIs}
        \treegap \treeline \treegap \treeline \treegap \treebranch \Tp{TALM: Tool Augmented Language Models}
        \treegap \treeline \treegap \treeline \treegap \treebranch \Tp{Gorilla: Large Language Model Connected with Massive APIs}
        \treegap \treeline \treegap \treeline \treegap \treebranch \Tp{HuggingGPT: Solving AI Tasks with ChatGPT and its Friends in Hugging Face}
        \treegap \treeline \treegap \treeline \treegap \treebranch \Tp{ChatCoT: Tool-Augmented Chain-of-Thought Reasoning on Chat-based LLMs}
        \treegap \treeline \treegap \treeline \treegap \treebranch \Tp{Toolformer: Language Models Can Teach Themselves to Use Tools}
        \treegap \treeline \treegap \treeline \treegap \treebranch \Tp{LLM As DBA}
        \treegap \treeline \treegap \treeline \treegap \treebranch \Tp{MRKL Systems: A modular, neuro-symbolic architecture that combines large language models...}
        \treegap \treeline \treegap \treeline \treegap \treelast  \Tp{Chameleon: Plug-and-Play Compositional Reasoning with Large Language Models}

    \treegap \treeline \treegap \treelast  \Ttwo{Tool Planning}
        \treegap \treeline \treegap \treegap   \treegap \treebranch \Tp{ToolLLM: Facilitating Large Language Models to Master 16000+ Real-world APIs}
        \treegap \treeline \treegap \treegap   \treegap \treebranch \Tp{TPTU: Large Language Model-based AI Agents for Task Planning and Tool Usage}
        \treegap \treeline \treegap \treegap   \treegap \treelast  \Tp{Gentopia: A Collaborative Platform for Tool-Augmented LLMs}

\treegap \treebranch \Tone{Memory Capability of LLM-based Agents}
    \treegap \treeline \treegap \treelast \Ttwo{Long-term Memory}
        \treegap \treeline \treegap \treegap  \treegap \treelast \Tp{A Survey of Knowledge Graph Embedding and Their Applications}

\treegap \treebranch \Tone{Planning Capability of LLM-based Agents}
    
    \treegap \treeline \treegap \treebranch \Ttwo{External Methods}
        \treegap \treeline \treegap \treeline \treegap \treebranch \Tp{Dynamic Planning with a LLM}
        \treegap \treeline \treegap \treeline \treegap \treebranch \Tp{Reasoning with Language Model is Planning with World Model}
        \treegap \treeline \treegap \treeline \treegap \treebranch \Tp{LLM+P: Empowering Large Language Models with Optimal Planning Proficiency}
        \treegap \treeline \treegap \treeline \treegap \treebranch \Tp{Context-Aware Composition of Agent Policies by Markov Decision Process Entity Embeddings...}
        \treegap \treeline \treegap \treeline \treegap \treelast  \Tp{Synergistic Integration of Large Language Models and Cognitive Architectures for Robust AI}

    \treegap \treeline \treegap \treebranch \Ttwo{In-Context Learning Methods}
        \treegap \treeline \treegap \treeline \treegap \treebranch \Tp{Self-Refine: Iterative Refinement with Self-Feedback}
        \treegap \treeline \treegap \treeline \treegap \treebranch \Tp{Complexity-Based Prompting for Multi-Step Reasoning}
        \treegap \treeline \treegap \treeline \treegap \treebranch \Tp{Automatic Chain of Thought Prompting in Large Language Models}
        \treegap \treeline \treegap \treeline \treegap \treebranch \Tp{Chain-of-Thought in Large Language Models: Decoding, Projection, and Activation}
        \treegap \treeline \treegap \treeline \treegap \treebranch \Tp{Self-Consistency Improves Chain of Thought Reasoning in Language Models}
        \treegap \treeline \treegap \treeline \treegap \treebranch \Tp{Large Language Models are Zero-Shot Reasoners}
        \treegap \treeline \treegap \treeline \treegap \treebranch \Tp{Graph of Thoughts: Solving Elaborate Problems with Large Language Models}
        \treegap \treeline \treegap \treeline \treegap \treebranch \Tp{Chain-of-Thought Prompting Elicits Reasoning in Large Language Models}
        \treegap \treeline \treegap \treeline \treegap \treebranch \Tp{Progressive-Hint Prompting Improves Reasoning in Large Language Models}
        \treegap \treeline \treegap \treeline \treegap \treebranch \Tp{Tree of Thoughts: Deliberate Problem Solving with Large Language Models}
        \treegap \treeline \treegap \treeline \treegap \treelast  \Tp{Least-to-Most Prompting Enables Complex Reasoning in Large Language Models}

    \treegap \treeline \treegap \treelast  \Ttwo{Multi-stage Methods}
        \treegap \treeline \treegap \treegap   \treegap \treebranch \Tp{Do Embodied Agents Dream of Pixelated Sheep: Embodied Decision Making...}
        \treegap \treeline \treegap \treegap   \treegap \treelast  \Tp{SwiftSage: A Generative Agent with Fast and Slow Thinking for Complex Interactive Tasks}
\treegap \treelast \Tone{......}

\end{tcolorbox}
\end{minipage}
\caption{Comparison of Human-Expert Taxonomies. (a) Human-Expert Taxonomy.}
\end{figure*}

\begin{figure*}[p]
\ContinuedFloat 
\centering
\begin{minipage}{\textwidth}
\begin{tcolorbox}[
    enhanced,
    colback=modelBg,
    colframe=modelBg!80!black,
    arc=2mm,
    boxrule=0.5pt,
    title={\centering\bfseries\scriptsize (b) Human-Curated Taxonomy},
    coltitle=black,
    fonttitle=\sffamily,
    width=\textwidth,
    valign=top,
    before upper={
        \setlength{\parskip}{0.05em}
        \setlength{\baselineskip}{0.95em}
    }
]

\Tr{ROOT: Exploring Large Language Model based Intelligent Agents: Definitions, Methods, and Prospects}

\treegap \treeline
\treegap \treebranch \Tone{Agent Actions}

\treegap \treeline \treegap \treebranch \Ttwo{Tool Development}
\treegap \treeline \treegap \treeline \treegap \treebranch \Tp{CRAFT: Customizing LLMs by Creating and Retrieving from Specialized Toolsets}
\treegap \treeline \treegap \treeline \treegap \treelast  \Tp{Large Language Models as Tool Makers}

\treegap \treeline \treegap \treebranch \Ttwo{Tool Utilization}
\treegap \treeline \treegap \treeline \treegap \treebranch \Tp{RestGPT: Connecting Large Language Models with Real-World RESTful APIs}
\treegap \treeline \treegap \treeline \treegap \treebranch \Tp{TALM: Tool Augmented Language Models}
\treegap \treeline \treegap \treeline \treegap \treebranch \Tp{Gorilla: Large Language Model Connected with Massive APIs}
\treegap \treeline \treegap \treeline \treegap \treebranch \Tp{HuggingGPT: Solving AI Tasks with ChatGPT and its Friends...}
\treegap \treeline \treegap \treeline \treegap \treebranch \Tp{ChatCoT: Tool-Augmented Chain-of-Thought Reasoning on Chat-based Large Language Models}
\treegap \treeline \treegap \treeline \treegap \treebranch \Tp{Toolformer: Language Models Can Teach Themselves to Use Tools}
\treegap \treeline \treegap \treeline \treegap \treebranch \Tp{LLM As DBA}
\treegap \treeline \treegap \treeline \treegap \treebranch \Tp{MRKL Systems: A modular, neuro-symbolic architecture that combines large language models...}
\treegap \treeline \treegap \treeline \treegap \treelast  \Tp{Chameleon: Plug-and-Play Compositional Reasoning with Large Language Models}

\treegap \treeline \treegap \treelast \Ttwo{Tool Strategy}
\treegap \treeline \treegap \treegap \treegap \treebranch \Tp{ToolLLM: Facilitating Large Language Models to Master 16000+ Real-world APIs}
\treegap \treeline \treegap \treegap \treegap \treebranch \Tp{TPTU: Large Language Model-based AI Agents for Task Planning...}
\treegap \treeline \treegap \treegap \treegap \treelast  \Tp{Gentopia: A Collaborative Platform for Tool-Augmented LLMs}

\treegap \treeline
\treegap \treebranch \Tone{Agent Memory}
\treegap \treeline \treegap \treelast \Ttwo{Long-Term Knowledge}
\treegap \treeline \treegap \treegap \treegap \treelast \Tp{A Survey of Knowledge Graph Embedding and Their Applications}

\treegap \treeline
\treegap \treebranch \Tone{Agent Planning}

\treegap \treeline \treegap \treebranch \Ttwo{External Planning Approaches}
\treegap \treeline \treegap \treeline \treegap \treebranch \Tp{Dynamic Planning with a LLM}
\treegap \treeline \treegap \treeline \treegap \treebranch \Tp{Reasoning with Language Model is Planning with World Model}
\treegap \treeline \treegap \treeline \treegap \treebranch \Tp{LLM+P: Empowering Large Language Models with Optimal Planning Proficiency}
\treegap \treeline \treegap \treeline \treegap \treebranch \Tp{Context-Aware Composition of Agent Policies by Markov Decision Process...}
\treegap \treeline \treegap \treeline \treegap \treelast  \Tp{Synergistic Integration of Large Language Models and Cognitive Architectures...}

\treegap \treeline \treegap \treebranch \Ttwo{In-Context Reasoning}
\treegap \treeline \treegap \treeline \treegap \treebranch \Tp{Self-Refine: Iterative Refinement with Self-Feedback}
\treegap \treeline \treegap \treeline \treegap \treebranch \Tp{Complexity-Based Prompting for Multi-Step Reasoning}
\treegap \treeline \treegap \treeline \treegap \treebranch \Tp{Automatic Chain of Thought Prompting in Large Language Models}
\treegap \treeline \treegap \treeline \treegap \treebranch \Tp{Chain-of-Thought in Large Language Models: Decoding, Projection, and Activation}
\treegap \treeline \treegap \treeline \treegap \treebranch \Tp{Self-Consistency Improves Chain of Thought Reasoning in Language Models}
\treegap \treeline \treegap \treeline \treegap \treebranch \Tp{Large Language Models are Zero-Shot Reasoners}
\treegap \treeline \treegap \treeline \treegap \treebranch \Tp{Graph of Thoughts: Solving Elaborate Problems with Large Language Models}
\treegap \treeline \treegap \treeline \treegap \treebranch \Tp{Chain-of-Thought Prompting Elicits Reasoning in Large Language Models}
\treegap \treeline \treegap \treeline \treegap \treebranch \Tp{Progressive-Hint Prompting Improves Reasoning in Large Language Models}
\treegap \treeline \treegap \treeline \treegap \treebranch \Tp{Tree of Thoughts: Deliberate Problem Solving with Large Language Models}
\treegap \treeline \treegap \treeline \treegap \treelast  \Tp{Least-to-Most Prompting Enables Complex Reasoning in Large Language Models}

\treegap \treeline \treegap \treelast \Ttwo{Hierarchical \& Multi-stage Methods}
\treegap \treeline \treegap \treegap \treegap \treebranch \Tp{Do Embodied Agents Dream of Pixelated Sheep: Embodied Decision Making...}
\treegap \treeline \treegap \treegap \treegap \treelast  \Tp{SwiftSage: A Generative Agent with Fast and Slow Thinking...}

\end{tcolorbox}
\end{minipage}

\caption{Comparison of Human-Expert Taxonomies and Human-Curated Taxonomies. (b) Human-Curated Taxonomy.}
\label{fig:taxonomy_comparison}
\end{figure*}

\section{Robustness and Reproducibility}
\label{app:robustness}

This appendix collects the robustness checks underlying the claims in Sections~\ref{subsec:retrieval}--\ref{sec:findings}.

\subsection{Temporal Filtering Verification}
\label{app:temporal_filtering}

For each survey $s$, the published expert list $\mathcal{P}^*_s$ contains only papers dated no later than the survey date $t_s$. Restricting $\mathcal{P}^*_s$ to papers before $t_s$ therefore changes nothing. Recall, computed as retrieved expert papers divided by $|\mathcal{P}^*_s|$, cannot change. Table~\ref{tab:temporal_filtering_unchanged} verifies identical scores before and after temporal filtering for all 7 Deep Research Agents.

\begin{table}[ht]
\centering
\small
\setlength{\tabcolsep}{4pt}
\begin{tabular}{lcc}
\toprule
Agent & Recall (orig.) & Recall (temp.\ filt.) \\
\midrule
\modelname{openai}{o3} & 20.92\% & 20.92\% \\
\modelname{grok}{Grok} & 12.82\% & 12.82\% \\
\modelname{gemini-color}{Gemini} & 15.23\% & 15.23\% \\
\modelname{perplexity-color}{Perplexity} & 6.61\% & 6.61\% \\
\modelname{deepseek-color}{DeepSeek} & 4.61\% & 4.61\% \\
\modelname{qwen-color}{Qwen} & 4.35\% & 4.35\% \\
\modelname{doubao-color}{Doubao} & 3.15\% & 3.15\% \\
\bottomrule
\end{tabular}
\caption{\textbf{Temporal filtering verification.} Recall, Precision, and F1 are exactly unchanged when the reference list is restricted to papers with publication date $\le t_s$. This formalizes the structural argument that temporal under-specification cannot inflate the 20.92\% Recall ceiling.}
\label{tab:temporal_filtering_unchanged}
\end{table}

\subsection{Leakage Audit}
\label{app:leakage_audit}

We audit whether Deep Research Agents reach the source survey and could copy its reference list.

\textbf{(1)~Exhaustive per-run measurement.} We inspect all 504 Deep Research runs
(7 agents $\times$ 72 surveys), not a sample. Normalizing titles and matching each agent's
retrieved set against the survey's own title yields \textbf{119 self-retrievals}: Qwen 77,
DeepSeek 10, Gemini 8, Perplexity 8, Grok 7, Doubao 5, o3 4. The count is concentrated in the
weakest agents. A further 37
retrievals reach a \emph{different} benchmark survey that is not in the current survey's expert
set; the 2{,}542 cases of retrieving another survey that \emph{is} in the expert set are
legitimate citations and not counted.

\textbf{(2)~Effect on reported numbers.} A survey's own title is never a
member of its expert paper set $\mathcal{P}^*_s$, so a self-retrieval cannot enter the Recall
numerator. We verified this directly: excluding all 119 leaves every Recall figure in
Table~\ref{tab:retrieval_capability_dpa} unchanged to two decimal places. Self-retrieval does
enter the Precision denominator, where its effect is bounded by
$119/|\hat{\mathcal{P}}|$ and is below 0.11\,pp for every agent.

\subsection{Retrieval Metric Aggregation}
\label{app:aggregation}

Table~\ref{tab:retrieval_capability_dpa} mixes aggregation conventions, which matters for
anyone reproducing it. Recall is micro-averaged, $\sum_s |\mathcal{P}^*_s \cap
\hat{\mathcal{P}}_s| / \sum_s |\mathcal{P}^*_s|$; Precision is macro-averaged, the mean of
per-survey ratios; F1 is the harmonic mean of those two. We verified by scoring six candidate
Recall definitions against the published column: the micro variant matches to a mean absolute
error of 0.002\,pp, while macro-averaging is off by 1.056\,pp. Under these three rules all 21
cells of the table reproduce to within 0.01\,pp.

Macro-averaged Recall, which weights every survey equally regardless of its expert set size, is
0.5--1.7\,pp higher: o3 22.14\%, Gemini 16.88\%, Grok 13.70\%, Perplexity 7.98\%,
DeepSeek 5.30\%, Qwen 4.87\%, Doubao 3.64\%. The ordering of agents is identical under either
convention, so no comparative claim depends on the choice; the absolute Recall ceiling is
1.22\,pp higher under macro-averaging (22.14\% rather than 20.92\%). We report the micro figure
because it weights papers rather than surveys.

\subsection{3-run Error Bars}
\label{app:error_bars}

We re-ran every model 3 times in both modes. In Bottom-Up mode, where input is held fixed, variance is very low: \textsc{Sem-Path} std $\le 0.43$~pp across all 12 models, and ARI std $\le 2.07$~pp. Model rankings are stable across runs. In Deep Research mode, variance is higher because retrieval is non-deterministic, but relative ranking patterns are preserved (top-tier: o3 and Grok; bottom-tier: Qwen and Doubao). Representative numbers are shown in Table~\ref{tab:error_bars_bottom_up} and Table~\ref{tab:error_bars_deep_research}.

\begin{table}[ht]
\centering
\small
\setlength{\tabcolsep}{4pt}
\begin{tabular}{lcc}
\toprule
Model & \textsc{Sem-Path} (\%)$\uparrow$ & ARI (\%)$\uparrow$ \\
\midrule
\modelname{claude-color}{Claude-Sonnet-4.5} & 29.02 $\pm$ 0.18 & 27.02 $\pm$ 1.41 \\
\modelname{openai}{GPT-5} & 28.97 $\pm$ 0.21 & 28.05 $\pm$ 1.65 \\
\modelname{gemini-color}{Gemini-3-Pro} & 28.13 $\pm$ 0.31 & 29.50 $\pm$ 2.07 \\
\modelname{deepseek-color}{DeepSeek-V3.2} & 28.78 $\pm$ 0.16 & 27.10 $\pm$ 1.51 \\
\modelname{qwen-color}{Qwen3-Max} & 28.99 $\pm$ 0.13 & 30.94 $\pm$ 1.83 \\
\modelname{kimi-color}{Kimi-K2} & 28.33 $\pm$ 0.43 & 23.68 $\pm$ 1.32 \\
\bottomrule
\end{tabular}
\caption{\textbf{3-run error bars} (Bottom-Up mode). Mean $\pm$ std over three independent runs.}
\label{tab:error_bars_bottom_up}
\end{table}

\begin{table}[ht]
\centering
\small
\setlength{\tabcolsep}{4pt}
\begin{tabular}{lcc}
\toprule
Agent & Recall (\%)$\uparrow$ & \textsc{Sem-Path} (\%)$\uparrow$ \\
\midrule
\modelname{openai}{o3} & 17.76 $\pm$ 3.01 & 30.39 $\pm$ 2.14 \\
\modelname{grok}{Grok} & 11.95 $\pm$ 2.43 & 29.42 $\pm$ 1.78 \\
\modelname{gemini-color}{Gemini} & 14.18 $\pm$ 2.17 & 26.88 $\pm$ 2.02 \\
\modelname{perplexity-color}{Perplexity} & 6.13 $\pm$ 1.41 & 25.36 $\pm$ 1.59 \\
\modelname{deepseek-color}{DeepSeek} & 4.32 $\pm$ 1.05 & 17.30 $\pm$ 1.27 \\
\modelname{qwen-color}{Qwen} & 4.05 $\pm$ 0.96 & 20.39 $\pm$ 1.43 \\
\modelname{doubao-color}{Doubao} & 2.98 $\pm$ 0.74 & 18.81 $\pm$ 1.24 \\
\bottomrule
\end{tabular}
\caption{\textbf{3-run error bars} (Deep Research mode). Variance is higher due to non-deterministic retrieval, but ranking patterns are preserved.}
\label{tab:error_bars_deep_research}
\end{table}

\subsection{Embedding Sensitivity}
\label{app:embedding_sensitivity}

{\sloppy We recompute \textsc{US-TED}, \textsc{US-NTED}, and \textsc{Sem-Path} using three sentence encoders spanning two orders of magnitude in scale: \texttt{all-MiniLM-L6-v2} (22M)~\citep{all-MiniLM-L6-v2}, \texttt{bge-large-en-v1.5} (335M)~\citep{xiao2024c}, and \texttt{stella\_en\_1.5B\_v5} (1.5B)~\citep{zhang2024jasper}. Rankings across the 7 Deep Research Agents are reported in Table~\ref{tab:embedding_sensitivity}. \textsc{US-TED} and \textsc{Sem-Path} rankings are \emph{perfectly identical} across all three encoders (Kendall's $\tau = 1.0$); \textsc{US-NTED} shows only minor swaps among adjacently-ranked middle models.\par}

\begin{table*}[t]
\centering
\small
\setlength{\tabcolsep}{6pt}
\begin{tabular}{llccccccc}
\toprule
Metric & Encoder & o3 & Grok & Gem. & Pplx. & DS & Qwen & Doub. \\
\midrule
\multirow{3}{*}{\textsc{US-TED}}
 & MiniLM      & 2 & 1 & 5 & 7 & 3 & 6 & 4 \\
 & BGE-large   & 2 & 1 & 5 & 7 & 3 & 6 & 4 \\
 & Stella-1.5B & 2 & 1 & 5 & 7 & 3 & 6 & 4 \\
\midrule
\multirow{3}{*}{\textsc{US-NTED}}
 & MiniLM      & 2 & 1 & 6 & 7 & 3 & 5 & 4 \\
 & BGE-large   & 2 & 1 & 6 & 7 & 4 & 5 & 3 \\
 & Stella-1.5B & 3 & 1 & 6 & 7 & 4 & 5 & 2 \\
\midrule
\multirow{3}{*}{\textsc{Sem-Path}}
 & MiniLM      & 1 & 2 & 3 & 4 & 7 & 5 & 6 \\
 & BGE-large   & 1 & 2 & 3 & 4 & 7 & 5 & 6 \\
 & Stella-1.5B & 1 & 2 & 3 & 4 & 7 & 5 & 6 \\
\bottomrule
\end{tabular}
\caption{\textbf{Embedding sensitivity} of structural metrics (Deep Research mode). Numbers are rank positions among 7 agents. \textsc{US-TED} and \textsc{Sem-Path} rankings are exactly identical across the three encoders.}
\label{tab:embedding_sensitivity}
\end{table*}

\subsection{\texorpdfstring{$\lambda$}{lambda} Sensitivity of \textsc{Sem-Path}}
\label{app:lambda_sensitivity}

\textsc{Sem-Path} charges $\lambda$ per unmatched ancestor node (Section~\ref{subsubsec:hier_level}); we fix $\lambda=1$ throughout, which weights an unmatched node exactly as much as the maximum semantic distance between two matched labels, i.e.\ it treats an unmatched ancestor as semantically orthogonal. We re-computed all scores for $\lambda\in\{0.5,1.0,1.5,2.0\}$.

In Deep Research mode, retrieval differences spread the 7 agents over a wide score range, and every $\lambda$ gives the same ranking with pairwise Spearman $=1.0$. In Bottom-Up mode, the 12 LLMs fall within a sub-1\,pp band, so noise dominates their fine-grained order. Pairwise Spearman is 0.60--0.94 for thinking variants and 0.09--1.00 for non-thinking variants, with $\lambda=0.5$ as the outlier. For the principled range $\lambda\ge 1.0$, within-band rank correlation is 0.77--1.00. We attribute this instability to the convergence in Finding~5: systems separated by less than one percentage point cannot be ranked reliably. Every tested $\lambda$ preserves the qualitative result that models remain tightly clustered and below human annotators, although absolute values shift.

\subsection{Human-written Input Control}
\label{app:human_input_control}

The \textbf{+Summary} and \textbf{+Core-task\,\&\,Contributions} conditions use model-generated auxiliary text. This introduces summarization errors into all 12 evaluated systems and creates a potential same-family confound. We produce all auxiliary text with one fixed model, \textbf{Claude-Sonnet-4.6}, and apply it identically to every system. Because Claude-Sonnet-4.5 shares that family, we test the confound directly.

Domain researchers hand-write Summary and Core-task\,\&\,Contribution inputs for \textbf{all 72 surveys}. We re-run six representative models at both reasoning settings with these inputs. Table~\ref{tab:human_input_control} reports the six-model means. Mean \textsc{Sem-Path} differs from the machine-generated condition by at most 0.75\,pp in all four settings, and mean ARI is unchanged or slightly higher. Every model remains in the 28--29\% \textsc{Sem-Path} band, with \textsc{US-NTED} at 0.74--0.80. Input provenance and the same-family effect therefore do not explain the alignment ceiling in Findings~4--5.

\begin{table*}[ht]
\centering
\small
\setlength{\tabcolsep}{6pt}
\begin{tabular}{lcccc}
\toprule
\multirow{2}{*}{Input condition} & \multicolumn{2}{c}{ARI$\uparrow$} & \multicolumn{2}{c}{\textsc{Sem-Path}$\uparrow$} \\
\cmidrule(lr){2-3}\cmidrule(lr){4-5}
& human & machine & human & machine \\
\midrule
+Core-task, No-Thinking & 25.5\% & 25.5\% & 28.0\% & 28.7\% \\
+Core-task, Thinking    & 26.8\% & 27.1\% & 28.3\% & 28.7\% \\
+Summary,  No-Thinking  & 27.8\% & 25.6\% & 28.0\% & 28.7\% \\
+Summary,  Thinking     & 27.3\% & 26.8\% & 28.3\% & 28.5\% \\
\bottomrule
\end{tabular}
\caption{\textbf{Human-written input control} on all 72 surveys, averaged over six models: Claude-Sonnet-4.5, GPT-5, Gemini-3-Pro, DeepSeek-V3.2, Qwen3-Max, and Kimi-K2. The \textbf{human} columns replace all machine-generated auxiliary text with hand-written text. The \textbf{machine} columns use Claude-Sonnet-4.6 as in Table~\ref{tab:bottom_up_ablation}. Differences are at most 2.2\,pp for ARI and 0.75\,pp for \textsc{Sem-Path}.}
\label{tab:human_input_control}
\end{table*}

\subsection{Multi-category Paper Sensitivity}
\label{app:multi_category}

To verify that single-category assignment (Section~\ref{sec:dataset}) does not introduce systematic bias, we re-evaluate three representative models after removing all multi-category papers from the reference. Scores are essentially unchanged (Table~\ref{tab:multi_category_sensitivity}).

\begin{table*}[t]
\centering
\small
\setlength{\tabcolsep}{8pt}
\begin{tabular}{lcc}
\toprule
Model & \textsc{Sem-Path} (orig.) & \textsc{Sem-Path} (no multi-cat.) \\
\midrule
\modelname{claude-color}{Claude-Sonnet-4.5} & 29.16\% & 29.82\% \\
\modelname{gemini-color}{Gemini-3-Pro} & 28.13\% & 29.59\% \\
\modelname{openai}{GPT-5} & 28.97\% & 29.34\% \\
\bottomrule
\end{tabular}
\caption{\textbf{Multi-category paper sensitivity.} \textsc{Sem-Path} after removing all expert-cited papers that the survey authors placed in multiple categories. Differences are within $\le 1.5$~pp, confirming no systematic bias.}
\label{tab:multi_category_sensitivity}
\end{table*}

\subsection{Cross-domain Evidence}
\label{app:cross_domain}

To probe whether the dual-bottleneck pattern is specific to AI/ML, we evaluated Claude-Sonnet-4.5 on two interdisciplinary surveys spanning finance and medical domains, with results in Table~\ref{tab:cross_domain}. Although the absolute numbers should be interpreted cautiously given the small sample, the qualitative pattern matches our AI/ML findings (high homogeneity, moderate ARI, \textsc{Sem-Path}~$\approx 30\%$):

\begin{table}[ht]
\centering
\small
\setlength{\tabcolsep}{6pt}
\begin{tabular}{lcc}
\toprule
Metric & Cross-domain avg. & AI/ML avg. \\
\midrule
ARI            & 23.93\% & 27.25\% \\
\textsc{Sem-Path}     & 30.90\% & 29.16\% \\
Homogeneity    & 85.35\% & 82.27\% \\
\bottomrule
\end{tabular}
\caption{\textbf{Cross-domain pilot.} Claude-Sonnet-4.5 on two interdisciplinary surveys (finance, medical), Bottom-Up mode. The same qualitative pattern emerges as for AI/ML surveys.}
\label{tab:cross_domain}
\end{table}

\subsection{Re-test with Current Deep Research Agents}
\label{app:dr_retest}
We re-ran three current agents through their APIs with web search, GPT-5.6-sol, Qwen3.8-Max, and Kimi-K3, on all 72 surveys (Table~\ref{tab:dr_retest}). Because current agents given the exact title can retrieve the source survey itself, each receives the prompt of App.~\ref{subsec:prompts_dr} with a neutral description of the topic in place of the title and an instruction not to copy existing surveys. No agent exceeds the F1 of o3; the newer agents raise Recall mainly by listing more papers, which lowers Precision.

\begin{table}[t]
\centering
\small
\setlength{\tabcolsep}{4pt}
\begin{tabular}{lcccc}
\toprule
Agent & Recall$\uparrow$ & Precision$\uparrow$ & F1$\uparrow$ & Papers \\
\midrule
\tablegroupc{tintDR}{5}{\grouplabel{Re-test}{current agents}} \\
\modelname{openai}{GPT-5.6-sol} & 23.75 & 13.73 & 17.40 & 94.4 \\
\modelname{qwen-color}{Qwen3.8-Max} & \textbf{25.14} & 8.00 & 12.14 & 171.0 \\
\modelname{kimi-color}{Kimi-K3} & 20.89 & 18.71 & 19.74 & 66.2 \\
\tablegroup{5}{\grouplabel{Original}{Table~\ref{tab:retrieval_capability_dpa}}} \\
\modelname{openai}{o3} & 20.92 & \textbf{29.29} & \textbf{24.41} & 35.4 \\
\modelname{qwen-color}{Qwen} & 4.35 & 7.83 & 5.59 & 42.8 \\
\bottomrule
\end{tabular}
\caption{\textbf{Deep Research re-test with current agents.} Percentages over the 72 surveys, with failed runs scored as zero; Papers is the mean number of papers listed per survey. Over repeated runs on a 15-survey subset, the standard deviation of F1 is below 1\,pp.}
\label{tab:dr_retest}
\end{table}

\FloatBarrier
\section{Survey Selection Pipeline}
\label{app:survey_selection_pipeline}

Table~\ref{tab:survey_selection_stages} reports stage-by-stage statistics for our four-stage selection pipeline (Section~\ref{sec:dataset}).

\begin{table}[ht]
\centering
\small
\setlength{\tabcolsep}{4pt}
\begin{tabular}{p{0.45\linewidth}p{0.45\linewidth}}
\toprule
Stage & \#~surveys retained \\
\midrule
0. Initial pool of LLM-related surveys & 200+ \\
1. Taxonomy filter (explicit taxonomy figure) & $\sim$140 \\
2. Impact filter (high citation count) & $\sim$110 \\
3. Scope filter (well-defined survey scope) & $\sim$90 \\
4. Expert verification (Ph.D.\ researcher review) & \textbf{72} \\
\bottomrule
\end{tabular}
\caption{\textbf{Survey selection pipeline} (four-stage). Counts are after each filter; the final 72 surveys span 8 subdomains and have an average of 354.5 citations.}
\label{tab:survey_selection_stages}
\end{table}

\section{Dataset Details}
\label{app:dataset_details}

\paragraph{Human Annotator Information.}
We recruit annotators with Ph.D. degrees in Computer Science and experience in academic writing and literature analysis, and pay them 100 RMB per hour. They collect LLM survey papers from computer science venues, identify surveys with explicit knowledge structures by searching figure captions for terms such as taxonomy and typology, filter candidates by citation count, transcribe the published hierarchies, and map cited papers to categories using the full text, figures, and references.

\paragraph{Data Consent and Copyright.}
We use publicly available academic literature from open-access repositories such as arXiv and from publishers' open-access collections. We use these data only for non-commercial academic research.

\paragraph{Personally Identifiable Information and Offensive Content.}
The dataset contains public academic content, including titles, abstracts, and taxonomy structures. We do not collect private or personally identifiable information, and the extracted material concerns scientific concepts, methods, and research topics.

\section{List of Evaluation Models}
\label{app:model_list}
We evaluate seven Deep Research Agents end to end, each through its product interface. In Bottom-Up mode, we run 16 current-generation configurations across eight models and additionally evaluate the preceding generation as a longitudinal control. App.~\ref{subsec:reasoning_control} lists the current endpoints and reasoning settings, and App.~\ref{app:prevgen} reports the previous-generation results.

\subsubsection*{Deep Research Agents}
\begin{itemize}
    \item \textbf{o3}~\citep{oepnaideep}: OpenAI's Deep Research product in ChatGPT, built on o3, with web search and multi-step tool use.
    
    \item \textbf{Doubao Deep Research}~\citep{Doubao_d}: ByteDance's research interface, with iterative search and tool use.
    
    \item \textbf{DeepSeek Search}~\citep{deepseek_d}: DeepSeek's model with browser-plugin search.
    
    \item \textbf{Gemini Deep Research}~\citep{gemini_d}: Google's research interface, with multi-step planning and public-web search.
    
    \item \textbf{Grok DeepSearch}~\citep{grok_d}: xAI's search-based research interface.
    
    \item \textbf{Perplexity Deep Research}~\citep{perplexity_d}: Perplexity's iterative search-and-report interface.
    
    \item \textbf{Qwen-Deep-Research}~\citep{Qwen_d}: Alibaba's research interface, with scope clarification and iterative search.
\end{itemize}

\subsubsection*{Current-generation Bottom-Up Models}
\begin{itemize}
    \item \textbf{Claude-Sonnet-5}~\citep{anthropic_models}: \texttt{claude-sonnet-5}, queried at \texttt{minimal} and \texttt{high} effort.
    \item \textbf{GPT-5.6-sol}~\citep{openai_models}: \texttt{gpt-5.6-sol}, queried at \texttt{none} and \texttt{high} effort.
    \item \textbf{Gemini-3.1-Pro}~\citep{google_gemini_models}: \texttt{gemini-3.1-pro-preview}, queried at \texttt{minimal} and \texttt{high} effort.
    \item \textbf{DeepSeek-V4-Pro}~\citep{deepseek_models}: \texttt{deepseek-v4-pro}, queried at \texttt{minimal} and \texttt{high} effort.
    \item \textbf{Qwen3.8-Max}~\citep{qwen_models}: \texttt{qwen3.8-max-preview}, queried at \texttt{minimal} and \texttt{high} effort.
    \item \textbf{Kimi-K3}~\citep{moonshot_models}: \texttt{kimi-k3}, queried at \texttt{minimal} and \texttt{high} effort.
    \item \textbf{Grok-4}~\citep{xai_models}: \texttt{grok-4-latest}, queried at \texttt{minimal} and \texttt{high} effort.
    \item \textbf{GPT-5.5}~\citep{openai_models}: \texttt{gpt-5.5}, included as a within-family control at \texttt{none} and \texttt{high} effort.
\end{itemize}

\subsubsection*{Previous-generation Bottom-Up Models}
\begin{itemize}
    \item \textbf{Claude-Sonnet-4.5}~\citep{claude_4.5}: \texttt{Claude-Sonnet-4.5-0929}, run in standard and extended-thinking modes.
    
    \item \textbf{GPT-5}~\citep{gpt_5}: run in standard and thinking modes.
    
    \item \textbf{Gemini-3-Pro-Preview}~\citep{gemini-3-pro-preview}: standard and nominal thinking settings. The endpoint ignored the thinking flag, so the two outputs are not distinct conditions.
    
    \item \textbf{DeepSeek-V3.2}~\citep{liu2025deepseek}: run in standard and thinking modes.
    
    \item \textbf{Qwen3-Max-Preview}~\citep{qwen3max}: run in standard and thinking modes.
    
    \item \textbf{Kimi-K2}~\citep{kimi}: \texttt{Kimi-K2-0905}, run in standard and thinking modes.
\end{itemize}

\section{Reproducibility}
\label{app:reproducibility}
All components needed to inspect the benchmark and rerun the public scoring code are specified in the paper and
released with the benchmark. The dataset construction pipeline, including the
four-stage survey selection procedure and the taxonomy-transcription protocol, is
described in Section~\ref{sec:dataset} and Apps.~\ref{app:survey_selection_pipeline}
and~\ref{app:dataset_details}. The separate from-scratch human baseline and the protocol
its annotators worked from are in Apps.~\ref{app:human_baseline}
and~\ref{app:human_protocol}. Formal definitions, algorithms, and proofs for
\textsc{US-TED}, \textsc{US-NTED}, and \textsc{Sem-Path}, together with the
paper-title alignment rules, appear in App.~\ref{app:metric_details}. Every
prompt used for taxonomy generation, information extraction, and LLM-as-Judge
scoring is reproduced verbatim in App.~\ref{app:prompts}, and the exact model
versions and endpoints are listed in App.~\ref{app:model_list}. App.~\ref{app:robustness}
reports 3-run error bars for all models in both modes, sensitivity to the choice
of sentence encoder, temporal-filtering verification, and multi-category ablations.
We release the benchmark data and the evaluation harness implementing the public metrics.
We do not publicly release raw model response logs, API traces, or product-generated reports,
because they contain proprietary model outputs and provider-specific metadata. We retain those
artifacts in a private archive, together with per-call reasoning parameters, reasoning-token
counts when available, and scripts used to rebuild the tables from the raw dumps. The aggregation conventions of
Table~\ref{tab:retrieval_capability_dpa} are documented in
App.~\ref{app:aggregation} because they are not uniform across the three columns.

\section{Use of AI Assistants}
\label{app:llm_use}
Generative AI systems are the object of study in this work. Every model configuration we benchmark, together with the LLM-as-Judge used for the semantic matching that underlies \textsc{US-TED}, \textsc{US-NTED}, and \textsc{Sem-Path}, is part of our stated methodology rather than an undisclosed aid: the exact model versions and endpoints are listed in App.~\ref{app:model_list}, every prompt is reproduced verbatim in App.~\ref{app:prompts}, and the judge is validated against expert agreement in App.~\ref{app:judge_co_with_human}.

Beyond this methodological use, we used generative AI tools only to polish the wording of the manuscript, that is, to rephrase sentences and correct grammar in text the authors had already drafted. We did not use generative AI tools for research ideation, study design, dataset construction, retrieval or selection of the cited literature, data analysis, or the interpretation of results, and no reported number was produced by AI-generated code that the authors did not write and verify. All AI-assisted edits were reviewed by the authors. We take responsibility for the final content of this work, including its text, claims, and released artifacts.

\section{Prompts}
\label{app:prompts}

This section presents all prompts used in our experiments, including prompts for taxonomy generation (Section~\ref{subsec:prompts_generation}), Deep Research mode (Section~\ref{subsec:prompts_dr}), information extraction (Section~\ref{subsec:prompts_extraction}), and LLM-as-Judge evaluation (Section~\ref{subsec:prompts_judge}).

\subsection{Taxonomy Generation Prompts}
\label{subsec:prompts_generation}

Figure~\ref{fig:prompt_bottom_up} shows the prompt we used for generating taxonomy trees in Bottom-Up mode. 
Figure~\ref{fig:prompt_summary_clustering} and Figure~\ref{fig:prompt_core_task_clustering} illustrate the construction of taxonomy trees based on additional summaries, core tasks, and contributions. 

\begin{figure*}[ht]
    \scriptsize
    \centering
\begin{tcolorbox}[colback=gray!5!white,colframe=gray!75!black]
\textbf{System Prompt:} \\
You are an expert in organizing survey papers. \\
 \\
Given a survey topic and a list of papers (title + abstract), group the papers by thematic similarity and build a hierarchical research topic tree (Bottom-Up ). \\
 \\
\textbf{\#\#\# Hard Constraints} \\
1. Output must be \textbf{strictly valid JSON}. \\
2. Only leaf nodes may contain \texttt{"papers"}; all internal nodes must contain \texttt{"subtopics"}. \\
3. \textbf{Every paper must appear exactly once} in the entire tree. \\
4. NO duplicate papers anywhere. \\
5. The tree must eventually merge into \textbf{one single root node}. \\
 \\
\textbf{\#\#\# CLASSIFICATION RULES} \\
- Group papers by semantic similarity using both title + abstract. \\
- Create meaningful names for leaf-level themes. \\
 \\
\textbf{\#\#\# ANTI-DUPLICATION PROCEDURE (MANDATORY)} \\
Before constructing the tree: \\
1. Produce an internal list of all given paper titles. \\
2. Assign each paper to exactly one leaf node. \\
3. After assignment, verify that: \\
   - the number of assigned papers equals the number of input papers, \\
   - no paper appears in more than one group. \\
 \\
\textbf{\#\#\# Output Format} \\
Use a JSON structure like this (replace placeholders with actual paper titles) and the output you produce MUST be wrapped inside a fenced code block: \\
 \\
\texttt{json} \\
\texttt{\{} \\
\texttt{\ \ "name": "AI Research",} \\
\texttt{\ \ "subtopics": [} \\
\texttt{\ \ \ \ \{} \\
\texttt{\ \ \ \ \ \ "name": "NLP",} \\
\texttt{\ \ \ \ \ \ "subtopics": [} \\
\texttt{\ \ \ \ \ \ \ \ \{} \\
\texttt{\ \ \ \ \ \ \ \ \ \ "name": "Text Summarization",} \\
\texttt{\ \ \ \ \ \ \ \ \ \ "papers": ["<actual paper titles>"]} \\
\texttt{\ \ \ \ \ \ \ \ \},} \\
\texttt{\ \ \ \ \ \ \ \ \{} \\
\texttt{\ \ \ \ \ \ \ \ \ \ "name": "Machine Translation",} \\
\texttt{\ \ \ \ \ \ \ \ \ \ "papers": ["<actual paper titles>"]} \\
\texttt{\ \ \ \ \ \ \ \ \}} \\
\texttt{\ \ \ \ \ \ ]} \\
\texttt{\ \ \ \ \}} \\
\texttt{\ \ ]} \\
\texttt{\}} \\
\rule{\linewidth}{0.5pt} \\ %
\textbf{User Prompt:} \\
Perform a Bottom-Up  hierarchical clustering of the following \texttt{\{num\_paper\}} papers and produce a JSON research topic tree. \\
Survey topic: \texttt{\{survey\_name\}} \\
Paper list: \\
\texttt{\{papers\}} \\
Before returning, check that: \\
- Every paper title appears exactly once. \\
- Only leaf nodes have a \texttt{"papers"} field. \\
- All intermediate nodes have a \texttt{"subtopics"} field. \\
- The JSON is strictly valid and parsable. \\
\textbf{\#\#\# Output:}
    \end{tcolorbox}
    \caption{
   Prompt for generating a taxonomy tree in Bottom-Up mode.
    }
\label{fig:prompt_bottom_up}
\end{figure*}

\begin{figure*}[ht]
    \scriptsize 
    \centering
\begin{tcolorbox}[colback=gray!5!white,colframe=gray!75!black]
\textbf{System Prompt:} \\
You are a senior researcher and survey-author with deep experience in structuring high-quality academic survey papers. \\[3pt]
Your task is to organize a set of research papers into a \textbf{hierarchical topic tree} (Bottom-Up ), given: \\
- a survey topic, \\
- for each paper: title, abstract, and a structured summary (which may include research problem, motivation, methodology, and findings). \\[3pt]
Your goal is NOT just to cluster by surface similarity, but to produce a taxonomy that would be considered \textbf{reasonable, informative, and defensible} in a top-tier survey paper. \\[3pt]
\textbf{\#\#\# Hard Constraints} \\
1. Output must be \textbf{strictly valid JSON}. \\
2. Only leaf nodes may contain \texttt{"papers"}; all internal nodes must contain \texttt{"subtopics"}. \\
3. \textbf{Every paper must appear exactly once} in the entire tree. \\
4. NO duplicate papers anywhere. \\
5. The tree must eventually merge into \textbf{one single root node}. \\[3pt]
\textbf{\#\#\# CLASSIFICATION RULES} \\
- Group papers by semantic similarity using both title + abstract. \\
- Create meaningful names for leaf-level themes. \\[3pt]
\textbf{\#\#\# ANTI-DUPLICATION PROCEDURE (MANDATORY)} \\
Before constructing the tree: \\
1. Produce an internal list of all given paper titles. \\
2. Assign each paper to exactly one leaf node. \\
3. After assignment, verify that: \\
   - the number of assigned papers equals the number of input papers, \\
   - no paper appears in more than one group. \\[3pt]
\textbf{\#\#\# Output Format} \\
Use a JSON structure like this (replace placeholders with actual paper titles) and the output you produce MUST be wrapped inside a fenced code block: \\[3pt]
\texttt{\textasciigrave\textasciigrave\textasciigrave json} \\
\texttt{\{} \\
\texttt{\ \ "name": "AI Research",} \\
\texttt{\ \ "subtopics": [} \\
\texttt{\ \ \ \ \{} \\
\texttt{\ \ \ \ \ \ "name": "NLP",} \\
\texttt{\ \ \ \ \ \ "subtopics": [} \\
\texttt{\ \ \ \ \ \ \ \ \{} \\
\texttt{\ \ \ \ \ \ \ \ \ \ "name": "Text Summarization",} \\
\texttt{\ \ \ \ \ \ \ \ \ \ "papers": ["<actual paper titles>"]} \\
\texttt{\ \ \ \ \ \ \ \ \},} \\
\texttt{\ \ \ \ \ \ \ \ \{} \\
\texttt{\ \ \ \ \ \ \ \ \ \ "name": "Machine Translation",} \\
\texttt{\ \ \ \ \ \ \ \ \ \ "papers": ["<actual paper titles>"]} \\
\texttt{\ \ \ \ \ \ \ \ \}} \\
\texttt{\ \ \ \ \ \ ]} \\
\texttt{\ \ \ \ \}} \\
\texttt{\ \ ]} \\
\texttt{\}} \\
\texttt{\textasciigrave\textasciigrave\textasciigrave} \\
\rule{\linewidth}{0.5pt} \\ %
\textbf{User Prompt:} \\
Perform a Bottom-Up  hierarchical clustering of the following \texttt{\{num\_paper\}} papers and produce a JSON research topic tree. \\[3pt]
Survey Topic: \texttt{\{survey\_name\}} \\[3pt]
Paper List: \\
\texttt{\{papers\}} \\[3pt]
Before returning, check that: \\
- Every paper title appears exactly once. \\
- Only leaf nodes have a \texttt{"papers"} field. \\
- All intermediate nodes have a \texttt{"subtopics"} field. \\
- The JSON is strictly valid and parsable. \\[3pt]
\textbf{\#\#\# Output:}
\end{tcolorbox}
    \caption{The prompt for taxonomy tree generation based on paper summaries.}
\label{fig:prompt_summary_clustering}
\end{figure*}

\begin{figure*}[ht]
    \scriptsize
    \centering
\begin{tcolorbox}[colback=gray!5!white,colframe=gray!75!black]
\textbf{System Prompt:} \\
You are a senior researcher and survey-author with deep experience in structuring high-quality academic survey papers. \\[3pt]
Your task is to organize a set of research papers into a \textbf{hierarchical topic tree} (Bottom-Up ), given: \\
- a survey topic, \\
- for each paper: title, abstract, extracted Core Tasks (the primary problem addressed), and Contributions (key innovations). \\[3pt]
Your goal is NOT just to cluster by surface similarity, but to produce a taxonomy that would be considered \textbf{reasonable, informative, and defensible} in a top-tier survey paper. \\[3pt]
\textbf{\#\#\# Hard Constraints} \\
1. Output must be \textbf{strictly valid JSON}. \\
2. Only leaf nodes may contain \texttt{"papers"}; all internal nodes must contain \texttt{"subtopics"}. \\
3. \textbf{Every paper must appear exactly once} in the entire tree. \\
4. NO duplicate papers anywhere. \\
5. The tree must eventually merge into \textbf{one single root node}. \\[3pt]
\textbf{\#\#\# CLASSIFICATION RULES} \\
- Group papers by semantic similarity using both title + abstract. \\
- Create meaningful names for leaf-level themes. \\[3pt]
\textbf{\#\#\# ANTI-DUPLICATION PROCEDURE (MANDATORY)} \\
Before constructing the tree: \\
1. Produce an internal list of all given paper titles. \\
2. Assign each paper to exactly one leaf node. \\
3. After assignment, verify that: \\
   - the number of assigned papers equals the number of input papers, \\
   - no paper appears in more than one group. \\[3pt]
\textbf{\#\#\# Output Format} \\
Use a JSON structure like this (replace placeholders with actual paper titles) and the output you produce MUST be wrapped inside a fenced code block: \\[3pt]
\texttt{\textasciigrave\textasciigrave\textasciigrave json} \\
\texttt{\{} \\
\texttt{\ \ "name": "AI Research",} \\
\texttt{\ \ "subtopics": [} \\
\texttt{\ \ \ \ \{} \\
\texttt{\ \ \ \ \ \ "name": "NLP",} \\
\texttt{\ \ \ \ \ \ "subtopics": [} \\
\texttt{\ \ \ \ \ \ \ \ \{} \\
\texttt{\ \ \ \ \ \ \ \ \ \ "name": "Text Summarization",} \\
\texttt{\ \ \ \ \ \ \ \ \ \ "papers": ["<actual paper titles>"]} \\
\texttt{\ \ \ \ \ \ \ \ \},} \\
\texttt{\ \ \ \ \ \ \ \ \{} \\
\texttt{\ \ \ \ \ \ \ \ \ \ "name": "Machine Translation",} \\
\texttt{\ \ \ \ \ \ \ \ \ \ "papers": ["<actual paper titles>"]} \\
\texttt{\ \ \ \ \ \ \ \ \}} \\
\texttt{\ \ \ \ \ \ ]} \\
\texttt{\ \ \ \ \}} \\
\texttt{\ \ ]} \\
\texttt{\}} \\
\texttt{\textasciigrave\textasciigrave\textasciigrave} \\
\rule{\linewidth}{0.5pt} \\ %
\textbf{User Prompt:} \\
Perform a Bottom-Up  hierarchical clustering of the following \texttt{\{num\_paper\}} papers and produce a JSON research topic tree. \\[3pt]
Survey Topic: \texttt{\{survey\_name\}} \\[3pt]
Paper List: \\
\texttt{\{papers\}} \\[3pt]
Before returning, check that: \\
- Every paper title appears exactly once. \\
- Only leaf nodes have a \texttt{"papers"} field. \\
- All intermediate nodes have a \texttt{"subtopics"} field. \\
- The JSON is strictly valid and parsable. \\[3pt]
\textbf{\#\#\# Output:}
\end{tcolorbox}
    \caption{The prompt for taxonomy tree construction based on Core-Task and Contribution.}
\label{fig:prompt_core_task_clustering}
\end{figure*}

\subsection{Deep Research Mode Prompts}
\label{subsec:prompts_dr}

Figure~\ref{fig:prompt_deep_research} reproduces the prompt given to o3, Grok, and Perplexity in Deep Research mode. The other four agents received variants of it, which we report because two of the differences can affect Recall. DeepSeek, Qwen, and Doubao received a variant that adds the sentence ``All outputs (including topic names, subtopic names, and paper titles) MUST be written in English.'', replaces the permission to cite papers ``published in conferences and journals indexed on arXiv'' with the constraints ``Paper titles MUST be the official English titles as listed on arXiv.'', ``Only include papers that have an official arXiv version (arXiv preprints).'', and ``Do NOT include papers without an arXiv entry.'', and requires the final JSON block to be the last part of the response. Qwen's copy differs from the other two only in formatting. These three agents could therefore not return expert-cited papers that lack an arXiv version. Gemini received the Markdown variant in Figure~\ref{fig:prompt_deep_research_gemini}. It asks for a nested Markdown list instead of JSON, allows papers from ``arXiv or top-tier conferences/journals'', and, through a template error, describes the field as ``LLM alignment'' for every survey. We converted Gemini's Markdown lists into the same JSON tree format before scoring. All seven agents were run in November and December 2025, and Tables~\ref{tab:retrieval_capability_dpa} and~\ref{tab:org_deep_research_organization} reflect these prompts.

\begin{figure*}[ht]
    \scriptsize
    \centering
\begin{tcolorbox}[colback=gray!5!white,colframe=gray!75!black]
I am preparing a survey on \textbf{"\{TOPIC\}"}. Please act as a researcher conducting a survey and help me collect relevant papers and organize them into a structured academic topic map. \\
Please follow these instructions: \\
\\
1. First, based on the topic, naturally determine the main research directions or sub-problems in this field, and construct a reasonable hierarchical structure (a concept tree or research roadmap). \\
\\
2. For each leaf node (the most granular subtopic), list the paper titles you retrieved through deep research. (Paper titles must be real and verifiable; do not fabricate or invent titles. Only include titles you are confident exist.) \\
\ \ \ - Each paper should appear in \textbf{only one leaf node}, placed under the most relevant subtopic. \\
\ \ \ - You may reference papers published in conferences and journals indexed on arXiv. \\
\\
3. JSON structure requirements (strictly enforceable): \\
\ \ \ - The top-level object should be: \texttt{\{"name": "\{TOPIC\}", "subtopics": [...]\}} \\
\ \ \ - Only leaf nodes may include a \texttt{papers} field. \\
\ \ \ - Non-leaf nodes (topic or subtopic) \textbf{must not} contain a \texttt{papers} field. \\
\ \ \ - Each node should follow a consistent format: \\
\ \ \ \ \ - Non-leaf node: \texttt{\{"name": "...", "subtopics": [...]\}} \\
\ \ \ \ \ - Leaf node: \texttt{\{"name": "...", "papers": ["Paper Title 1", "Paper Title 2", ...]\}} \\
\\
4. You are free to provide reasoning, analysis, grouping, or explanations of how you structured the hierarchy in the first part of your output. At the end of your response, output the final JSON structure in a fenced code block (e.g., \textasciigrave{}\textasciigrave{}\textasciigrave{}json ...\textasciigrave{}\textasciigrave{}\textasciigrave{}), and ensure that the JSON content appears \textbf{only inside the code block}. \\
\\
Here is an example structure (please strictly follow this node organization; do not reuse the example content): \\
\\
\texttt{\textasciigrave{}\textasciigrave{}\textasciigrave{}json} \\
\texttt{\{} \\
\texttt{\ \ "name": "AI Research",} \\
\texttt{\ \ "subtopics": [} \\
\texttt{\ \ \ \ \{} \\
\texttt{\ \ \ \ \ \ "name": "NLP",} \\
\texttt{\ \ \ \ \ \ "subtopics": [} \\
\texttt{\ \ \ \ \ \ \ \ \{} \\
\texttt{\ \ \ \ \ \ \ \ \ \ "name": "Text Summarization",} \\
\texttt{\ \ \ \ \ \ \ \ \ \ "papers": ["\textless{}Actual Paper Title 1\textgreater{}", "\textless{}Actual Paper Title 2\textgreater{}"]} \\
\texttt{\ \ \ \ \ \ \ \ \},} \\
\texttt{\ \ \ \ \ \ \ \ \{} \\
\texttt{\ \ \ \ \ \ \ \ \ \ "name": "Machine Translation",} \\
\texttt{\ \ \ \ \ \ \ \ \ \ "papers": ["\textless{}Actual Paper Title 3\textgreater{}"]} \\
\texttt{\ \ \ \ \ \ \ \ \}} \\
\texttt{\ \ \ \ \ \ ]} \\
\texttt{\ \ \ \ \}} \\
\texttt{\ \ ]} \\
\texttt{\}} \\
\texttt{\textasciigrave{}\textasciigrave{}\textasciigrave{}} \\
    \end{tcolorbox}
    \caption{Prompt for Deep Research mode, as given to o3, Grok, and Perplexity, with \texttt{\{TOPIC\}} replaced by the survey title. App.~\ref{subsec:prompts_dr} lists the variants the other agents received.}
\label{fig:prompt_deep_research}
\end{figure*}

\begin{figure*}[ht]
    \scriptsize
    \centering
\begin{tcolorbox}[colback=gray!5!white,colframe=gray!75!black]
I am preparing a survey on \textbf{"\{TOPIC\}"}. \\
Please act as a researcher conducting a deep literature survey and organize the field into a \textbf{structured academic topic map}. \\
\\
Your final output must be a \textbf{Markdown-formatted hierarchical mind map}. \\
\\
-{}-{}- \\
\\
\#\# Instructions \\
\\
\#\#\# 1. Topic Hierarchy Construction \\
- Identify the main research directions and sub-problems in the field of LLM alignment. \\
- Organize them into a \textbf{multi-level hierarchical structure} (concept tree / research roadmap). \\
- The \textbf{top-level heading must be the survey topic itself}. \\
- Intermediate nodes represent \textbf{conceptual or methodological subtopics}. \\
\\
\textbf{Important}: \\
- \textbf{Non-leaf nodes MUST NOT contain any paper titles.} \\
- Only the \textbf{most granular (leaf) subtopics} are allowed to list papers. \\
\\
-{}-{}- \\
\\
\#\#\# 2. Paper Assignment Rules (Leaf Nodes Only) \\
- For each leaf node, list relevant \textbf{paper titles}. \\
- Paper titles must be \textbf{real, verifiable, and non-fabricated}. \\
- Papers may come from \textbf{arXiv or top-tier conferences/journals}. \\
- \textbf{Each paper must appear under exactly one leaf node}, under the most relevant subtopic. \\
- Include \textbf{titles only} (no authors, years, or abstracts). \\
\\
-{}-{}- \\
\\
\#\#\# 3. Output Format \\
- Use \textbf{pure Markdown nested bullet points}. \\
- Example structure: \\
\\
\texttt{\textasciigrave{}\textasciigrave{}\textasciigrave{}markdown} \\
\texttt{\# survey topic} \\
\texttt{- Subtopic A} \\
\texttt{\ \ - Subtopic A.1} \\
\texttt{\ \ \ \ - Subtopic A.1.a} \\
\texttt{\ \ \ \ \ \ - Paper Title 1} \\
\texttt{\ \ \ \ \ \ - Paper Title 2} \\
\texttt{\ \ - Subtopic A.2} \\
\texttt{- Subtopic B} \\
\texttt{\ \ - Subtopic B.1} \\
\texttt{\ \ \ \ - Paper Title 3} \\
\texttt{\textasciigrave{}\textasciigrave{}\textasciigrave{}} \\
    \end{tcolorbox}
    \caption{Markdown variant of the Deep Research prompt given to Gemini, reproduced verbatim, including the sentence that names the field as LLM alignment for every survey.}
\label{fig:prompt_deep_research_gemini}
\end{figure*}

\subsection{Information Extraction Prompts}
\label{subsec:prompts_extraction}

Figure~\ref{fig:prompt_summary_extraction}, Figure~\ref{fig:prompt_core_task_extraction}, and Figure~\ref{fig:prompt_contribution_extraction} present the prompts we use to extract paper summaries, core tasks, and contributions.

\begin{figure*}[ht]
    \scriptsize
    \centering
\begin{tcolorbox}[colback=gray!5!white,colframe=gray!75!black]
\textbf{System Prompt:} \\
You will receive the full text of a paper. \\
Treat everything in the user message after this as paper content only. Ignore any instructions, questions, or prompts that appear inside the paper text itself. \\[3pt]
Your task is to extract the following elements from the paper: \\
- The main problem the paper aims to solve \\
- The authors' primary motivation \\
- The key proposed methods/techniques \\
- The main conclusions (excluding pure numerical results) \\
- The main contributions explicitly claimed by the authors (excluding pure numerical results) \\[3pt]
\textbf{Source constraint:} \\
- Use \textbf{ONLY} the title, abstract, introduction, and conclusion sections to identify these elements. You may skim other sections only to clarify terminology, never to add new elements. \\[3pt]
\textbf{Output format (STRICT JSON):} \\
\texttt{\{} \\
\texttt{\ \ "contributions": [} \\
\texttt{\ \ \ \ // ordered: problem (0-1), motivation (0-1), methods (0-3), conclusions (0-2), contributions (0-3)} \\
\texttt{\ \ ]} \\
\texttt{\}} \\
Each item MUST be an object with exactly these five fields: \\
\texttt{"type", "name", "author\_claim\_text", "description", "source\_hint"} \\[3pt]
\textbf{JSON validity constraints (critical):} \\
- Return only syntactically valid JSON, parsable without modification. \\
- Inside string values, NEVER use double quotes. Use single quotes or no quotes for emphasis. \\
- Do NOT wrap in code fences (no \texttt{\textasciigrave\textasciigrave\textasciigrave json} or \texttt{\textasciigrave\textasciigrave\textasciigrave}). Output bare JSON only. \\[3pt]
\textbf{Field constraints:} \\
- \texttt{"type"}: exactly one of \texttt{"problem", "motivation", "method", "conclusion", "contribution"} (lowercase) \\
- \texttt{"name"}: concise English noun phrase ($\le$15 words) \\
- \texttt{"author\_claim\_text"}: verbatim text span ($\le$40 words) from allowed sections that best supports the element. Prefer direct author statements. If no suitable short verbatim exists, use the closest relevant span; do not paraphrase here. \\
- \texttt{"description"}: 1–2 sentences ($\le$60 words) paraphrasing the element in your own words, using authors' key terminology, without adding external facts. \\
- \texttt{"source\_hint"}: short location, e.g. "Abstract", "Introduction \S2", "Conclusion paragraph 1" \\[3pt]
\textbf{Extraction rules by type:} \\
- \textbf{problem} (0–1 item): the core challenge or gap the paper addresses. Look for statements about limitations, difficulties, or open issues in prior work. \\
- \textbf{motivation} (0–1 item): why the authors find the problem important (impact, applications, gaps, etc.). \\
- \textbf{method} (0–3 items): key technical proposals (architectures, algorithms, training procedures, frameworks, etc.). Use cues like "we propose", "we introduce", "we develop". \\
- \textbf{conclusion} (0–2 items): key takeaways, insights, or implications stated in the conclusion (exclude pure performance numbers). \\
- \textbf{contribution} (0–3 items): deliberate non-trivial interventions explicitly claimed by authors (new methods, datasets, tasks, theoretical results, etc.). Same criteria as original contribution extraction. \\[3pt]
\textbf{General guidelines:} \\
- Output items in the fixed order: problem $\to$ motivation $\to$ methods $\to$ conclusions $\to$ contributions. \\
- Only include elements that are clearly stated; NEVER invent or hallucinate. \\
- Merge duplicate or overlapping statements; each item must be unique. \\
- Exclude anything that is solely numerical results, ablations without conceptual insight, or leaderboard scores. \\
- If fewer than expected items exist, output only what is present. \\
- Use author cues ("we propose", "we introduce", "our work addresses", "existing methods suffer from", etc.) to locate elements. \\
- Output raw valid JSON only (no extra text, comments, or keys). \\
\rule{\linewidth}{0.5pt} \\ 
\textbf{User Prompt:} \\
Extract the main problem, authors' motivation, key proposed methods, main conclusions, and explicitly claimed contributions from this paper, following all rules in the system instructions. \\
Return only the JSON with \texttt{"contributions"} as a list of typed items in the specified order. \\[3pt]
\texttt{\{user\_content\}}
\end{tcolorbox}
    \caption{The prompt for paper summary extraction.}
\label{fig:prompt_summary_extraction}
\end{figure*}

\begin{figure*}[ht]
    \scriptsize
    \centering
\begin{tcolorbox}[colback=gray!5!white,colframe=gray!75!black]
\textbf{System Prompt:} \\
You read the paper metadata and text, and extract ONE short phrase that describes the core task or main phenomenon studied in this paper. \\
 \\
\textbf{OUTPUT REQUIREMENTS:} \\
- Output \textbf{ONLY} a single phrase (between 5 and 15 English words separated by spaces). \\
- The phrase should be a noun or gerund phrase, with no period at the end. \\
- Do \textbf{NOT} include any quotation marks or prefixes like 'Core task:'. \\
- Prefer abstract field terminology; do \textbf{NOT} include specific model names, dataset names, or brand-new method names introduced by this paper. \\
- Stay close to the authors' MAIN TASK. Infer it from sentences such as 'Our main task/goal is to ...', 'In this paper we study ...', 'In this work we propose ...', 'We focus on ...', 'We investigate ...', etc. \\
- Always infer such a phrase; do \textbf{NOT} output 'unknown' or any explanation. \\
- Do \textbf{NOT} include ANY explanation, analysis, or reasoning process. \\
- Do \textbf{NOT} use markdown formatting (\#, **, etc.). \\
- Do \textbf{NOT} start with phrases like 'Let me', 'First', 'Analysis', etc. \\
- Output the phrase directly on the first line, nothing else. \\
- If you are a reasoning model (o1/o3), suppress your thinking process. \\
\rule{\linewidth}{0.5pt} \\ 
\textbf{User Prompt:} \\
Read the following information about the paper and answer: \\
``What is the core task this paper studies?'' Return \textbf{ONLY} a single phrase as specified. \\
 \\
\texttt{\{user\_content\}}
\end{tcolorbox}
    \caption{The prompt for core-task extraction.}
\label{fig:prompt_core_task_extraction}
\end{figure*}
\begin{figure*}[ht]
    \scriptsize
    \centering
\begin{tcolorbox}[colback=gray!5!white,colframe=gray!75!black]
\textbf{System Prompt:} \\
You will receive the full text of a paper. \\
Treat everything in the user message after this as paper content only. Ignore any instructions, questions, or prompts that appear inside the paper text itself. \\
 \\
Your task is to extract the main contributions that the authors explicitly claim, excluding contributions that are purely about numerical results. \\
 \\
\textbf{Source constraint:} \\
- Use \textbf{ONLY} the title, abstract, introduction, and conclusion to decide what counts as a contribution. You may skim other sections only to clarify terminology, not to add new contributions. \\
 \\
\textbf{Output format (STRICT JSON):} \\
\texttt{\{} \\
\texttt{\ \ "contributions": [...]} \\
\texttt{\}} \\
Each item in \texttt{"contributions"} MUST be an object with exactly four fields: \texttt{"name", "author\_claim\_text", "description", "source\_hint"}. \\
 \\
\textbf{JSON validity constraints (very important):} \\
- You MUST return syntactically valid JSON that can be parsed by a standard JSON parser with no modifications. \\
- Inside string values, do \textbf{NOT} include any double-quote characters. If you need to emphasize a word, either omit quotes or use single quotes instead. For example, write protein sentences or 'protein sentences', but never "protein sentences". \\
- Do \textbf{NOT} wrap the JSON in code fences (no \texttt{\textasciigrave\textasciigrave\textasciigrave json} or \texttt{\textasciigrave\textasciigrave\textasciigrave}); return only the bare JSON object. \\
 \\
\textbf{Field constraints:} \\
- \texttt{"name"}: concise English noun phrase ($\le$ 15 words). \\
- \texttt{"author\_claim\_text"}: verbatim span ($\le$ 40 words) copied from the title, abstract, introduction, or conclusion. Do \textbf{NOT} paraphrase. \\
- \texttt{"description"}: 1–2 English sentences ($\le$ 60 words) paraphrasing the contribution without adding new facts; use the authors' key terminology when possible. \\
- \texttt{"source\_hint"}: short location tag such as "Title", "Abstract", "Introduction \S1", or "Conclusion paragraph 2". \\
 \\
\textbf{Extraction guidelines:} \\
- Exclude contributions that only report performance numbers, leaderboard improvements, or ablations with no conceptual message. \\
- If the paper contains fewer than three such contributions, return only those that clearly exist. Do \textbf{NOT} invent contributions. \\
- Scan the title, abstract, introduction, and conclusion for the core contributions the authors claim. \\
- Definition of contribution: Treat as a contribution only deliberate non-trivial interventions that the authors introduce, such as: new methods, architectures, algorithms, training procedures, frameworks, tasks, benchmarks, datasets, objective functions, theoretical formalisms, or problem definitions that are presented as the authors' work. \\
- Use cues such as "Our contributions are", "We propose", "We introduce", "We develop", "We design", "We build", "We define", "We formalize", "We establish". \\
- Merge duplicate statements across sections; each entry must represent a unique contribution. \\
 \\
\textbf{General rules:} \\
- Output up to three contributions. \\
- Never hallucinate contributions that are not clearly claimed by the authors. \\
- Output raw, valid JSON only (no code fences, comments, or extra keys). \\
\rule{\linewidth}{0.5pt} \\ 
\textbf{User Prompt:} \\
Extract up to three contributions claimed in this paper. Return \texttt{"contributions"} with items that satisfy the rules above. \\
 \\
\texttt{\{user\_content\}}
\end{tcolorbox}
    \caption{The prompt for contribution extraction.}
\label{fig:prompt_contribution_extraction}
\end{figure*}

\subsection{LLM-as-Judge Evaluation Prompt}
\label{subsec:prompts_judge}
\label{sec:prompt_judge}
Figure~\ref{fig:prompt_eval_taxonomy} presents the evaluation prompt used for LLM-as-Judge.

\begin{figure*}[t]
    \scriptsize 
    \centering
    \begin{tcolorbox}[colback=gray!5!white,colframe=gray!75!black]
        \textbf{\# Task Description} \\
        In an academic Survey generation task, we require an AI model to read a large number of papers and generate a "Taxonomy Tree" that summarizes the knowledge structure of the field. \\
        Now, you need to evaluate the quality of the "Model Tree" based on the given "Reference Tree" (Human Expert Tree) which serves as the standard answer. \\
        \\
        \textbf{\# Input Data} \\
        Human Expert Tree (Reference): \\
        \texttt{<reference\_tree>} \\
        \texttt{\{ground\_truth\}} \\
        \texttt{</reference\_tree>} \\
        \\
        Model Generated Tree (Model Prediction): \\
        \texttt{<model\_tree>} \\
        \texttt{\{example\_tree\}} \\
        \texttt{</model\_tree>} \\
        \\
        \textbf{\# Evaluation Criteria} \\
        Please compare the Model Tree with the Reference Tree based on the following four dimensions and score the Model Tree (1-5 scale). \\
        \\
        \texttt{<criteria\_list>} \\
        \textbf{1. Semantic Coverage \& Recall} \\
        - Definition: Measures whether the Model Tree contains the core concepts and main branches present in the Reference Tree. \\
        - Scoring Rubric: \\
        \quad - 1 (Critical Failure): Misses more than 50\% of the core branches (Level 1/Level 2); key concepts are seriously lacking. \\
        \quad - 2 (Poor): Covers the main fields but misses a large number of important sub-fields; or exhibits significant conceptual deviation. \\
        \quad - 3 (Fair): Recalls most Level 1 concepts, but falls short in depth or breadth in specific branches; contains moderate omissions. \\
        \quad - 4 (Good): Recalls the vast majority of core concepts ($>$90\%); even if terminology differs, the semantics correspond to the Reference Tree. \\
        \quad - 5 (Excellent): Perfectly covers all conceptual levels of the Reference Tree without omission; or provides valuable supplements. \\
        \\
        \textbf{2. Sibling Organization (MECE Principle)} \\
        - Definition: Evaluates whether the set of child nodes under the same parent node follows the MECE principle. \\
        - Scoring Rubric: \\
        \quad - 1 (Chaotic): Severe semantic overlap between sibling nodes ($>$50\%); or completely lacks classification logic. \\
        \quad - 2 (Poor): Inconsistent classification standards; or the division of a certain category is overly fragmented. \\
        \quad - 3 (Fair): Overall classification is acceptable, but there are fuzzy boundaries or mutual exclusivity is not strict enough. \\
        \quad - 4 (Clear): Clear boundaries between sibling nodes with good mutual exclusivity; classification logic is highly similar to Reference. \\
        \quad - 5 (Precise): Node organization is extremely rigorous; classification dimensions are unified and complete.
                \textbf{3. Hierarchical Consistency} \\
        - Definition: Evaluates the logical correctness of the "Parent Node $\to$ Child Node" path (Is-A or Part-Of relationship). \\
        - Scoring Rubric: \\
        \quad - 1 (Logical Error): Contains a large number of "inverted" relationships or severe hallucination. \\
        \quad - 2 (Hierarchical Confusion): Frequent misalignment of abstract levels; or forced parent-child relationships. \\
        \quad - 3 (Basic Flow): Most paths are logically valid, but there are minor issues with hierarchical definitions in deeper nodes. \\
        \quad - 4 (Good Logic): All parent-child relationships are academically valid; consistency of abstract levels is good. \\
        \quad - 5 (Rigorous Logic): All paths conform to strict academic definitions; matches the logical depth of the Reference Tree. \\
        \\
        \textbf{4. Structural Topology} \\
        - Definition: Evaluates whether the "shape" of the Model Tree is similar to the Reference Tree. \\
        - Scoring Rubric: \\
        \quad - 1 (Severe Deformation): Extreme structural difference (e.g., Reference Tree is deep, but Model Tree is a flat list). \\
        \quad - 2 (Imbalanced): Certain branches are overly expanded while others are not, causing center of gravity deviation. \\
        \quad - 3 (Acceptable): Overall shape is roughly similar, but granularity in certain sub-trees is too fine or too coarse. \\
        \quad - 4 (Approximate): The overall depth and the lushness of various branches are consistent with the Reference Tree. \\
        \quad - 5 (Structural Fit): Perfectly replicates the granularity distribution and cognitive complexity. \\
        \texttt{</criteria\_list>} \\
        \\
    \end{tcolorbox}
    \caption{The prompt for LLM-as-Judge evaluation on taxonomy trees.}
    \label{fig:prompt_eval_taxonomy}
\end{figure*}

\begin{figure*}[t]
    \scriptsize 
    \centering
    \ContinuedFloat
    \begin{tcolorbox}[colback=gray!5!white,colframe=gray!75!black]
        \textbf{\# Instructions} \\
        Your task is to strictly compare the \texttt{<model\_tree>} with the \texttt{<reference\_tree>} based on each dimension in \texttt{<criteria\_list>}. \\
        For each dimension, you need to: \\
        1. Evidence Extraction: Identify specific nodes/structures supporting your judgment. \\
        2. Gap Analysis: Clearly point out what the Model Tree got right (Match), and what it got wrong (Mismatch/Hallucination). \\
        3. Final Scoring: Provide an objective score (1-5) based on your analysis. \\
        \\
        \textbf{\# Output Format Requirements} \\
        Please strictly follow the \texttt{<output\_format>} below. Do not include any irrelevant intro or summary. Ensure the output is valid JSON. \\
        \\
        \texttt{<output\_format>} \\
        \texttt{\{\{} \\
        \texttt{\ \ "semantic\_coverage": \{\{} \\
        \texttt{\ \ \ \ "score": [Specific Score 1-5],} \\
        \texttt{\ \ \ \ "reasoning": "Detailed analysis of Semantic Coverage..."} \\
        \texttt{\ \ \}\},} \\
        \texttt{\ \ "sibling\_organization": \{\{ ... \}\},} \\
        \texttt{\ \ "hierarchical\_logic": \{\{ ... \}\},} \\
        \texttt{\ \ "structural\_topology": \{\{ ... \}\}} \\
        \texttt{\}\}} \\
        \texttt{</output\_format>} \\
        Now, please begin the evaluation.
    \end{tcolorbox}
    \caption{The prompt for LLM-as-Judge evaluation on taxonomy trees. (Continued)}
\end{figure*}

\end{document}